\documentclass{article} % For LaTeX2e
\usepackage{amsmath, latexsym}
\usepackage{amsfonts}
\usepackage{mathtools}
\usepackage{amsthm}
\usepackage{dsfont}
\usepackage[dvipsnames]{xcolor}
\usepackage[colorinlistoftodos]{todonotes}
\usepackage{booktabs}
\usepackage{xfrac}

\usepackage{algorithm}

\usepackage[most]{tcolorbox}
\usepackage{xparse}
\usepackage{lipsum}
\usepackage{changepage}
\usepackage{enumitem}

\newcommand{\squishlist}{
  \begin{list}{$\bullet$}
    { \setlength{\itemsep}{0pt}      \setlength{\parsep}{3pt}
      \setlength{\topsep}{3pt}       \setlength{\partopsep}{0pt}
      \setlength{\leftmargin}{1.5em} \setlength{\labelwidth}{1em}
      \setlength{\labelsep}{0.5em} } }

\newcommand{\squishlisttwo}{
  \begin{list}{$\bullet$}
    { \setlength{\itemsep}{0pt}    \setlength{\parsep}{0pt}
      \setlength{\topsep}{0pt}     \setlength{\partopsep}{0pt}
      \setlength{\leftmargin}{2em} \setlength{\labelwidth}{1.5em}
      \setlength{\labelsep}{0.5em} } }

\newcommand{\squishend}{
    \end{list}  }

\newtheorem{prop}{Proposition}

\newtheorem{thm}{Theorem}

\newtheorem{cor}{Corollary}
\newtheorem{defn}{Definition}

\newcommand{\calA}{{\cal A}}

\newcommand{\calE}{{\cal E}}

\newcommand{\calL}{{\cal L}}

\newcommand{\calX}{{\cal X}}

\newcommand{\grad}{\nabla}
\newcommand{\Etp}{\calE_{t^+}}

\newcommand{\Xtp}{X_{t^+}}

\newcommand{\ft}{f^X}
\newcommand{\fpi}{f^\pi}

\newcommand{\pa}{\text{pa}}

\newcommand{\thetaf}{{\theta_{\text{fs}}}}

\newcommand{\thetah}{{\theta_{\text{hs}}}}
\newcommand{\thetaV}{{\theta_{\text{V}}}}
\newcommand{\om}{{\omega}}
\newcommand{\MI}{{\mathbb{I}}}
\newcommand{\entH}{{\mathbb{H}}}
\newcommand{\E}{\mathbb{E}}
\newcommand{\V}{\mathbb{V}}

\newcommand{\hlaa}{\text{GENE}_\text{A}}
\newcommand{\hlab}{\text{GENE}_\text{B}}
\newcommand{\hlac}{\text{GENE}_\text{C}}

\newcommand{\red}[1]{\textcolor{red}{#1}}
\newcommand{\blue}[1]{\textcolor{blue}{#1}}
\newcommand{\betaim}{\beta_{\text{IM}}}
\newcommand{\dbyd}[2]{\frac{\partial #1}{\partial #2}}
\usetikzlibrary{calc}
\usepackage{zref-savepos}

\newcounter{NoTableEntry}
\renewcommand*{\theNoTableEntry}{NTE-\the\value{NoTableEntry}}

\newcommand*{\notableentry}{%
  \multicolumn{2}{c|}{%
    \stepcounter{NoTableEntry}%
    \vadjust pre{\zsavepos{\theNoTableEntry t}}% top
    \vadjust{\zsavepos{\theNoTableEntry b}}% bottom
    \zsavepos{\theNoTableEntry l}% left
    \hspace{0pt plus 1filll}%
    \zsavepos{\theNoTableEntry r}% right
    \tikz[overlay]{%
      \draw[black]
        let
          \n{llx}={\zposx{\theNoTableEntry l}sp-\zposx{\theNoTableEntry r}sp},
          \n{urx}={0},
          \n{lly}={\zposy{\theNoTableEntry b}sp-\zposy{\theNoTableEntry r}sp},
          \n{ury}={\zposy{\theNoTableEntry t}sp-\zposy{\theNoTableEntry r}sp}
        in
        (\n{llx}, \n{lly}) -- (\n{urx}, \n{ury})
        (\n{llx}, \n{ury}) -- (\n{urx}, \n{lly})
      ;
    }% 
  }%
}
\DeclareMathSymbol{\shortminus}{\mathbin}{AMSa}{"39}

\newcommand{\doop}{\mathrm{do}}
\usepackage[accepted]{icml2021}
\usepackage{times}

\usepackage{booktabs}
\usepackage{multirow}
\usepackage[normalem]{ulem}
\useunder{\uline}{\ul}{}
\usepackage{hyperref}
\usepackage{url}
\usepackage[utf8]{inputenc} % allow utf-8 input
\usepackage[T1]{fontenc}    % use 8-bit T1 fonts
\usepackage[authoryear, sort&compress, round]{} 
\usepackage{afterpage}
\usepackage{amsfonts}
\usepackage{amsmath}
\usepackage{amstext}
\usepackage{bm}
\usepackage{subfigure}
\usepackage{caption}
\usepackage{colortbl}
\usepackage{float}
\usepackage{placeins}
\usepackage{tikz}
\usepackage{times}
\usetikzlibrary{shapes,snakes}
\usetikzlibrary{positioning, arrows.meta}
\tikzset{main node/.style={circle,draw,minimum size=0.8cm,inner sep=0pt},}
\usetikzlibrary{decorations.pathreplacing}
\usetikzlibrary{decorations.pathmorphing}
\tikzset{snake it/.style={decorate, decoration=snake}}
\usepackage[font=small,labelfont=bf]{caption}

\newcommand{\RNum}[1]{\uppercase\expandafter{\romannumeral #1\relax}}

\usepackage{sidecap}
\usepackage{microtype}
\usepackage{graphicx}
\usepackage{selectp}
\icmltitlerunning{Counterfactual Credit Assignment}

\begin{document}

\twocolumn[
\icmltitle{Counterfactual Credit Assignment \\
           in Model-Free Reinforcement Learning}

\icmlsetsymbol{equal}{*}

\begin{icmlauthorlist}
\icmlauthor{Thomas Mesnard}{equal,dm}
\icmlauthor{Th\'eophane Weber}{equal,dm}
\icmlauthor{Fabio Viola}{dm}
\icmlauthor{Shantanu Thakoor}{dm}
\icmlauthor{Alaa Saade}{dm}
\icmlauthor{Anna Harutyunyan}{dm}
\icmlauthor{Will Dabney}{dm}
\icmlauthor{Tom Stepleton}{dm}
\icmlauthor{Nicolas Heess}{dm}
\icmlauthor{Arthur Guez}{dm}
\icmlauthor{\'Eric Moulines}{cmap}
\icmlauthor{Marcus Hutter}{dm}
\icmlauthor{Lars Buesing}{dm}
\icmlauthor{R\'emi Munos}{dm}
\end{icmlauthorlist}

\icmlaffiliation{dm}{DeepMind}
\icmlaffiliation{cmap}{INRIA XPOP, CMAP, \'Ecole Polytechnique, Palaiseau, France}

\icmlcorrespondingauthor{Th\'eophane Weber}{theophane@deepmind.com}
\icmlcorrespondingauthor{Thomas Mesnard}{mesnard@deepmind.com}
\icmlkeywords{Machine Learning, ICML}

\vskip 0.3in
]
\printAffiliationsAndNotice{\icmlEqualContribution}

\begin{abstract}
Credit assignment in reinforcement learning is the problem of measuring an action’s influence on future rewards. 
In particular, this requires separating \emph{skill} from \emph{luck}, i.e.\ disentangling the effect of an action on rewards from that of external factors and subsequent actions.
To achieve this, we adapt the notion of counterfactuals from causality theory to a model-free RL setup. 
The key idea is to condition value functions on \emph{future} events, by learning to extract relevant information from a trajectory. 
We formulate a family of policy gradient algorithms that use these future-conditional value functions as baselines or critics, and show that they are provably low variance. To avoid the potential bias from conditioning on future information, we constrain the hindsight information to not contain information about the agent's actions. We demonstrate the efficacy and validity of our algorithm on a number of illustrative and challenging problems.

\end{abstract}
\section{Introduction}

Reinforcement learning (RL) agents act in their environments and learn to achieve desirable outcomes by maximizing a reward signal. A key difficulty is the problem of \emph{credit assignment} \citep{minsky1961steps}, i.e.\ to understand the relation between actions and outcomes, and to determine to what extent an outcome was caused by external, uncontrollable factors. In doing so we aim to disentangle the relative aspects of `skill' and `luck' in an agent's performance.

One possible solution to this problem is for the agent to build a model of the environment, and use it to obtain a more fine-grained understanding of the effects of an action. While this topic has recently generated a lot of interest \citep{heess2015learning, ha2018world,hamrick2019analogues,kaiser2019model,schrittwieser2019mastering}, it remains difficult to model complex, partially observed environments.

In contrast, model-free reinforcement learning algorithms such as policy gradient methods \citep{williams1992simple,sutton2000policy} perform simple time-based credit assignment, where events and rewards happening after an action are credited to that action, \emph{post hoc ergo propter hoc}. While unbiased in expectation, this coarse-grained credit assignment typically has high variance, and the agent will require a large amount of experience to learn the correct relation between actions and rewards. 
Another issue is that existing model-free methods are not capable of \emph{counterfactual reasoning}, i.e.\ reasoning about what would have happened had different actions been taken \emph{with everything else remaining the same}. Given a trajectory, model-free methods can in fact only learn about the actions that were actually taken to produce the data, and this limits the ability of the agent to learn efficiently.

As environments grow in complexity due to partial observability, scale, long time horizons, and increasing number of agents, actions taken by an agent will only affect a vanishing part of the outcome, making it increasingly difficult to learn from classical reinforcement learning algorithms. We need better credit assignment techniques.

In this paper, we investigate a new method of credit assignment for model-free reinforcement learning which we call \emph{Counterfactual Credit Assignment} (CCA). CCA leverages \emph{hindsight} information to implicitly perform counterfactual evaluation--an estimate of the return for actions other than the ones which were chosen. These counterfactual returns can be used to form unbiased and lower variance estimates of the policy gradient by building future-conditional baselines.
Unlike classical Q functions, which also provide an estimate of the return for all actions but do so by averaging over all possible futures, our methods provide trajectory-specific counterfactual estimates, i.e.\ an estimate of the return for different actions, but keeping as many of the external factors constant between the return and its counterfactual estimate\footnote{From from a causality standpoint, one-step action-value functions are interventional concepts (``What would happen if") instead of counterfactuals (``What would have happened if").}. Such a method would perform finer-grained credit assignment and could greatly improve data efficiency in environments with complex credit assignment structures. 
Our method is inspired by ideas from causality theory, but does not require learning a model of the environment.

Our main contributions are: a) introducing a family of novel policy gradient estimators that leverage hindsight information and generalizes previous approaches, b) proposing a practical instantiation of this algorithm  with sufficiency conditions for unbiasedness and guarantees for lower variance, c) introducing a set of environments which further our understanding of when credit assignment is made difficult due to exogenous noise, long-term effects and task interleaving, and thus leads to poor policy learning, d) demonstrating the improved performance of our algorithm on these environments, e) formally connecting our results to notions from causality theory, in particular treatment effects and counterfactuals, further linking the causal inference and reinforcement learning literature.

\section{Counterfactual Credit Assignment}
\label{sec:highlevelidea}
\subsection{Notation}\label{subsec:notation}
\emph{We use capital letters for random variables and lowercase for the value they take.} 
Consider a generic MDP $(\calX, \calA, p, r, \gamma)$. Given a current state $x \in \calX$ and assuming an agent takes action $a \in \calA$, the agent receives reward $r(x,a)$ and transitions to a state $y \sim p(\cdot|x,a)$. The state (resp. action, reward) of the agent at step $t$ is denoted $X_t$ (resp. $A_t$, $R_t$). The initial state of the agent $X_0$ is a fixed $x_0$.
The agent acts according to a policy $\pi$, i.e.\ action $A_t$ is sampled from the policy $\pi_\theta(\cdot|X_t)$ where $\theta$ are the policy parameters, and aims to optimize the expected discounted return $\E[G]= \E[\sum_t \gamma^t R_t]$. The return $G_t$ from step $t$ is $G_t=\sum_{t'\geq t} \gamma^{t'-t} R_{t'}$. Note $G=G_0$. Finally, we define the score function $s_\theta(\pi_\theta, a, x) = \grad_\theta \log \pi_\theta(a|x)$; the score function at time $t$ is denoted $S_t=\grad_\theta \log \pi_\theta(A_t|X_t)$. In the case of a partially observed environment, we assume the agent receives an observation ${{E}}_t$ at every time step, and simply define $X_t$ to be the set of all previous observations, actions and rewards $X_t=(O_{\leq t})$, with $O_t=({{E}}_t, A_{t-1}, R_{t-1})$.\footnote{Previous actions and rewards are provided as part of the observation as it is generally beneficial to do so in partially observable Markov decision processes.}
We use $P$ to denote the probability distribution of random variables according to the data distribution, e.g. $P(X)$ denote the probability distribution of a random variable $X$, while we use lowercase letters for distributions according to the model, e.g. $p(x)$ is the distribution of $x$ according to the model $p$.

\subsection{Policy gradient algorithms}

We begin by recalling two forms of policy gradient algorithms and the credit assignment assumptions they make. The first is the REINFORCE algorithm introduced by \citet{williams1992simple}, which we will also call the single-action policy gradient estimator. The gradient of $\E[G]$ is given by:
\begin{equation}\grad_\theta \E[G] = \E\Big[ \sum_{t\geq 0} \gamma^t\: S_t\: \left(G_t-V(X_t)\right)\label{eq:REINFORCE} \Big]\text{,}\end{equation}
where $V(X_t)=\E[G_t|X_t]$. Let's note here that $V(X_t)$ (resp. $Q(X_t, A_t)=\E[G_t|X_t, A_t]$) is the value function (resp. Q-function) for the policy $\pi_{\theta}$ but for notation simplicity the dependence on the policy will be implicit through the rest of this paper.

The appeal of this estimator lies in its simplicity and generality: to evaluate it, the only requirement is the ability to simulate trajectories, and compute both the score function and the return.\\

Let us note two credit assignment features of the estimator. 
First, the score function $S_t$ is multiplied not by the whole return $G$, but by the return from time $t$. Intuitively, action $A_t$ can only affect states and rewards coming after time $t$, and it is therefore pointless to credit action $A_t$ with past rewards. Second, subtracting the value function $V(X_t)$ from the return $G_t$ does not bias the estimator and typically reduces variance, since the resulting estimate makes an action $A_t$ more likely proportionally not to the return, but to which extent the return was higher than what was expected before the action was taken \cite{williams1992simple}.
Such a function will be called a \emph{baseline} in the following.  In theory, the baseline can be any function of $X_t$; note in particular that for a partially observed environment, this means the baseline can be any function of \emph{past} observations.
It is however typically assumed that it does not depend on any variable `from the future' (including the action about to be taken, $A_t$), i.e.\ with time index greater than $t$, since including variables which are (causally) affected by the action generally results in a biased estimator \citep{weber2019credit}.

This estimator updates the policy through the score term; note however the learning signal only updates the policy $\pi_\theta(a\vert X_t)$ for the taken action $A_t=a$ (other actions are only updated through normalization of action probabilities).
The policy gradient theorem from \citet{sutton2000policy}, which we will also call all-action policy gradient, shows it is possible to provide learning signal to all actions, given we have access to a Q-function, $Q(x,a)=\E[G_t|X_t=x, A_t=a]$, which we will call a \emph{critic} in the following. The gradient of $\E[G]$ is given by: \begin{equation}
    \label{eq:AAPG}
\grad_\theta \E[G] = \E\Big[ \sum_{t\geq 0} \gamma^t \sum_{a\in\calA} \grad_\theta \pi(a|X_t) Q(X_t,a) \Big].\end{equation}
A particularity of the all-actions policy gradient estimator is that the term at time $t$ for updating the policy $\grad \pi(a|X_t)Q(X_t,a)$ depends only on current information (past information in a POMDP); this is in contrast with the score function estimates above which depend on the return, a function of the entire trajectory.
\subsection{Intuitive example on hindsight reasoning and skill versus luck}
\label{sec:story}

Imagine a scenario in which Alice just moved to a new city, is learning to play soccer, and goes to the local soccer field to play a friendly game with a group of other kids she has never met. As the game goes on, Alice does not seem to play at her best and makes some mistakes. It turns out however her partner Megan is a strong player, and eventually scores the goal that makes the game a victory. What should Alice learn from this game? 

When using the single-action policy gradient estimate, the outcome of the game being a victory, and assuming a $\pm 1$ reward scheme, all her actions taken during the game are made more likely; this is in spite of the fact that during this particular game she may not have played well and that the victory is actually due to her strong teammate. From an RL point of view, her actions are wrongly credited for the victory and positively reinforced as a result; effectively, Alice was lucky rather than skillful.
Regular baselines do not mitigate this issue, as Alice did not a priori know the skill of Megan, resulting in an assumption that Megan was of average strength and therefore a guess that their team had a $50\%$ chance of winning. This could be fixed by understanding that Megan's strong play were not a consequence of Alice's play, that her skill was a priori unknown but known in hindsight, and that it is therefore valid to retroactively include her skill level in the baseline. A hindsight baseline, conditioned on Megan's estimated skill level, would therefore be closer to $1$, driving the advantage estimate (and corresponding learning signal) close to $0$. 

As pointed out by \citet{buesing2018woulda}, situations in which hindsight information is helpful in understanding a trajectory are frequent.  In that work, the authors adopt a model-based framework, where hindsight information is used to ground counterfactual trajectories (i.e.\ trajectories under \textit{different actions, but same randomness}). 
Our proposed approach follows a similar intuition, but is model-free: we attempt to \emph{measure}---instead of model---  information known in hindsight to compute a \emph{future-conditional baseline}, but in a way that maintains unbiasedness. As we will see later, this corresponds to a constraint that the captured information must not have been caused by the agent.

\subsection{Future-conditional (FC-PG) and Counterfactual (CCA-PG) Policy Gradient Estimators}

Intuitively, our approach for assigning proper credit to action $A_t$ relies on measuring statistics $\Phi_t$ that capture relevant information from the trajectory (e.g.\ including observations $O_{t'}$ at times $t'$ greater than $t$). We then learn value functions or critics which are conditioned on the additional hindsight information contained in $\Phi_t$.  In general, these future-conditional values and critics would be biased for use in a policy gradient algorithm; we therefore use an importance correction term to eliminate this bias.

\begin{thm}[Future-Conditional Policy Gradient (FC-PG) estimators]
\label{thm:FC-PG}
Let $\Phi_t$ be an arbitrary random variable. Assuming that $\frac{\pi(a|X_t)}{{\mathbb P}(a|X_t, \Phi_t)}<\infty$ for all $a$, the following is the single-action unbiased estimator of the gradient of $\E[G]$:
\begin{align}
\grad_\theta \E[G] =\E\Big[ \sum_t \gamma^t\: S_t \Big(G_t-\frac{\pi(A_t|X_t)}{{\mathbb P}(A_t|X_t, \Phi_t)}V(X_t, \Phi_t)\Big) \Big]
\end{align}
where  $V(x, \phi)=\E[G_t|X_t=x, \Phi_t=\phi]$ is the future $\Phi$-conditional value function
\footnote{Note more generally that any function of $X_t$ and $\Phi_t$ can in fact be used as a valid baseline.}.

With no requirements on $\Phi_t$, we also have an all-action unbiased estimator:
\begin{align*}
\hspace{-3pt}\grad_\theta \E[G] \hspace{-1pt}=\hspace{-1pt} \E\big[ \sum_{t,a} \gamma^t \grad_\theta \log \pi(a|X_t) P(a|X_t,\Phi_t) Q(X_t,\Phi_t,a) \big]
\end{align*}
where $Q(x, \phi, a)=\E[G_t|X_t=x, \Phi_t=\phi, A_t=a]$ is the future-conditional Q function (critic). Furthermore, we have $Q(X_t, a)=\E\left[Q(X_t,\Phi_t,a) \frac{P(a|X_t,\Phi_t)}{\pi(a|X_t)}\right]$.
\end{thm}

Intuitively, the $\frac{\pi(a|X_t)}{{\mathbb P}(a|X_t, \Phi_t)}<\infty$ condition means that knowing $\Phi_t$ should not preclude any action $a$ which was possible for $\pi$ from having potentially produced $\Phi_t$. A counterexample is $\Phi_t=A_t$; knowing $\Phi_t$ precludes any action $a\not = A_t$ from having produced $\Phi_t$.
Typically, $\Phi_t$ will be chosen to a function of the present and future trajectory $(X_s, A_s, R_s)_{s\geq t}$. The estimators above are very general and generalize similar estimators (HCA) introduced by \citet{harutyunyan2019hindsight} (see App.~\ref{hca-cca-fc} for a discussion of how HCA can be rederived from FC-PG) and different choices of $\Phi$ will have varying properties. $\Phi$ may be hand-crafted using domain knowledge, or, as we will see later, learned using appropriate objectives.
Note that in general an FC-PG estimator doesn't necessarily have lower variance (a good proxy for fine-grained credit assignment) than the classical policy gradient estimator; this is due to the variance introduced by the importance weighting scheme. It would be natural to study an estimator where this effect is nullified through independence of the action and statistics $\Phi$ (resulting in a ratio of $1$).

The resulting advantage estimate could thus be interpreted not just as an estimate of `what outcome should I expect', but also a measure of 'how (un)lucky did I get?' and `what other outcomes might have been possible in this precise situation, had I acted differently'. It will in turn provide finer-grained credit for action $A_t$ in a sense to be made precise below.

\begin{thm}[Counterfactual Policy Gradient (CCA-PG)]\label{prop:cfac}
If $A_t$ is independent from $\Phi_t$ given $X_t$, the following is an unbiased \emph{single-action} estimator of the gradient of $\E[G]$:
\begin{align}\label{eq:CFAC-baseline}
\grad_\theta \E[G] =\E\left[ \sum_t \gamma^t\: S_t \left(G_t-V(X_t, \Phi_t)\right) \right].
\end{align}
Furthermore, the hindsight advantage estimate has no higher variance than the forward one: $$\E\left[\left(G_t-V(X_t, \Phi_t)\right)^2 \right] \leq \E\left[\left(G_t-V(X_t)\right)^2 \right].$$
Similarly, for the \emph{all-action} estimator:
\begin{align}\label{eq:CFAC-critic}
    \hspace{-2pt}\grad_\theta \E[G] = \E\Big[ \sum_t \gamma^t \sum_a \grad_\theta \pi(a|X_t) Q(X_t,\Phi_t,a) \Big].
\end{align}
Also, we have for all $a$, $$Q(X_t,a)=\E[Q(X_t,\Phi_t,a)|X_t, A_t=a]$$
\end{thm}

The benefit of the first estimator (equation~\ref{eq:CFAC-baseline}) is clear: under the specified condition, and compared to the regular policy gradient estimator, the CCA estimator also has no bias, but the variance of its advantage estimate $G_t-V(X_t,\Phi_t)$  (the critical component behind variance of the overall estimator) is no higher. 

For the all-action estimator, the benefits of CCA (equation~\ref{eq:CFAC-critic}) are less self-evident, since this estimator has \emph{higher} variance than the regular all action estimator (which has variance $0$). 
The interest here lies in bias due to learning imperfect Q functions. Both estimators require learning a Q function from data; any error in Q leads to a bias in $\pi$. 
Learning $Q(X_t,a)$ requires averaging over all possible trajectories initialized with state $X_t$ and action $a$: in high variance situations, this will require a lot of data. In contrast, if the agent could measure a quantity $\Phi_t$ which has a high impact on the return but is not correlated to the agent action $A_t$, it could be far easier (and more data-efficient) to learn $Q(X_t,\Phi_t,a)$. This is because $Q(X_t,\Phi_t,a)$ computes the averages of the return $G_t$ conditional on $(X_t,\Phi_t,a)$; if $\Phi_t$ has a high impact on $G_t$, the variance of that conditional return will be lower, and learning its average will in turn be simpler.
Interestingly, note also that $Q(X_t,\Phi_t,a)$ (in contrast to $Q(X_t,a)$) is a \emph{trajectory-specific} estimate of the return for a counterfactual action.

\subsection{Learning the relevant statistics}
\label{subsec:learnstats}
The previous section proposes a sufficient condition on $\Phi$ for useful estimators to be derived. A question remains - how to compute such a $\Phi$ from the trajectory, and can such $\Phi$ be generally guaranteed to exist? In this section, we suggest three possible methods to do so, focusing on one which we will still call CCA for simplicity.

\paragraph{Handcrafted information from prior knowledge.}\textcolor{white}{a}\\
In many cases, domain knowledge may indicate the existence of information unknown at time $t$ but observed in hindsight but which are independent (unaffected) by the action of our agent at time $t$.

For example, when the dynamics of the environment are partially driven by exogenous factors independent of the agent's actions, those factors can be a posteriori included in the baseline at time $t$ \citep{mao2018variance}.
Similarly, in multi-agent environments, actions of other agents at the same time step \citep{foerster2018counterfactual} can be included in the baseline\footnote{More ideas in this direction can be found under the centralized training and decentralized execution (CDTE) framework.}; and so can the full state (at time $t$) of the simulator when learning control from pixels \citep{andrychowicz2018learning}, or the opponent's private observations (at time $t$) in Starcraft II \citep{vinyals2019grandmaster}; finally, when using independent multi-dimensional actions, the baseline for one dimension of the action vector can include the actions in other dimensions \citep{wu2018variance}.

Note however that all of these require privileged information, both in the form of feeding information to the baseline which is potentially inaccessible to the agent, and in knowing that this information is in fact independent from the agent's action $A_t$ and therefore won't bias the baseline. Next, we will suggest approaches which seek to replicate a similar effect, but in a more general fashion and from an agent-centric point of view, where the agent learns itself which information from the future can be used to improve its baseline at time $t$.

\paragraph{Generative CCA and posterior value functions}\textcolor{white}{a}\\
Next, we consider a family of approaches, also suggested by \citet{nota2021posterior} under the name \emph{posterior value functions}, which leverages generative models of data to identify such a $\Phi$. It is also closely related to the explicitly model-based approach of \citet{buesing2018woulda}.
Consider some arbitrary variable of interest $Y_t$, a function of the entire trajectory, therefore known in hindsight. Assume we learn a conditional generative model $p(Y_t|X_t,A_t)$ with a latent $\varepsilon_t$ conditionally independent from $A_t$: 
$$p(Y_t|X_t,A_t) = \int_{\varepsilon_t} p(\varepsilon_t|X_t) p(Y_t|X_t, A_t, \varepsilon_t)\:d \varepsilon_t.$$
This generative model can take many forms, such as a variational autoencoder \citep{kingma2013auto, rezende2014stochastic}, a normalizing flow\footnote{In which case $P(Y_t|X_t, A_t, \varepsilon_t)$ is a dirac distribution defined by an invertible function.}\citep{papamakarios2019normalizing}, a distributional model, or, as in \citet{nota2021posterior}, a state-space model (in which case $\varepsilon_t$ is the state $S_t$ at time $t$). 
Several choices for $Y_t$ can be made, such as future observations $(X_t')_{t'\geq t}$, the entire return $G_t$, or any function of the trajectory which we believe captures relevant information. Note $Y_t$ is not assumed to be independent from the action.

Note that when $Y_t$ is the return, generative CCA is effectively modeling the distribution of the return, an approach pioneered by distributional RL \citep{bellemare2017distributional} (though distributional RL typically uses different methods - more often quantile regression-based -  to learn the distribution of the return).

The probabilistic model $\int_{\varepsilon_t} p(\varepsilon_t|X_t) p(Y_t|X_t, A_t, \varepsilon_t)$ induces a posterior $p(\varepsilon_t | X_t, A_t, Y_t)$\footnote{It also induces a more accurate posterior $p(\varepsilon_t | X_t, A_t, \tau)$ where $\tau$ is the entire trajectory, by treating $\tau$ as an auxiliary variable.}.
The following theorem defines a variable $\Phi_{\varepsilon_t}$ which is independent from the action $A_t$:
\begin{thm}
Assume $p$ perfectly models $Y_t$, and that $p(\varepsilon_t|X_t,A_t,Y_t)$ is the exact posterior of that model.
Consider a sample $\Phi_{\varepsilon_t} \sim p(\varepsilon_t|X_t,A_t,Y_t)$. Then $\Phi_{\varepsilon_t}$ is conditionally independent from $A_t$ given $X_t$.
\end{thm}
This may seem counterintuitive, as \citet{nota2021posterior} notes, the posterior is not independent from the action (in particular, a collection of samples from the posterior is not independent from the action). Nevertheless, a single sample from the posterior recovers conditional independence properties from the prior. 
This readily defines a valid hindsight value function $V(X_t, \Phi_{\varepsilon_t,t})$. Going further, we can average this random hindsight value function, and still get a valid baseline:
\begin{cor}
The posterior value function 
$$V(X_t,A_t,Y_t) = \int_{\varepsilon_t} p(\varepsilon_t|X_t,A_t,Y_t) V(X_t,\varepsilon_t) d\varepsilon_t$$ is uncorrelated to the action $A_t$ and is therefore a valid baseline\footnote{It may seem strange the value takes the action as input - however, the action is required to compute the value, but the value itself is only \emph{uncorrelated} to the action.}.
\end{cor}

The same exact technique applies to the Q function:
\begin{cor}
The posterior Q function 
$$Q(X_t,a, A_t,Y_t) = \int_{\varepsilon_t} p(\varepsilon_t|X_t,A_t,Y_t) Q(X_t, a, \varepsilon_t) d\varepsilon_t$$ is uncorrelated to the action $A_t$ and is therefore a valid all-action critic.
\end{cor}

Note that a possible approach to learn both the generative model and the posterior is to train a variational auto-encoder, in which case the loss used to train both the posterior (encoder) and the model (prior and decoder) is :
\begin{align}
\int_\epsilon q(\epsilon|Y_t) \Big(&\log p(G_t|X_t,A_t,\epsilon_t)\notag \\ + &\log P(\epsilon_t|X_t) - \log q(\epsilon| X_t, A_t, Y_t) \Big) d\epsilon
\end{align}
For a Gaussian distribution with fixed variance, $p(Y|X_t,A_t,\epsilon)$ simply becomes proportional to $\E[(G_t-Q(X_t,A_t,\epsilon_t))^2]$ (where $Q$ is the mean of the distribution). In the case of a state-space VAE \citep{buesing2016stochastic}, Generative CCA is very similar to the approach proposed by \citep{venuto2021policy}.

\paragraph{CCA.}\textcolor{white}{a}\\
The previous approach suggests a model-based approach to identifying hindsight statistics which can be included in a valid baseline. While general, this method does require learning a generative model of some statistics of the observations.
The main approach we investigate in this paper is a model-free approach which \emph{learns to extract} $\Phi$ directly from the trajectory.  We suppose that $\Phi$ is some function $\varphi$ of the entire trajectory, i.e.\ $\Phi_t=\varphi(({X}, {A}, {R}))$. For this approach (which we will adopt for the rest of paper and study experimentally), we adopt a method related to domain adversarial training techniques, see for instance \citet{ganin2016domain}. 

The learning signal will be guided by two objectives: First, we encourage $\Phi_t$ to be predictive of the outcome by training a baseline $V(X_t,\Phi_t)$, since theorem~\ref{prop:cfac} highlights that hindsight features which are predictive of the return lead to a decreased variance of the advantage estimate. 
Second, we encourage $\Phi_t$ to be conditionally independent from $A_t$ by minimizing the loss $\mathcal{L}_{\text{IM}}$, as it is required for the estimator to be valid. 
We will induce conditional independence between $\Phi_t$ and $A_t$ by ensuring an action classifier does not gain any information about $A_t$ when given access to $\Phi_t$. To do so, we suppose we can estimate the posterior distribution $P(A_t|X_t, \Phi_t)$ and consider an associated loss $\mathcal{L}_{\text{IM}}$ which is $0$ if only if $\Phi_t$ and $A_t$ are conditionally independent given $X_t$. 
The following gives an example of such a loss:
\begin{prop}\label{propIM}
Assume that $$E_\Phi[\textrm{KL}(P(A_t|X_t))||P(A_t|X_t,\Phi_t)]=0.$$ Then $A_t$ and $\Phi_t$ are independent conditionally on $X_t$.
\end{prop}

\begin{SCfigure*}
\includegraphics[width=0.55\textwidth]{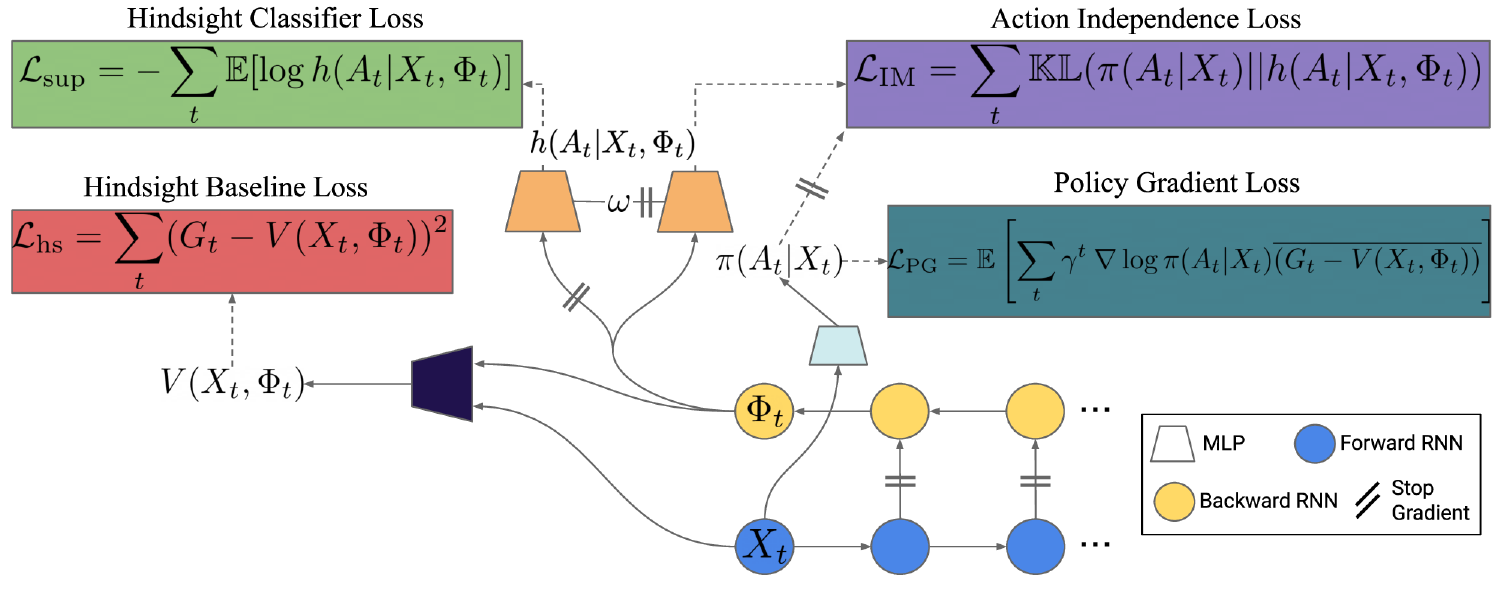}
\caption{{\scriptsize\textbf{Counterfactual Credit Assignment in a nutshell:}
(1) The backward RNN which in this example computes the hindsight features is shaped by the hindsight baseline loss. This ensures that it is predictive of the return. (2) However, to have an unbiased baseline, this hindsight feature $\Phi_t$ needs to be independent from the action $A_t$. To that end, we first train a hindsight predictor that tries to predict what action has been taken a time $t$ from $X_t$ and $\Phi_t$. (3) Then the action independence loss helps removing any information about $A_t$ from the hindsight feature $\Phi_t$ (This only enforces that the output of the backward RNN $\Phi_t$ is independent of the action $A_t$. However, this could potentially translate in $\Phi_t$ being independent from further actions). This loss only impacts the backward RNN and no gradient is being applied to the hindsight predictor MLP. (4) Finally, the policy gradient loss helps improving the policy while no gradient is being sent to the hindsight baseline (i.e\ as expressed by the bar notation).}
}
\label{fig.implementation}
\end{SCfigure*}

To summarize, we want $\Phi$ to be predictive of the return while being independent of the action that is currently being credited. The corresponding hindsight conditional baseline would capture the `luck' part of the outcome while the corresponding advantage estimate would capture the `skill' aspect of it. We detail our agent components and losses in the next section. See also Fig.~\ref{fig.implementation} for a depiction of the resulting architecture and Appendix~\ref{sec:implementation-details} for more details.

\subsection{Practical implementation of CCA-PG}

\textbf{Agent components:}
\begin{itemize}
    \itemsep0em 
    \item \textbf{Agent network}: Our algorithm can generally be applied to arbitrary environments (e.g. POMDPs), so we assume the agent constructs an internal state $X_t$ from past observations $(O_{t'})_{t'\leq t}$ using an arbitrary network, for instance an RNN, i.e.\ $X_t = \text{RNN}_\thetaf(O_t, X_{t-1})$\footnote{Obviously, if the environment is fully observed, a feed-forward network suffices.}. From $X_t$ the agent computes a policy $\pi_{\thetaf}(a|X_t)$, where $\thetaf$ denotes the parameters of the representation network and policy. 
    \item \textbf {Hindsight network}: Additionally, we assume the agent uses a hindsight network $\varphi$ with parameters $\thetah$ which computes a hindsight statistic $\Phi_t=\varphi(({X}, {A}, {R}))$ which
    may depend arbitrarily on the vectors of observations, agent states and actions (in particular, it may depend on observations from timesteps $t'\geq t$).
    \item \textbf {Value network}: The third component is a future-conditional value network  $V_\thetaV(X_t, \Phi_t)$, with parameters $\thetaV$.
    \item \textbf {Hindsight predictor}: The last component is a probabilistic predictor $h_\om$ with parameters $\om$ that takes $X_t, \Phi_t$ as input and outputs a distribution over $A_t$ which is used to enforce the independence condition.
\end{itemize}

\textbf{Learning objectives:}
\begin{itemize}
    \itemsep0em 
    
    \item The first loss is the hindsight baseline loss $\calL_\text{hs}=\sum_t (G_t-V_\thetaV(X_t,\Phi_t))^2$.
    
    \item The second loss is an \emph{independence maximization} (IM) loss  $\calL_{\text{IM}}(X_t)$ which is non-negative and zero if and only if $A_t$ and $\Phi_t$ are conditionally independent given $X_t$. As proposed in the previous section, we choose the Kullback-Leibler divergence\footnote{Other losses are possible, cf. Appendix} between the distributions $P(A_t|X_t)$ and $P(A_t|X_t, \Phi_t)$. In this case, the KL can be estimated by $\sum_a P(a|X_t) \left(\log P(a|X_t)-\log P(a|X_t,\Phi_t) \right)$; $\log P(a|X_t)$ is simply the policy $\pi(a|X_t)$; the posterior $P(a|X_t,\Phi_t)$ is generally not known exactly, but we estimate it with the probabilistic predictor $h_\om(A_t|X_t,\Phi_t)$, which we train with the next loss.
    
    \item The third loss is the hindsight predictor loss, which we train by minimizing the supervised learning loss $\calL_\text{sup}=-\sum_t \E[\log h_\om(A_t|X_t,\Phi_t)]$ on samples $(X_t,A_t,\Phi_t)$ from the trajectory (note that this is a proper scoring rule, i.e. the optimal solution to the loss is the true probability $P(a|X_t,\Phi_t)$).

    \item The last loss is the policy gradients surrogate objective, implemented as $\mathcal{L}_\text{PG}=\sum_t \log \pi_\theta(A_t|X_t) \overline{(G_t-V(X_t,\Phi_t))}$, where the bar notation indicates that 
    the quantity is treated as a constant from the point of view of gradient computation, as is standard.
\end{itemize}

The overall loss is therefore $ \mathcal L = \calL_\text{PG} + \lambda_{\text{hs}}\calL_\text{hs} + \lambda_{\text{sup}}\calL_\text{sup} + \lambda_{\text{IM}}\calL_\text{IM} $. We again want to highlight the very special role played by $\om$ here: only $\calL_\text{sup}$ is optimized with respect to $\om$ (the parameters of the probabilistic predictor), while all the other losses are optimized treating $\om$ as a constant.

\section{Connections to causality}

In this section we provide a formal connection between the CCA-PG estimator and counterfactuals in causality theory (this connection is investigated in greater depth in appendices~\ref{sec:causalRL} and~\ref{sec:simple_ex_causality}). 

To this end, we assume that the MDP $(\calX, \calA, p, r, \gamma)$ in question is generated by an underlying structural causal models (SCM) analogous to \citep{buesing2018woulda, zhang2020designing}. 
In this setting the trajectory $(X_s,A_s,R_s)_{s\geq t}$ and return $G_t$ resulting from the agent-environment interaction is represented as the output of a deterministic function $f$ taking as input the current state $X_t$, the action $A_t$, and a set of exogenous random variables $\bm{\calE}$ which do not have any causal ancestors (in the graph). 
The latter represent the randomness required for sampling all future actions, transitions, and rewards.
Such a "reparametrization" of trajectories and return is always possible, i.e.\ there is always an SCM (possibly non unique) that induces the same 
joint distribution $P$ as the original MDP. 
Intuitively, $\bm{\calE}$ represent all factors external to $A_t$ which affect the outcome\footnote{Note that from this point of view, actions at future time-step are effectively `chance' from the point of view of computing credit for action $A_t$}.

Under that assumption, if the value of $\bm{\calE}$ was known, by virtue of being independent from the action $A_t$, they could be included in a future conditional baseline $V(X_t, \bm{\calE})$; such a baseline would still be unbiased, but since it is conditioned on factors affecting the outcome, has the potential to be significantly more accurate than a forward baseline $V(X_t)$ and lead to lower variance estimator of the gradient. In general, the agent will not know the exact value of $\bm{\calE}$; the CCA-PG attempts to implicitly estimate the value of $\bm{\calE}$ using hindsight information, and leverages that estimate to form more accurate baselines or critics.

SCMs allow to formally define the notion of counterfactual. Given an observed trajectory $\tau = (X_s,A_s,R_s)_{s\geq t}$, we define the counterfactual trajectory $\tau'$ for an alternative action $A_t'=a_t'$ as a the output of the following procedure:
\begin{itemize}
    \itemsep0em 
    \item Abduction: infer the exogenous noise variables ${\epsilon}$ under the factual observation: ${\epsilon} \sim P({\calE}|\tau)$.
    \item Intervention: Fix the value of $A_t'$ to $a_t'$ (mutilating incoming causal arrows).
    \item Prediction: Evaluate the counterfactual outcome $\tau'$ conditional on the fixed values ${\calE}$ and $A_t=a_t'$, yielding $\tau' = f(x_t, a_t',\epsilon)$
\end{itemize}
The counterfactual distribution will be denoted $P(\tau'|\text{observe}(\tau), \text{do}(A_t'=a_t'))$. Note that it typically requires knowledge of the model (SCM) to be computed; samples from the models which do not expose the exogenous variables $\bm{\calE}$ are typically not sufficient to identify the SCM, as several SCMs may correspond to the same distribution.
However, under the CCA assumptions and an additional faithfulness assumption, we can show that the counterfactual return is indeed identifiable and is equal to the future conditional state-action value function:

\begin{thm}\label{thm:causal}
Assume the causal model is faithful (i.e. that conditional independence assumptions are reflected in the graph structure and not only in the parameters). If $\Phi_t$ is conditionally independent from $A_t$ given $X_t$, then the counterfactual distribution, having observed only $\Phi_t$, is identifiable from samples of $(X_t,\Phi_t,A_t)$, and we have
\begin{align}
&\mathbb{E}[G(\tau')| \tau' \sim P(\tau'|X_t=x, \text{observe}(\Phi_t=\phi), \text{do}(A_t'=a)\big]=\notag\\
&Q(X_t=x, A_t=a, \Phi_t=\phi)\end{align}
\end{thm}

The CCA estimator can therefore be understood as follows. Whether we want to compute counterfactuals or posterior-value functions, we need access to a model of the environment. CCA forgoes modeling the environment by discarding information from the whole trajectory $\tau$ which would require a model, and left with still useful information $\Phi_t$ with which counterfactuals can be computed in a model-free way, therefore leading to valid baselines and critics.

\section{Numerical experiments}\label{sec:exps}
Given its guarantees on lower variance and unbiasedness, we run all our experiments on the single action version of CCA-PG and leave the all-action version for future work. We first investigate a bandit with feedback task, then a task that requires short and long-term credit assignment (i.e. Key-to-Door), and finally an interleaved multi-task setup where each episode is composed of randomly sampled and interleaved tasks. All results for Key-to-Door and interleaved multi-task are reported as median performances over 10 seeds with quartiles represented by a shaded area.
\subsection{Bandit with Feedback}
We first demonstrate the benefits of hindsight value functions in a toy problem designed to highlight these. We consider a contextual bandit problem with feedback. 
At each time step, the agent receives a context $-N\leq C\leq N$ (where $N$ is an environment parameter), and based on the context, chooses an action $-N\leq A \leq N$. The agent receives a reward $R=-(C-A)^2+\epsilon_r$, where the exogenous noise $\epsilon_r$ is sampled from $\mathcal{N}(0, \sigma_r)$, as well as a feedback vector $F$ which is a function of $C,A$ and $\epsilon_r$. More details about this problem as well as variants are presented in Appendix~\ref{sec:bandits_extra}.

For this problem, the optimal policy is to choose $A=C$, resulting in average reward of $0$. However, the reward signal $R$ is corrupted by the exogenous noise $\epsilon_r$, uncorrelated to the action. The higher the standard deviation, the more difficult proper credit assignment becomes, as high rewards are more likely due to a high value of $\epsilon_r$ than an appropriate choice of action. On the other hand, the feedback $F$ contains information about $C$, $A$ and $\epsilon_r$. If the agent can extract information $\Phi$ from $F$ in order to capture information about $\epsilon_r$ and use it to compute a hindsight value function, the effect of the perturbation $\epsilon_r$ may be removed from the advantage estimate, resulting in a significantly lower variance estimator. However, if the agent blindly uses $F$ to compute the hindsight value information, information about the action will `leak' into the hindsight value, leading to an advantage estimate of $0$ and no learning: intuitively, the agent will assume the outcome is entirely controlled by chance, and that all actions are equivalent, resulting in a form of learned helplessness.

\begin{figure}[h!]
    \centering
    \includegraphics[width=0.24\textwidth]{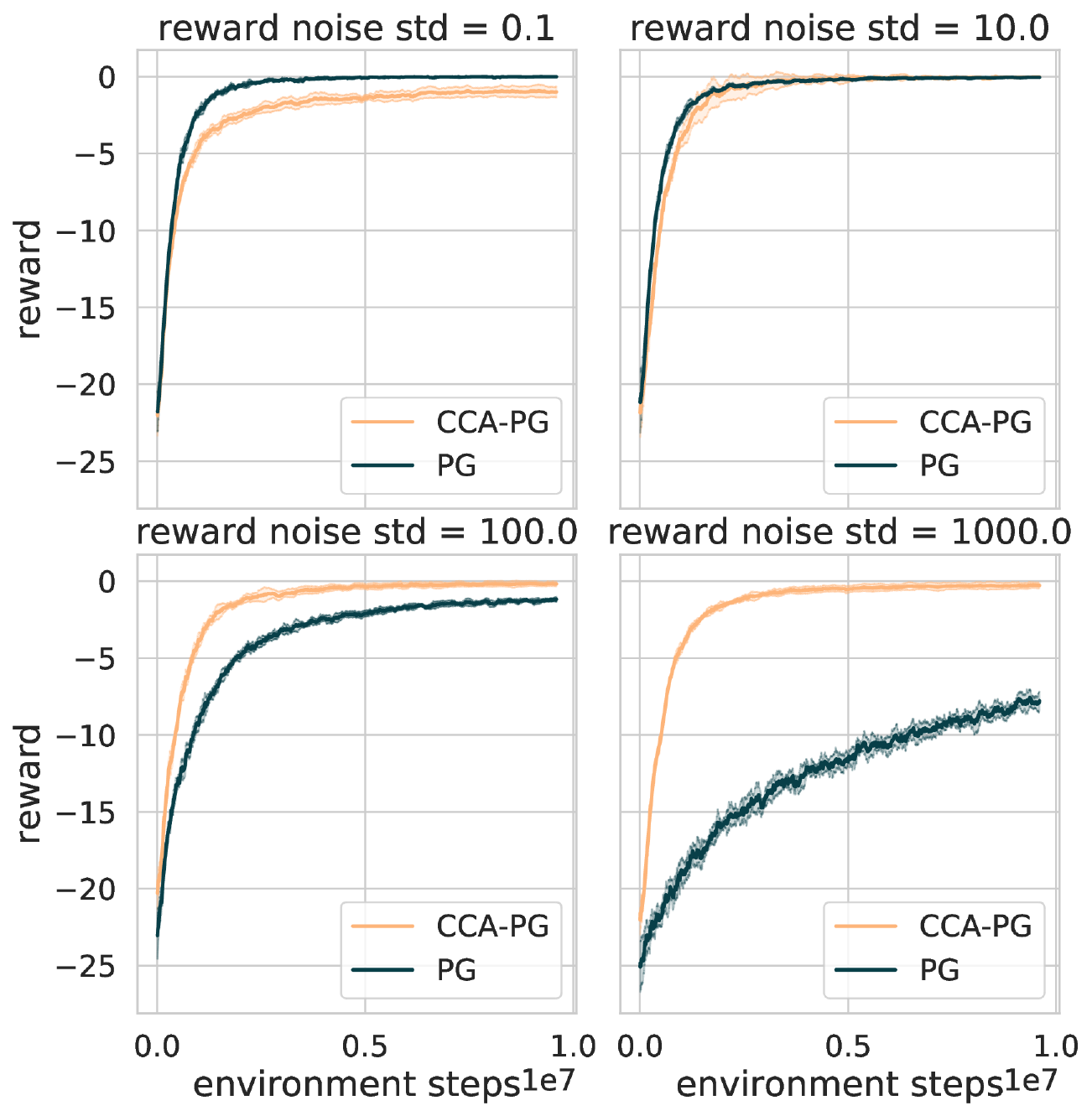}
    \includegraphics[width=0.235\textwidth]{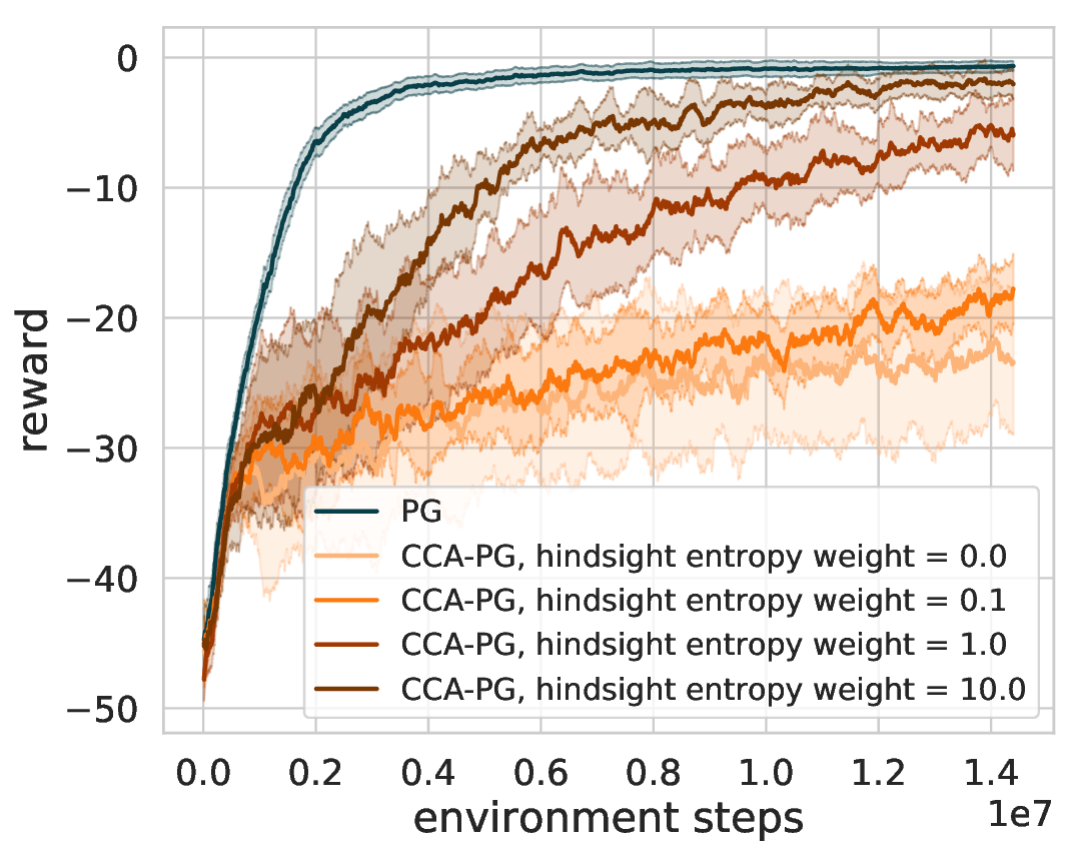}
    \caption{\textbf{Top:} Comparison of CCA-PG and PG in contextual bandits with feedback, for various levels of reward noise~$\sigma_r$. Results are averaged over 6 independent runs with standard deviation represented by a shaded area. \textbf{Bottom:} Performance of CCA-PG on the bandit task, for different values of $\lambda_{\text{IM}}$. Properly enforcing the independence constraint prevents the degradation of performance.}
    \label{fig.MIM_is_necessary}
    \label{fig.contextual_bandits}
\end{figure}

We investigate the proposed algorithm with $N=10$. As can be seen on Fig.~\ref{fig.contextual_bandits}, increasing the variance of the exogenous noise leads to dramatic decrease of performance for the vanilla PG estimator without the hindsight baseline; in contrast, the CCA-PG estimator is generally unaffected by the exogenous noise. For very low level of exogenous noise however, CCA-PG suffers from a decrease in performance. This is due to the agent computing a hindsight statistic $\Phi$ which is not perfectly independent from $A$, leading to bias in the policy gradient update. 
The agent attributes part of the reward to chance, despite the fact that in low-noise regime, the outcome is  entirely due to the agent's action.
To demonstrate this effect, and evaluate the importance of the independence constraint on performance, we run an ablation where we test lower values of the weight $\lambda_{\text{IM}}$ of the independence maximization loss (leading to a larger mutual information between $\Phi$ and $A$) and indeed observed that the performance is dramatically degraded, as seen in Fig.~\ref{fig.MIM_is_necessary}.

\subsection{Key-to-Door environments}
\paragraph{Task Description.}

We investigate new versions of the Key-To-Door family of environments, initially proposed by \citet{hung2019optimizing}, as a testbed of tasks where credit assignment is hard and is necessary for success. In this partially observable grid-world environment (cf.\ Fig.~\ref{fig.ktd} in the appendix), the agent has to pick up a key in the first room, for which it has \textit{no immediate reward}. In the second room, the agent can pick up 10 apples, that each give immediate rewards. In the final room, the agent may open a door (only if it is carrying a key), and receive a small reward for doing so.
In this task, a single action (i.e picking up the key) has a direct impact on the reward it receives in the final room, however this signal is hard to detect as the episode return is largely driven by its performance in the second room (i.e picking up apples). 

We now consider two instances of the Key-To-Door family that illustrate the difficulty of credit assignment in the presence of extrinsic variance.
In the Low-Variance-Key-To-Door environment, each apple is worth a reward of 1 and opening the final door also gets a reward of 1. Thus, an agent that solves the apple phase perfectly sees very little variance in its episode return and the learning signal for picking up the key and opening the door is relatively strong.

High-Variance-Key-To-Door keeps the overall structure of the Key-To-Door task. The door keeps giving a deterministic reward of 1 when the key was grabbed but now the reward for each apple is randomly sampled to be either 1 or 10, and fixed within the episode. In this setting, even an agent that is skilled at picking up apples sees a large variance in episode returns, and thus the learning signal for picking up the key and opening the door is comparatively weaker. Appendix~\ref{sec:ktd_env_details} has some additional discussion illustrating the difficulty of learning in such a setting.

\paragraph{Results}
We test CCA-PG on these environments, and compare it against Actor-Critic~\cite{williams1992simple}, as well as State-conditional HCA and Return-conditional HCA ~\citep{harutyunyan2019hindsight} as baselines. An analysis of the relation between HCA and CCA is described in Appendix~\ref{hca-cca-fc}. We test using both a backward-LSTM (referred to as CCA-PG RNN) or an attention model (referred to as CCA-PG Attn) for the hindsight function. Details for experimental setup are provided in Appendix~\ref{sec:ktd_architecture}.

\begin{figure}[h!]
    \centering
     \includegraphics[width=0.23\textwidth]{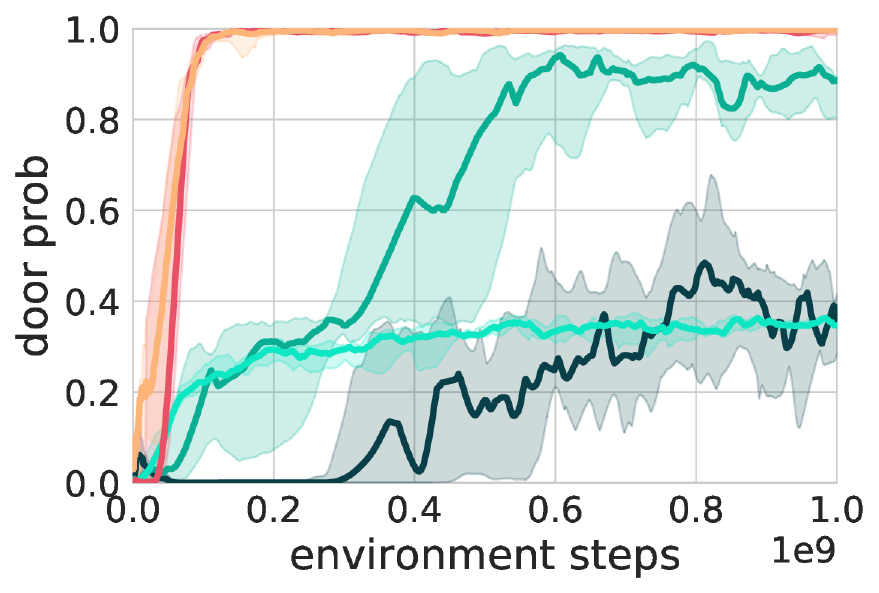}
     \label{fig.hvktd.door}
     \includegraphics[width=0.22\textwidth]{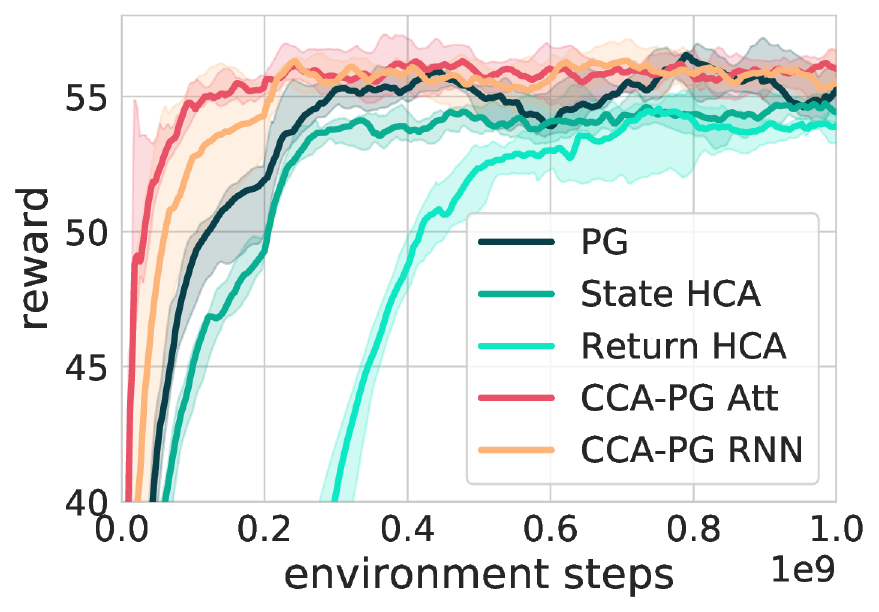}
     \label{fig.hvktd.reward}
     \includegraphics[width=0.23\textwidth]{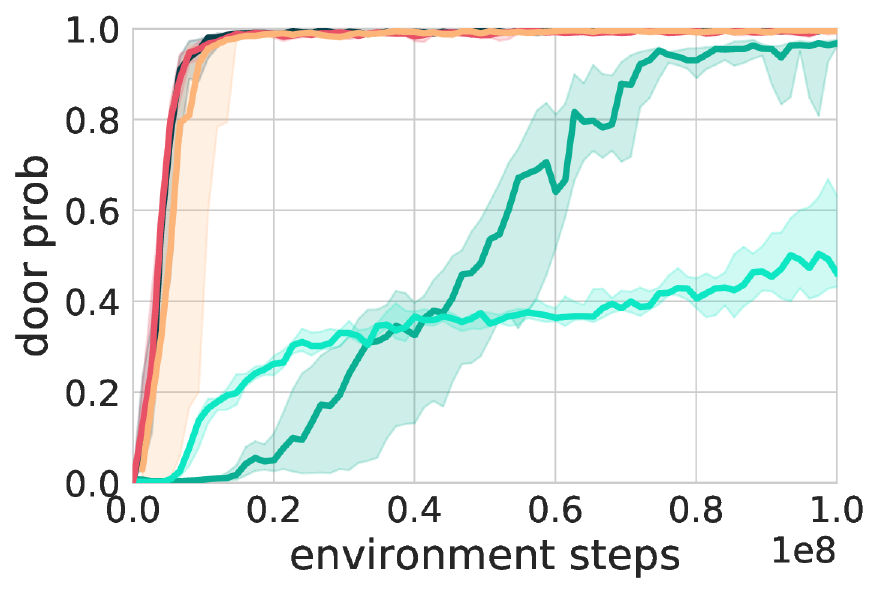}
     \label{fig.ktd.door}
     \includegraphics[width=0.22\textwidth]{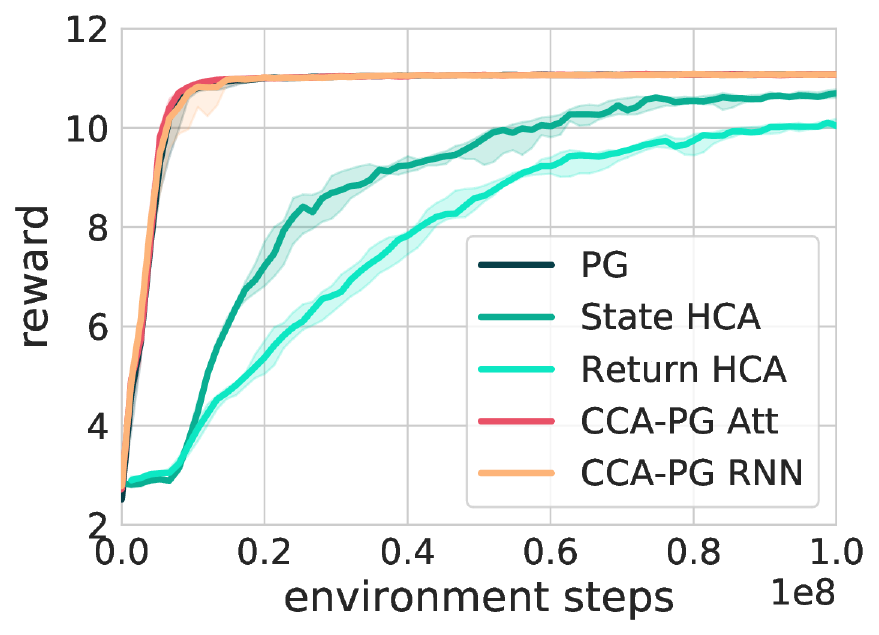}
     \label{fig.ktd.reward}
    \caption{
    Probability of opening the door and total reward obtained on the {\bf High-Variance-Key-To-Door} task (top two) and the {\bf Low-Variance-Key-To-Door} task (bottom two).}
    \label{fig.hvktd}
\end{figure}

We evaluate agents both on their ability to maximize total reward, as well as to solve the specific credit assignment problem of picking up the key and opening the door. 
Fig.~\ref{fig.hvktd} compares CCA-PG with the baselines on the High-Variance-Key-To-Door task. Both CCA-PG architectures outperform the baselines in terms of total reward, as well as probability of picking up the key and opening the door. 

This experiment highlights the capacity of CCA-PG to learn and incorporate trajectory-specific external factors into its baseline, resulting in lower variance estimators. Despite being a difficult task for credit assignment, CCA-PG is capable of solving it quickly and consistently. On the other hand, vanilla actor-critic is greatly impacted by this external variance, and needs around $3.10^9$ environment steps to have an 80\% probability of opening the door. CCA-PG also outperforms State- and Return- Conditional HCA, which do use hindsight information but in a more limited way than CCA-PG.

On the Low-Variance-Key-To-Door task (Fig.~\ref{fig.hvktd}), due to the lack of extrinsic variance, standard actor-critic is able to perfectly solve the environment. However, it is interesting to note that CCA-PG still matches this perfect performance. On the other hand, the other hindsight methods struggle with both door-opening and apple-gathering. This might be explained by the fact that both these techniques do not guarantee lower variance, and rely strongly on their learned hindsight classifiers for their policy gradient estimators, which can be harmful when these quantities are not perfectly learned. See Appendix~\ref{sec:ktd_additional_results} for additional experiments and ablations on these environments.

These experiments demonstrate that CCA-PG is capable of efficiently leveraging hindsight information to mitigate the challenge of external variance and learn strong policies that outperform baselines. At the same time, it suffers no drop in performance when used in cases where external variance is minimal.

\subsection{Task Interleaving}
\label{sec.interleaving.results}
\paragraph{Motivation.} 

In the real world, human activity can be seen as solving a large number of loosely related problems. At an abstract level, one could see this lifelong learning process as solving problems not in a sequential, but an interleaved fashion instead. These problems are not solved sequentially, as one may temporarily engage with a problem and only continue engaging with it or receive feedback from its earlier actions significantly later. The structure of this interleaving will also typically vary over time.

Despite this very complex structure and receiving high variance rewards from the future, humans are able to quickly make sense of these varying episodes and correctly credit their actions. This paradigm is quite different from the environments that are typically investigated by the reinforcement learning community. Indeed, focus is mostly put on agents trained on a single task, with an outcome mostly caused by the agent's actions, where long term credit assignment is not required and where every episode will be structurally the same. 

\begin{figure*}[ht!]
\centering
\includegraphics[width=0.8\textwidth]{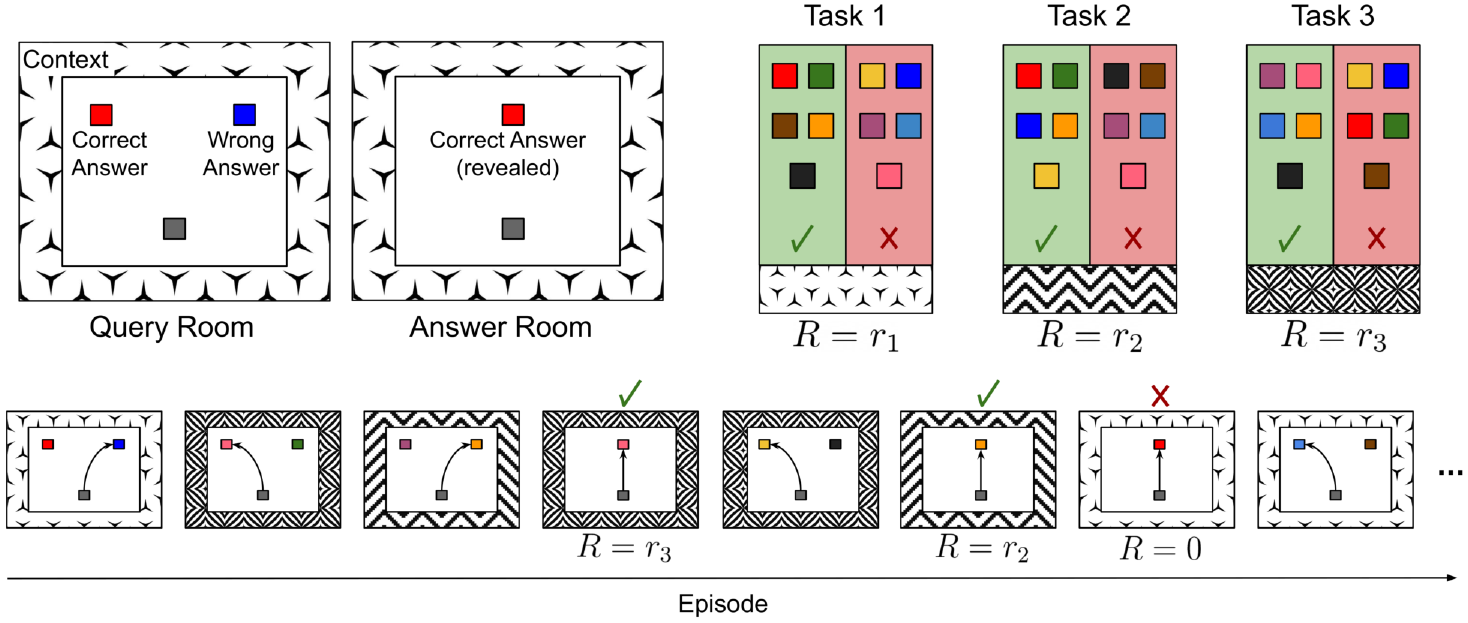}
\caption{
{\bf Task Interleaving Description.} \textbf{Top left:} Delayed feedback contextual bandit problem. Given a context shown as a surrounding visual pattern, the agent has to decide to pick up one of the two colored squares where only one will be rewarding. The agent is later teleported to the second room where it is provided with the reward associated with its previous choice and a visual cue about which colored square it should have picked up. \textbf{Top right:} Different tasks with each a different color mapping, visual context and associated reward. \textbf{Bottom:} Example of a generated episode, composed of randomly sampled tasks and color pairs.}
\label{fig.interleaving}
\end{figure*}

To better understand the effects of interleaving on agent learning, we introduce a new class of environments capturing the structural properties mentioned above. In contrast to most work on multi-task learning, we do not assume a clear delineation between subtasks, nor focus on skill retention. The agent will encounter multiple tasks in a single episode in an interleaved fashion (switching between tasks will occur before a task gets completed), and will have to detect the implicitly boundaries between them.
\paragraph{Task Description.} 
This task consists of pairs of query-answer rooms with different visual contexts that each indicates a different subtask.
In the query room, the agent gets to pick between two colored boxes (out of 10 possible colors). Later, in the answer room, the agents gets to observe which of the two boxes was rewarding in the first room, and receives a reward if it picked the correct box (there is always exactly one rewarding color in the query room). The mapping of colors to whether it is rewarding or not is specific to each subtask and fixed across training. Each subtask would be relatively easy to solve if encountered in an isolated fashion. However, each episode is composed of \textit{randomly sampled subtasks} and color pairs within those subtasks. Furthermore, query rooms and answer rooms of the sampled subtasks are presented in a random (interleaved) order which differs from one episode to another. Each episode are 140 steps long and it takes at least 9 steps for the agent to reach one colored square from its initial position. A visual example of what an episode looks like can be seen in Fig.~\ref{fig.interleaving}. 

There are six tasks, each classified as `easy' or `hard'; easy tasks have high reward signals (i.e. easier for agents to pick up on), while hard tasks have low rewards. In the 2 tasks setup (resp. 4 tasks and 6 tasks), there is one (resp. two and two) `easy’ and one (resp. two and four) `hard’ task. More details about the experimental setup can be found in~\ref{sec:interleaving}.

In addition to the total reward, we record the probability of picking up the correct square for the easy and hard tasks separately. Performance in the hard tasks will indicate ability to do fine-grained credit assignment. 
\begin{figure}[h!]
    \includegraphics[width=0.22\textwidth]{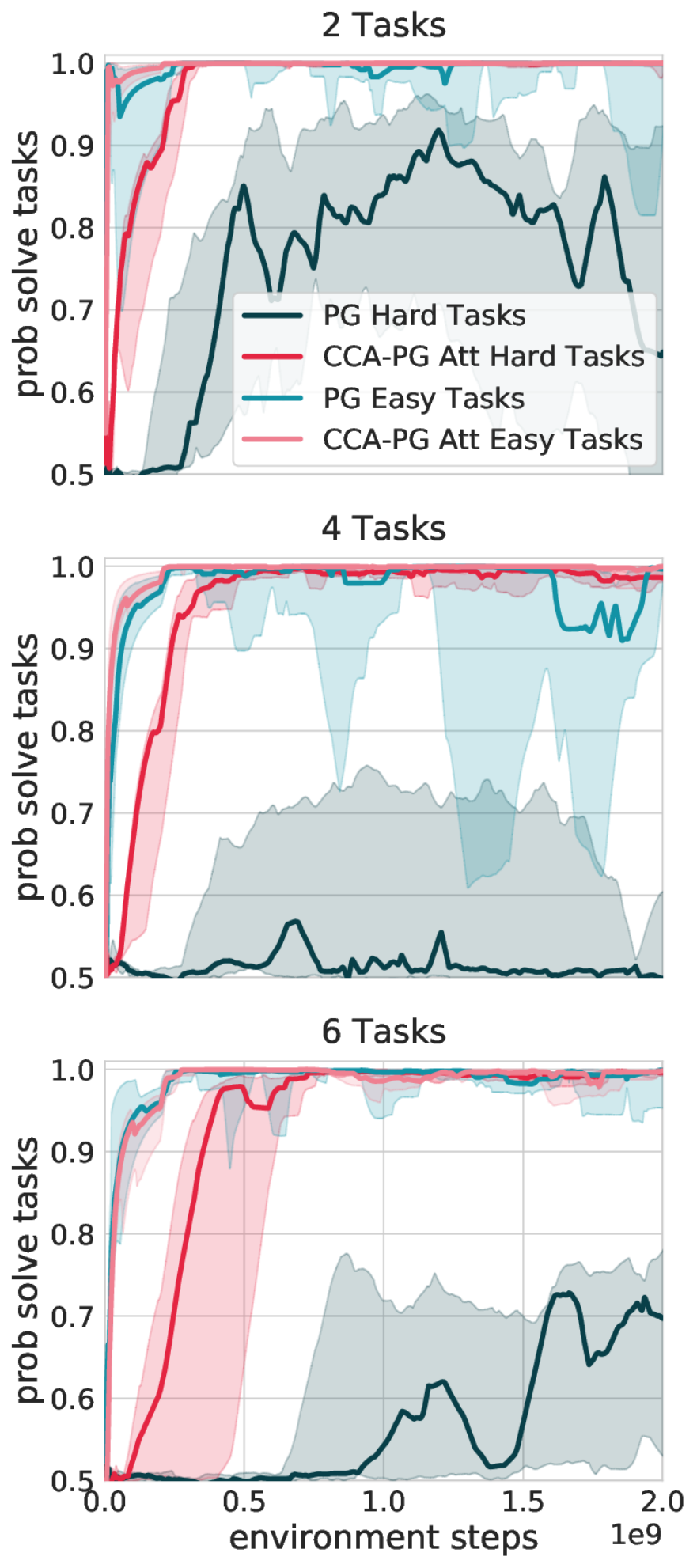}
    \label{fig.interleaving_prob}
    \includegraphics[width=0.22\textwidth]{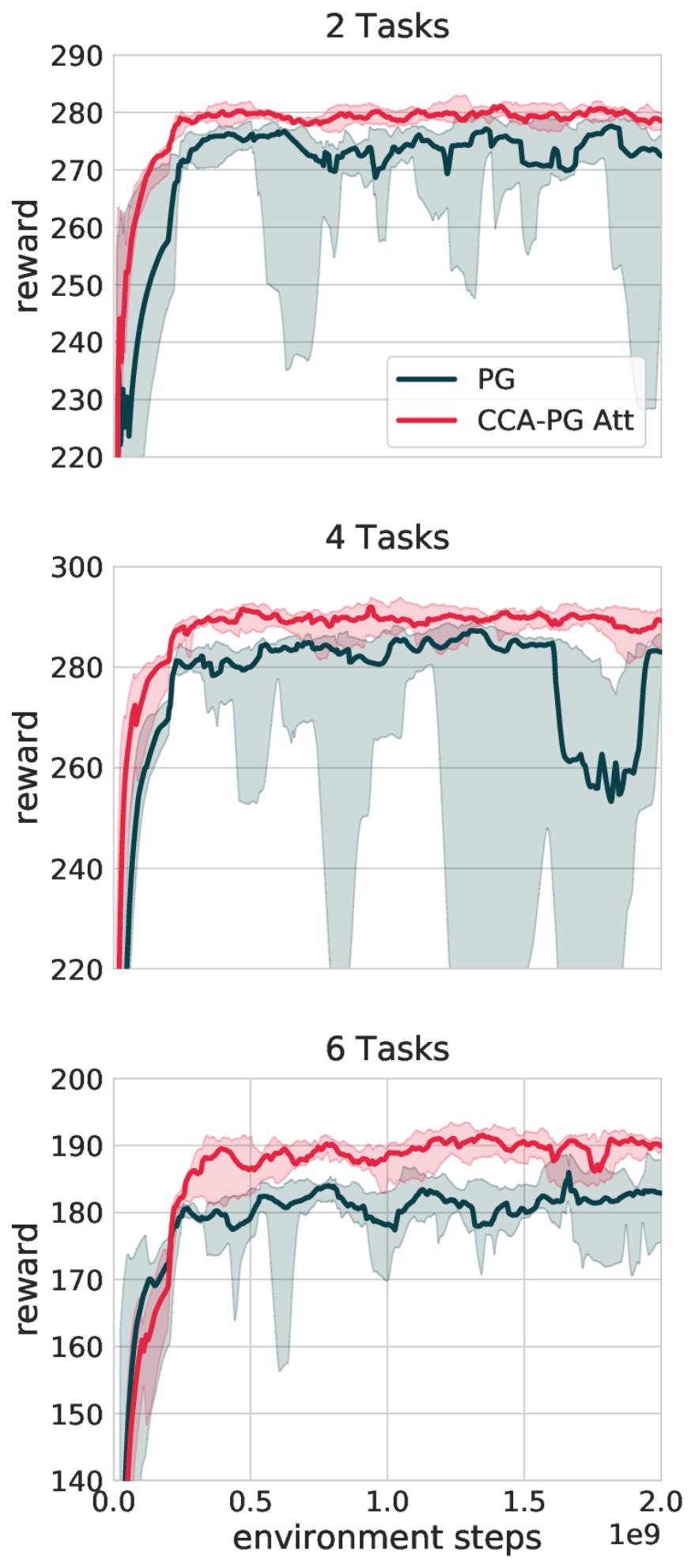}
    \label{fig.interleaving_reward}
    \caption{
        Probability of solving `easy’ and `hard’ tasks (left) and total reward (right) obtained for the {\bf Multi Task Interleaving.} Left plots: Median over 10 seeds after doing a mean over the performances in `easy’ or `hard’ tasks.}
    \label{fig.interleaving_results}
\end{figure}
\paragraph{Results.}  While CCA-PG is able to perfectly solve both the `easy’ and `hard’ tasks in the three setups in less than $5.10^8$ environment steps~(Fig.~\ref{fig.interleaving_results}), actor-critic is only capable to solve the ’easy’ tasks for which the associated rewards are large. Even after $2.10^9$ environment steps, actor-critic is still greatly impacted by the variance and remains incapable of solving `hard’ tasks in any of the three settings. CCA-PG also outperforms actor-critic in terms of the total reward obtained in each setting. State-conditional and Return-conditional HCA were also evaluated on this task but results are not reported as almost no learning was taking place on the ’hard’ tasks. More results along with an ablation study can be found in Appendix~\ref{sec:interleaving}. 

Through efficient use of hindsight, CCA-PG is able to take into account trajectory-specific factors such as the kinds of rooms encountered in the episode and their associated rewards.

In the case of the Multi-Task Interleaving environment, an informative hindsight function would capture the reward for different contexts and exposes as $\Phi_t$ all rewards obtained in the episode except those associated with the current context. This experiment again highlights the capacity of CCA-PG to solve hard credit assignment problems in a context where the return is affected by multiple distractors, while PG remains highly sensitive to them.

\section{Related work}\label{sec:lit}
This paper builds on work from \citet{buesing2018woulda} which shows how causal models and real data can be combined to generate counterfactual trajectories and perform off-policy evaluation for RL. Their results however require an explicit model of the environment. In contrast, our work proposes a model-free approach, and focuses on policy improvement. 
\cite{heess2015learning} also presents model-based counterfactuals policy gradient estimates, however they focus on continuous actions and differentiable environments.
\citet{oberst2019counterfactual} also investigates counterfactuals in reinforcement learning, point out the issue of non-identifiability of the correct SCM, and suggest a sufficient condition for identifiability; we discuss this issue in appendix~\ref{sec:simple_ex_causality}. 
Closely related to our work is Hindsight Credit Assignment, a concurrent approach  from \citet{harutyunyan2019hindsight}. In this paper, the authors also investigate value functions and critics that depend on future information. However, the information the estimators depend on is fixed (future state or return) instead of being an arbitrary functions of the trajectory. Our FC estimators generalizes both the HCA and CCA estimators while CCA further characterizes which statistics of the future provide a useful estimator. Relations between HCA, CCA and FC are discussed in appendix~\ref{hca-cca-fc}.
The HCA approach is further extended by \citet{young2019variance}, and \citet{zhang2019IAE} who minimize a surrogate for the variance of the estimator, but that surrogate cannot be guaranteed to actually lower the variance. Similarly to state-HCA, they treat each reward separately instead of taking a trajectory-centric view as CCA. \citep{alipov2021towards} proposes practical modifications to the HCA algorithms to improve its experimental results on Atari.
\citet{guez2019himo} also investigates future-conditional value functions. Similarly to us, they learn statistics of the future $\Phi$ from which returns can be accurately predicted, and show that doing so leads to learning better representations (but use regular policy gradient estimators otherwise). Instead of enforcing a information-theoretic constraint, they bottleneck information through the size of the encoding $\Phi$. In domain adaptation \citep{ganin2016domain,tzeng2017adversarial}, robustness to the training domain can be achieved by constraining the agent representation not to be able to discriminate between source and target domains, a  mechanism similar to the one constraining hindsight features not being able to discriminate the agent's actions. Also closely related to our paper, \citet{bica2020estimating} also leverages a similar mechanism to compute counterfactuals, for a different purpose than ours (computing treatment effects vs. policy improvement operators).

Both \citet{andrychowicz2017hindsight} and \citet{rauber2017hindsight} leverage the idea of using hindsight information to learn goal-conditioned policies. \citet{hung2019optimizing} leverages attention-based systems and episode memory to perform long term credit assignment; however, their estimator will in general be biased. \citet{ferret2019credit} looks at the question of transfer learning in RL and leverages transformers to derive a heuristic to perform reward shaping. \citet{arjona2019rudder} also addresses the problem of long-term credit assignment by redistributing delayed rewards earlier in the episode but their approach still fundamentally uses time as a proxy for credit. 

As mentioned in \ref{subsec:learnstats}, previous research also leverages the fact that baselines can include information unknown to the agent at time $t$ and potentially revealed in hindsight, but which is not affected by action $A_t$ \citep{wu2018variance, mao2018variance, foerster2018counterfactual, andrychowicz2018learning, vinyals2019grandmaster}. However, those methods require access to expert information unlike our approach. 

Furthermore, leveraging generative models to produce information that relies on hindsight information without being affected by the action $A_t$, also suggested by \citep{nota2021posterior}, is an exciting new route which could be the source of a wide variety of new methods. Finally, \citep{venuto2021policy} proposes a method similar to generative CCA / posterior value functions (they use an information bottleneck between prior and posterior, which can be interpreted as a variational autoencoder model of the return), combined with additional techniques to improve performance.
\section{Conclusion}
In this paper we have considered the problem of credit assignment in RL. Building on insights from causality theory and structural causal models, we have investigated the concept of future-conditional value functions. Contrary to common practice these allow baselines and critics to condition on future events thus separating the influence of an agent's actions on future rewards from the effects of other random events thus reducing the variance of policy gradient estimators.  A key difficulty lies in the fact that unbiasedness relies on accurate estimation and minimization of mutual information. Learning inaccurate hindsight classifiers will result in miscalibrated estimation of luck, leading to bias in learning.
Future research will investigate how to scale these algorithms to more complex environments, and the benefits of the more general FC-PG and all-actions estimators.

\clearpage
\bibliography{bibliography}

\begin{thebibliography}{47}
\providecommand{\natexlab}[1]{#1}
\providecommand{\url}[1]{\texttt{#1}}
\expandafter\ifx\csname urlstyle\endcsname\relax
  \providecommand{\doi}[1]{doi: #1}\else
  \providecommand{\doi}{doi: \begingroup \urlstyle{rm}\Url}\fi

\bibitem[Alipov et~al.(2021)Alipov, Simmons-Edler, Putintsev, Kalinin, and
  Vetrov]{alipov2021towards}
Alipov, V., Simmons-Edler, R., Putintsev, N., Kalinin, P., and Vetrov, D.
\newblock Towards practical credit assignment for deep reinforcement learning.
\newblock \emph{arXiv preprint arXiv:2106.04499}, 2021.

\bibitem[Andrychowicz et~al.(2017)Andrychowicz, Wolski, Ray, Schneider, Fong,
  Welinder, McGrew, Tobin, Abbeel, and Zaremba]{andrychowicz2017hindsight}
Andrychowicz, M., Wolski, F., Ray, A., Schneider, J., Fong, R., Welinder, P.,
  McGrew, B., Tobin, J., Abbeel, O.~P., and Zaremba, W.
\newblock Hindsight experience replay.
\newblock In \emph{Advances in neural information processing systems}, pp.\
  5048--5058, 2017.

\bibitem[Andrychowicz et~al.(2020)Andrychowicz, Baker, Chociej, Jozefowicz,
  McGrew, Pachocki, Petron, Plappert, Powell, Ray,
  et~al.]{andrychowicz2018learning}
Andrychowicz, O.~M., Baker, B., Chociej, M., Jozefowicz, R., McGrew, B.,
  Pachocki, J., Petron, A., Plappert, M., Powell, G., Ray, A., et~al.
\newblock Learning dexterous in-hand manipulation.
\newblock \emph{The International Journal of Robotics Research}, 39\penalty0
  (1):\penalty0 3--20, 2020.

\bibitem[Arjona-Medina et~al.(2019)Arjona-Medina, Gillhofer, Widrich,
  Unterthiner, Brandstetter, and Hochreiter]{arjona2019rudder}
Arjona-Medina, J.~A., Gillhofer, M., Widrich, M., Unterthiner, T.,
  Brandstetter, J., and Hochreiter, S.
\newblock Rudder: Return decomposition for delayed rewards.
\newblock In \emph{Advances in Neural Information Processing Systems}, pp.\
  13544--13555, 2019.

\bibitem[Bellemare et~al.(2017)Bellemare, Dabney, and
  Munos]{bellemare2017distributional}
Bellemare, M.~G., Dabney, W., and Munos, R.
\newblock A distributional perspective on reinforcement learning.
\newblock In \emph{Proceedings of the 34th International Conference on Machine
  Learning-Volume 70}, pp.\  449--458. JMLR. org, 2017.

\bibitem[Bica et~al.(2020)Bica, Alaa, Jordon, and van~der
  Schaar]{bica2020estimating}
Bica, I., Alaa, A.~M., Jordon, J., and van~der Schaar, M.
\newblock Estimating counterfactual treatment outcomes over time through
  adversarially balanced representations.
\newblock \emph{arXiv preprint arXiv:2002.04083}, 2020.

\bibitem[Buesing et~al.(2016)Buesing, Weber, and
  Mohamed]{buesing2016stochastic}
Buesing, L., Weber, T., and Mohamed, S.
\newblock Stochastic gradient estimation with finite differences.
\newblock In \emph{NIPS2016 Workshop on Advances in Approximate Inference},
  2016.

\bibitem[Buesing et~al.(2019)Buesing, Weber, Zwols, Racaniere, Guez, Lespiau,
  and Heess]{buesing2018woulda}
Buesing, L., Weber, T., Zwols, Y., Racaniere, S., Guez, A., Lespiau, J.-B., and
  Heess, N.
\newblock Woulda, coulda, shoulda: Counterfactually-guided policy search.
\newblock \emph{2019 International Conference for Learning Representations
  (ICLR)}, 2019.

\bibitem[Chung et~al.(2015)Chung, Gulcehre, Cho, and Bengio]{chung2015gated}
Chung, J., Gulcehre, C., Cho, K., and Bengio, Y.
\newblock Gated feedback recurrent neural networks.
\newblock In \emph{International conference on machine learning}, pp.\
  2067--2075, 2015.

\bibitem[Ferret et~al.(2019)Ferret, Marinier, Geist, and
  Pietquin]{ferret2019credit}
Ferret, J., Marinier, R., Geist, M., and Pietquin, O.
\newblock Credit assignment as a proxy for transfer in reinforcement learning.
\newblock \emph{arXiv preprint arXiv:1907.08027}, 2019.

\bibitem[Foerster et~al.(2018)Foerster, Farquhar, Afouras, Nardelli, and
  Whiteson]{foerster2018counterfactual}
Foerster, J.~N., Farquhar, G., Afouras, T., Nardelli, N., and Whiteson, S.
\newblock Counterfactual multi-agent policy gradients.
\newblock In \emph{Thirty-Second AAAI Conference on Artificial Intelligence},
  2018.

\bibitem[Ganin et~al.(2016)Ganin, Ustinova, Ajakan, Germain, Larochelle,
  Laviolette, Marchand, and Lempitsky]{ganin2016domain}
Ganin, Y., Ustinova, E., Ajakan, H., Germain, P., Larochelle, H., Laviolette,
  F., Marchand, M., and Lempitsky, V.
\newblock Domain-adversarial training of neural networks.
\newblock \emph{The Journal of Machine Learning Research}, 17\penalty0
  (1):\penalty0 2096--2030, 2016.

\bibitem[Glasserman \& Yao(1992)Glasserman and Yao]{glasserman1992some}
Glasserman, P. and Yao, D.~D.
\newblock Some guidelines and guarantees for common random numbers.
\newblock \emph{Management Science}, 38\penalty0 (6):\penalty0 884--908, 1992.

\bibitem[Goyal et~al.(2019)Goyal, Lamb, Hoffmann, Sodhani, Levine, Bengio, and
  Sch{\"o}lkopf]{goyal2019recurrent}
Goyal, A., Lamb, A., Hoffmann, J., Sodhani, S., Levine, S., Bengio, Y., and
  Sch{\"o}lkopf, B.
\newblock Recurrent independent mechanisms.
\newblock \emph{arXiv preprint arXiv:1909.10893}, 2019.

\bibitem[Guez et~al.(2019)Guez, Viola, Weber, Buesing, Kapturowski, Precup,
  Silver, and Heess]{guez2019himo}
Guez, A., Viola, F., Weber, T., Buesing, L., Kapturowski, S., Precup, D.,
  Silver, D., and Heess, N.
\newblock Value-driven hindsight modelling.
\newblock \emph{https://openreview.net/forum?id=rJxBa1HFvS}, 2019.

\bibitem[Ha \& Schmidhuber(2018)Ha and Schmidhuber]{ha2018world}
Ha, D. and Schmidhuber, J.
\newblock World models.
\newblock \emph{arXiv preprint arXiv:1803.10122}, 2018.

\bibitem[Hamrick(2019)]{hamrick2019analogues}
Hamrick, J.~B.
\newblock Analogues of mental simulation and imagination in deep learning.
\newblock \emph{Current Opinion in Behavioral Sciences}, 29:\penalty0 8--16,
  2019.

\bibitem[Harutyunyan et~al.(2019)Harutyunyan, Dabney, Mesnard, Azar, Piot,
  Heess, van Hasselt, Wayne, Singh, Precup, et~al.]{harutyunyan2019hindsight}
Harutyunyan, A., Dabney, W., Mesnard, T., Azar, M.~G., Piot, B., Heess, N., van
  Hasselt, H.~P., Wayne, G., Singh, S., Precup, D., et~al.
\newblock Hindsight credit assignment.
\newblock In \emph{Advances in Neural Information Processing Systems}, pp.\
  12467--12476, 2019.

\bibitem[Heess et~al.(2015)Heess, Wayne, Silver, Lillicrap, Erez, and
  Tassa]{heess2015learning}
Heess, N., Wayne, G., Silver, D., Lillicrap, T., Erez, T., and Tassa, Y.
\newblock Learning continuous control policies by stochastic value gradients.
\newblock In \emph{Advances in Neural Information Processing Systems}, pp.\
  2944--2952, 2015.

\bibitem[Hinton et~al.(2012)Hinton, Srivastava, and Swersky]{hinton2012neural}
Hinton, G., Srivastava, N., and Swersky, K.
\newblock Neural networks for machine learning lecture 6a overview of
  mini-batch gradient descent.
\newblock \emph{Cited on}, 14\penalty0 (8), 2012.

\bibitem[Hung et~al.(2019)Hung, Lillicrap, Abramson, Wu, Mirza, Carnevale,
  Ahuja, and Wayne]{hung2019optimizing}
Hung, C.-C., Lillicrap, T., Abramson, J., Wu, Y., Mirza, M., Carnevale, F.,
  Ahuja, A., and Wayne, G.
\newblock Optimizing agent behavior over long time scales by transporting
  value.
\newblock \emph{Nature communications}, 10\penalty0 (1):\penalty0 1--12, 2019.

\bibitem[Kaiser et~al.(2019)Kaiser, Babaeizadeh, Milos, Osinski, Campbell,
  Czechowski, Erhan, Finn, Kozakowski, Levine, et~al.]{kaiser2019model}
Kaiser, L., Babaeizadeh, M., Milos, P., Osinski, B., Campbell, R.~H.,
  Czechowski, K., Erhan, D., Finn, C., Kozakowski, P., Levine, S., et~al.
\newblock Model-based reinforcement learning for atari.
\newblock \emph{arXiv preprint arXiv:1903.00374}, 2019.

\bibitem[Kingma \& Ba(2014)Kingma and Ba]{kingma2014adam}
Kingma, D.~P. and Ba, J.
\newblock Adam: A method for stochastic optimization.
\newblock \emph{arXiv preprint arXiv:1412.6980}, 2014.

\bibitem[Kingma \& Welling(2013)Kingma and Welling]{kingma2013auto}
Kingma, D.~P. and Welling, M.
\newblock Auto-encoding variational bayes.
\newblock \emph{arXiv preprint arXiv:1312.6114}, 2013.

\bibitem[Mao et~al.(2018)Mao, Venkatakrishnan, Schwarzkopf, and
  Alizadeh]{mao2018variance}
Mao, H., Venkatakrishnan, S.~B., Schwarzkopf, M., and Alizadeh, M.
\newblock Variance reduction for reinforcement learning in input-driven
  environments.
\newblock In \emph{International Conference on Learning Representations}, 2018.

\bibitem[Minsky(1961)]{minsky1961steps}
Minsky, M.
\newblock Steps toward artificial intelligence.
\newblock \emph{Proceedings of the IRE}, 49\penalty0 (1):\penalty0 8--30, 1961.

\bibitem[Nota et~al.(2021)Nota, Thomas, and Da~Silva]{nota2021posterior}
Nota, C., Thomas, P., and Da~Silva, B.~C.
\newblock Posterior value functions: Hindsight baselines for policy gradient
  methods.
\newblock In \emph{International Conference on Machine Learning}, pp.\
  8238--8247. PMLR, 2021.

\bibitem[Oberst \& Sontag(2019)Oberst and Sontag]{oberst2019counterfactual}
Oberst, M. and Sontag, D.
\newblock Counterfactual off-policy evaluation with gumbel-max structural
  causal models.
\newblock \emph{arXiv preprint arXiv:1905.05824}, 2019.

\bibitem[Papamakarios et~al.(2019)Papamakarios, Nalisnick, Rezende, Mohamed,
  and Lakshminarayanan]{papamakarios2019normalizing}
Papamakarios, G., Nalisnick, E., Rezende, D.~J., Mohamed, S., and
  Lakshminarayanan, B.
\newblock Normalizing flows for probabilistic modeling and inference.
\newblock \emph{arXiv preprint arXiv:1912.02762}, 2019.

\bibitem[Parisotto et~al.(2019)Parisotto, Song, Rae, Pascanu, Gulcehre,
  Jayakumar, Jaderberg, Kaufman, Clark, Noury,
  et~al.]{parisotto2019stabilizing}
Parisotto, E., Song, H.~F., Rae, J.~W., Pascanu, R., Gulcehre, C., Jayakumar,
  S.~M., Jaderberg, M., Kaufman, R.~L., Clark, A., Noury, S., et~al.
\newblock Stabilizing transformers for reinforcement learning.
\newblock \emph{arXiv preprint arXiv:1910.06764}, 2019.

\bibitem[Pearl(2009{\natexlab{a}})]{pearl2009causality}
Pearl, J.
\newblock \emph{Causality}.
\newblock Cambridge university press, 2009{\natexlab{a}}.

\bibitem[Pearl(2009{\natexlab{b}})]{pearlbook}
Pearl, J.
\newblock Causality: Models, reasoning, and inference.
\newblock 2009{\natexlab{b}}.

\bibitem[Rauber et~al.(2017)Rauber, Ummadisingu, Mutz, and
  Schmidhuber]{rauber2017hindsight}
Rauber, P., Ummadisingu, A., Mutz, F., and Schmidhuber, J.
\newblock Hindsight policy gradients.
\newblock \emph{arXiv preprint arXiv:1711.06006}, 2017.

\bibitem[Rezende \& Viola(2018)Rezende and Viola]{rezende2018taming}
Rezende, D.~J. and Viola, F.
\newblock Taming {VAEs}.
\newblock \emph{arXiv preprint arXiv:1810.00597}, 2018.

\bibitem[Rezende et~al.(2014)Rezende, Mohamed, and
  Wierstra]{rezende2014stochastic}
Rezende, D.~J., Mohamed, S., and Wierstra, D.
\newblock Stochastic backpropagation and approximate inference in deep
  generative models.
\newblock In \emph{International conference on machine learning}, pp.\
  1278--1286. PMLR, 2014.

\bibitem[Schrittwieser et~al.(2019)Schrittwieser, Antonoglou, Hubert, Simonyan,
  Sifre, Schmitt, Guez, Lockhart, Hassabis, Graepel,
  et~al.]{schrittwieser2019mastering}
Schrittwieser, J., Antonoglou, I., Hubert, T., Simonyan, K., Sifre, L.,
  Schmitt, S., Guez, A., Lockhart, E., Hassabis, D., Graepel, T., et~al.
\newblock Mastering atari, go, chess and shogi by planning with a learned
  model.
\newblock \emph{arXiv preprint arXiv:1911.08265}, 2019.

\bibitem[Sutton et~al.(2000)Sutton, McAllester, Singh, and
  Mansour]{sutton2000policy}
Sutton, R.~S., McAllester, D.~A., Singh, S.~P., and Mansour, Y.
\newblock Policy gradient methods for reinforcement learning with function
  approximation.
\newblock In \emph{Advances in neural information processing systems}, pp.\
  1057--1063, 2000.

\bibitem[Tzeng et~al.(2017)Tzeng, Hoffman, Saenko, and
  Darrell]{tzeng2017adversarial}
Tzeng, E., Hoffman, J., Saenko, K., and Darrell, T.
\newblock Adversarial discriminative domain adaptation.
\newblock In \emph{Proceedings of the IEEE Conference on Computer Vision and
  Pattern Recognition}, pp.\  7167--7176, 2017.

\bibitem[Vaswani et~al.(2017)Vaswani, Shazeer, Parmar, Uszkoreit, Jones, Gomez,
  Kaiser, and Polosukhin]{vaswani2017attention}
Vaswani, A., Shazeer, N., Parmar, N., Uszkoreit, J., Jones, L., Gomez, A.~N.,
  Kaiser, {\L}., and Polosukhin, I.
\newblock Attention is all you need.
\newblock In \emph{Advances in neural information processing systems}, pp.\
  5998--6008, 2017.

\bibitem[Venuto et~al.(2021)Venuto, Lau, Precup, and Nachum]{venuto2021policy}
Venuto, D., Lau, E., Precup, D., and Nachum, O.
\newblock Policy gradients incorporating the future.
\newblock \emph{arXiv preprint arXiv:2108.02096}, 2021.

\bibitem[Vinyals et~al.(2019)Vinyals, Babuschkin, Czarnecki, Mathieu, Dudzik,
  Chung, Choi, Powell, Ewalds, Georgiev, et~al.]{vinyals2019grandmaster}
Vinyals, O., Babuschkin, I., Czarnecki, W.~M., Mathieu, M., Dudzik, A., Chung,
  J., Choi, D.~H., Powell, R., Ewalds, T., Georgiev, P., et~al.
\newblock Grandmaster level in starcraft {II} using multi-agent reinforcement
  learning.
\newblock \emph{Nature}, 575\penalty0 (7782):\penalty0 350--354, 2019.

\bibitem[Weber et~al.(2019)Weber, Heess, Buesing, and Silver]{weber2019credit}
Weber, T., Heess, N., Buesing, L., and Silver, D.
\newblock Credit assignment techniques in stochastic computation graphs.
\newblock In \emph{The 22nd International Conference on Artificial Intelligence
  and Statistics}, pp.\  2650--2660, 2019.

\bibitem[Williams(1992)]{williams1992simple}
Williams, R.~J.
\newblock Simple statistical gradient-following algorithms for connectionist
  reinforcement learning.
\newblock \emph{Machine learning}, 8\penalty0 (3-4):\penalty0 229--256, 1992.

\bibitem[Wu et~al.(2018)Wu, Rajeswaran, Duan, Kumar, Bayen, Kakade, Mordatch,
  and Abbeel]{wu2018variance}
Wu, C., Rajeswaran, A., Duan, Y., Kumar, V., Bayen, A.~M., Kakade, S.,
  Mordatch, I., and Abbeel, P.
\newblock Variance reduction for policy gradient with action-dependent
  factorized baselines.
\newblock \emph{2018 International Conference for Learning Representations
  (ICLR)}, 2018.

\bibitem[Young(2019)]{young2019variance}
Young, K.
\newblock Variance reduced advantage estimation with $\delta$-hindsight credit
  assignment.
\newblock \emph{arXiv preprint arXiv:1911.08362}, 2019.

\bibitem[Zhang(2020)]{zhang2020designing}
Zhang, J.
\newblock Designing optimal dynamic treatment regimes: A causal reinforcement
  learning approach.
\newblock In \emph{International Conference on Machine Learning}, pp.\
  11012--11022. PMLR, 2020.

\bibitem[Zhang et~al.(2019)Zhang, Zhao, Liu, Bian, Huang, Qin, and
  Tie-Yan]{zhang2019IAE}
Zhang, P., Zhao, L., Liu, G., Bian, J., Huang, M., Qin, T., and Tie-Yan, L.
\newblock Independence-aware advantage estimation.
\newblock \emph{https://openreview.net/forum?id=B1eP504YDr}, 2019.

\end{thebibliography}
\bibliographystyle{icml2021}
\clearpage
\FloatBarrier
\onecolumn
\appendix
\begin{center}
{\LARGE\bf Appendix\par}
\end{center}
\vskip 0.1in
\hrule 
\section{Algorithmic and implementation details}
\label{sec:implementation-details}
\subsection{Constrained optimization}

Corollary~\ref{prop:cfac} requires an independence assumption between $A_t$ and $\Phi_t$, conditional on $X_t$. We can therefore cast the problem of learning $\Phi_t$ as a constrained optimization problem, where the loss $\calL_{\text{hs}}$ measures how predictive of the return $\Phi_t$ is, and the constraint enforces a maximum (tolerance) value of  $\betaim$ for the independence maximization loss $\calL_{\text{IM}}$ (exact independence is obtained by $\betaim=0$, but this is hard to achieve exactly in practice).

The resulting optimization problem for finding an appropriate counterfactual baseline is given by:
\begin{align}
    &\min_{\theta} \E\left[\calL_{\text{hs}}\right]\label{eq:CCA}
   \:\:\:\:\: \text{subject to:}\:\: \forall t\:\: \calL_\text{IM}(X_t) \leq \betaim
\end{align}

The resulting hindsight baseline can then be used in the policy gradient estimate. There are two problems remaining to solve. First, the form of the (IM) loss used requires knowing the exact hindsight probability $\mathbb{P}(A_t|X_t,\Phi_t)$. As explained in the main text, we replace it by the classifier $h$, tracking the optimal classifier by stochastically minimizing the supervised loss (optimizing it only with respect to the parameters of the hindsight classifier).  Second, we relax the constraint using a Lagrangian method (the Lagrangian parameter can either be set as a hyperparameter, or optimized using an algorithm like GECO \citep{rezende2018taming}). 

\subsection{Parameter updates}

The corresponding parameter updates are as follows:

\noindent For each trajectory $(X_t, A_t, R_t)_{t\geq 0}$, compute the parameter updates :
\begin{itemize}
\item $\Delta \thetaf= -\lambda_{\text{PG}}\sum_t \gamma^t \grad_\thetaf \log \pi(A_t|X_t) (G_t-V(X_t,\Phi_t)) + \lambda_{\text{H}} \sum_t\grad_\thetaf \calL_{\text{H}}(t) + \lambda_{\text{hs}} \sum_t\grad_\thetaf \calL_{\text{hs}}(t)$\\
where $\calL_{\text{H}}(t) = -\sum_a\pi(a|X_t)\log \pi(a|X_t)$ is an entropy bonus.
\item $\Delta \thetah=  \lambda_{\text{hs}}(t) \sum_t\grad_\thetah \calL_{\text{hs}} + \lambda_{\text{IM}} \sum_t \grad_\thetah (\calL_{\text{IM}}(t)-\beta \entH[A_t|X_t])$
\item $\Delta \omega= \sum_t\grad_\omega \calL_{\text{sup}}(t)$
\item $\Delta \lambda_{\text{IM}} = - \lambda_{\text{IM}}\sum_t(\calL_{\text{IM}}(t)-\beta \entH[A_t|X_t])$ (when using GECO)
\end{itemize}

\subsection{Design choices}

Here we detail practical choices for two aspects of the general CCA algorithm. These concern a) the form of the hindsight function, b) the form of the independence maximization constraint.

\subsubsection*{Choice of the hindsight function $\varphi$}
In principle, this function can take any form: in practice, we investigated two architectures. The first is a backward RNN, where $(\Phi_t, B_t) = \text{RNN}(X_t, B_{t+1})$, where $B_t$ is the state of the backward RNN. Backward RNNs are justified in that they can extract information from arbitrary length sequences, and allow making the statistics $\Phi_t$ a function of the entire trajectory. They also have the inductive bias of focusing more on near-future observations. 
The second is a transformer~\citep{vaswani2017attention,parisotto2019stabilizing}. Alternative networks could be used, such as attention-based networks~ \citep{hung2019optimizing} or RIMs~\citep{goyal2019recurrent}.

\subsubsection*{Independence maximization constraint $\calL_{\text{IM}}$}
We investigated two IM losses. The first is the conditional mutual information
$\MI(A_t;\Phi_t|X_t)=\E_{\Phi_t|X_t} [\entH[A_t|X_t]-\entH[A_t|X_t,\Phi_t]]$,
where $\mathbb{H}[A|B]$ denotes the conditional entropy  $\entH[A|B]=-\sum_a P(A=a|B) \log P(A=a|B)$. The expectation can be stochastically approximated by the trajectory sample value $\entH(A_t|X_t)-\entH(A_t|X_t,\Phi_t)$. 
The first term is simply the entropy of the policy $-\sum_a \pi(a|X_t) \log \pi(a|X_t)$. The second term is estimated using the $h$ network.
The second we investigated is the Kullback-Leibler divergence, $\mathbb{KL}(\pi(A_t|X_t)||\mathbb{P}(A_t|X_t,\Phi_t)) = \sum_a \pi(a|X_t) \log \pi(a|X_t) - \sum_a \pi(a|X_t) \log \mathbb{P}(a|X_t,\Phi_t)$. Again, we approximate the second term using $h$. We did not see significant differences between the two, with the KL slightly outperforming the mutual information.

\FloatBarrier 

\section{Additional Experimental Details}
\label{sec:add_exps}
\subsection{Bandits}
\label{sec:bandits_extra}

\subsubsection{Environment}
\label{sec:bandits_env}

Our bandit with feedback environment is defined by two positive integers $(N,K)$, a noise level $\sigma_r>0$ and three arbitrary matrices $U,V,W$, where $U,V \in \mathbb{R}^{K\times N}$ and $W\in \mathbb{R}^K$. For each replication of the experiment (i.e. each seed), these matrices are sampled from a standard Gaussian distribution and kept constant throughout all episodes.
For each episode (of length $1$, since this is a bandit problem tackled without meta-learning), we sample a context $-N \leq C \leq N$. Given $C$, an agent chooses an action $-N\leq A\leq N$. The agent then receives a reward $R=-(C-A)^2 + \epsilon_r$, where $\epsilon_r$ is sampled from $\mathcal{N}(0,\sigma_r)$. The agent additionally receives a $K$-dimensional feedback vector $F=U_C + V_A + W\epsilon_r$, where $U_C$ (resp. $V_A$) denotes the $\text{C}^{\text{th}}$ (resp. $\text{A}^{\text{th}}$) column of U (resp. V).

The choices above were made without any particular intent: we would expect the intuitions to generalize for other noise distributions and feedback functions. In section~\ref{sec:bandits_additional_results}, we investigate a decentralized multiagent variant of this problem where the exogenous noise actually corresponds to other players' actions.

\subsubsection{Architecture}
\label{sec:bandits_arch}
For the bandit problems, the agent architecture is as follows:
\begin{itemize}
    \item The hindsight feature $\Phi$ is computed by a backward RNN. We tried multiple cores for the RNN: GRU \citep{chung2015gated} with $32$ hidden units, a recurrent adder ($h_t=h_{t-1}+\text{MLP}(x_t)$, where the $\text{MLP}$ has two layers of $32$ units), or an exponential averager ($h_t=\lambda h_{t-1} + (1-\lambda) \text{MLP}(x_t)$).
    \item The hindsight classifier $h_\omega$ is a simple MLP with two hidden layers with $32$ units each.
    \item The policy and value functions are computed as the output of a simple linear layer with concatenated observation and feedback as input.
    \item All weights are jointly trained with Adam~\citep{kingma2014adam}.
    \item Hyperparameters are chosen as follows: learning rate $4e{\shortminus4}$, entropy loss $4e{\shortminus3}$, independence maximization tolerance $\betaim=0.1$, $\lambda_{\text{sup}}=\lambda_{\text{hs}}=1$, $\lambda_{\text{IM}}$ is set through Lagrangian optimization (with GECO).
\end{itemize}
    
\subsubsection{Additional results}

\label{sec:bandits_additional_results}
\paragraph{Multi-agent Bandit Problem:}
In the multi-agent version, the environment is composed of $M$ replicas of the bandit with feedback task. Each agent $i=1,\ldots,M$ interacts with its own version of the environment, but feedbacks and rewards are coupled across agents. The multi-agent bandit is obtained by modifying the single agent version as follows:

\begin{itemize}
    \item The contexts $C^i$ are sampled i.i.d. from $\{-N,\ldots,N\}$. $C$ and $A$ now denote the concatenation of all agents' contexts and actions.
    \item The feedback tensor is $(M,K)$ dimensional, and is computed as $W_c\mathbf{1}(C)+W_a\mathbf{1}(A)+\epsilon_f$; where the $W$ are now three dimensional tensors. Effectively, the feedback for agent $i$ depends on the context and actions of all other agents. 
    \item The terminal joint reward is $\sum_i -(C^i-A^i)^2$ for all agents.
\end{itemize}

The multi-agent version does not require the exogenous noise $\epsilon_r$, as other agents play the role of exogenous noise; it is a minimal implementation of the example found in section~\ref{sec:story}.

\begin{figure}[h!]
\centering
\includegraphics[width=0.45\textwidth]{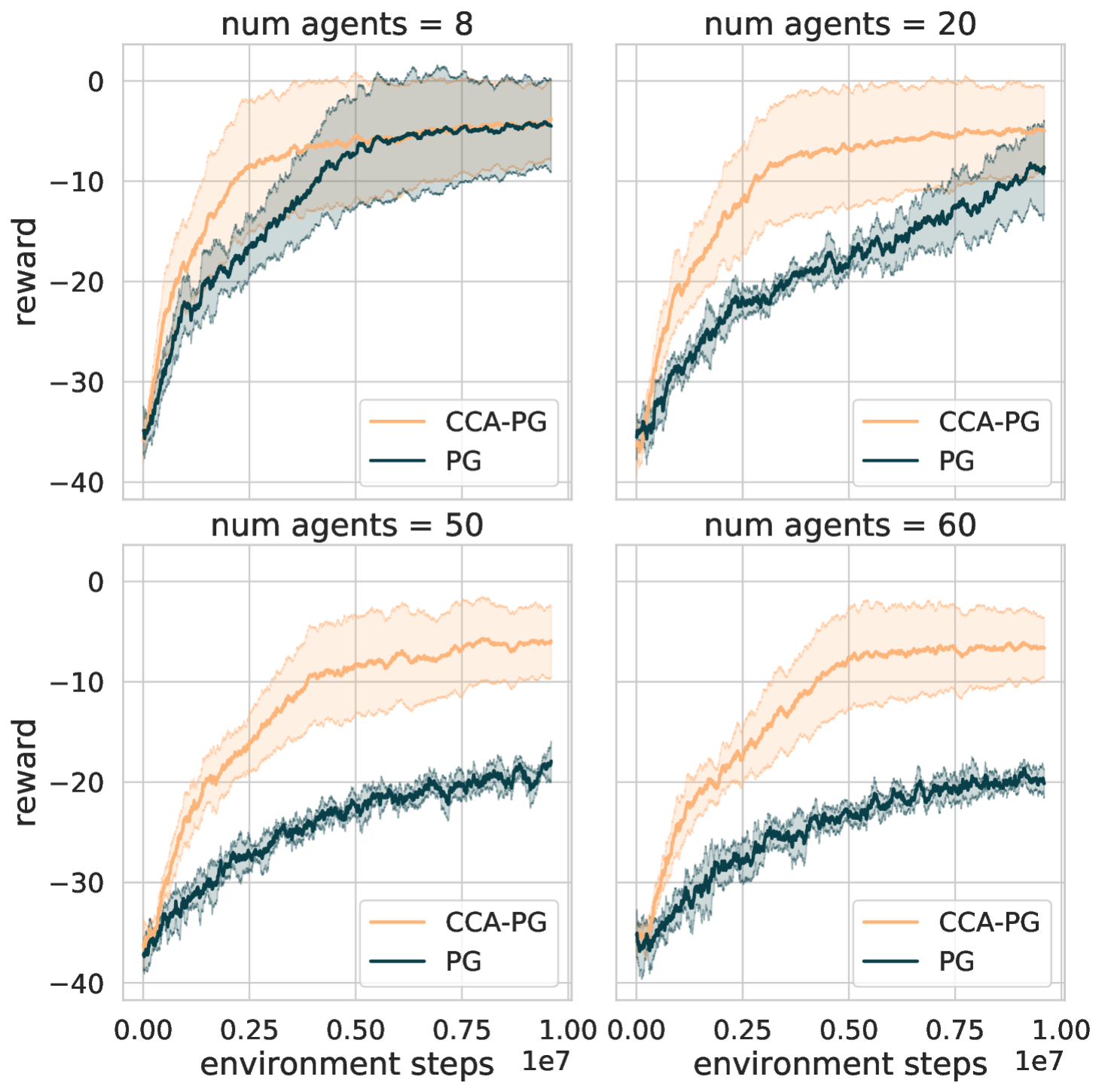}
\caption{
{\bf Multi-agent versions of the bandit problem.} CCA-PG agents outperform vanilla PG ones.}
\label{fig.multiagent}
\end{figure}

We report results from the multi-agent version of the environment in Fig.~\ref{fig.multiagent}. As the number of interacting agents increases, the effective variance of the vanilla PG estimator increases as well, and the performance of each agent decreases. In contrast, CCA-PG agents learn faster and reach higher performance (though they never learn the optimal policy).

\FloatBarrier
\subsection{Key to Door Tasks}
\label{sec:ktd}
\subsubsection{Environment details}
\label{sec:ktd_env_details}

\begin{figure}[H]
    \centering
    \includegraphics[width=0.45\textwidth]{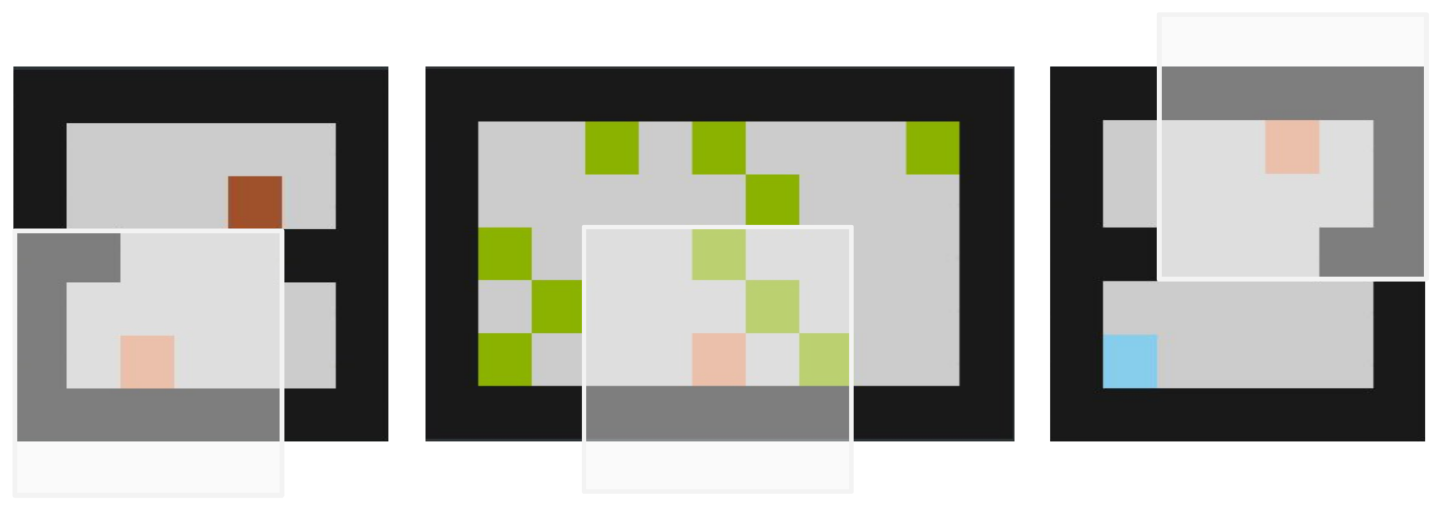}
    \caption{
    {\bf Key-To-Door environments visual.} The agent is represented by the beige pixel, key by brown, apples by green, and the final door by blue. The agent has a partial field of view, highlighted in white.}
    \label{fig.ktd}
\end{figure}

Observations returned by the key-to-door family of environments for each of the three phases can be visualized in Fig.~\ref{fig.ktd}.

\begin{table*}[ht!]
\centering
\begin{tabular}{@{}cp{3.5cm}|p{3cm}|p{3cm}@{}}
\toprule
 & & Lucky\newline(high apple reward) & Unlucky\newline(low apple reward) \\ \midrule
\multicolumn{1}{c|}{Hindsight Advantage Estimate} & Skillful\newline(Got key + Door)            & 1                         & 1                          \\ \cmidrule(l){2-4} 
\multicolumn{1}{c|}{}                                     & Unskillful\newline(Did not get key or door) & 0                         & 0                          \\ \midrule
\multicolumn{1}{c|}{Forward Advantage Estimate}   & Skillful\newline(Got key + Door)            & 46                        & -44                        \\ \cmidrule(l){2-4} 
\multicolumn{1}{c|}{}                                     & Unskillful\newline(Did not get key or door) & 45                        & -45                        \\ \bottomrule
\end{tabular}
\caption{The advantage estimate of the action of picking up a key in High-Variance-Key-To-Door, as computed by an agent that always picks up every apple, and never picks up the key or the door. We see that an advantage estimate learned using hindsight clearly differentiates between the skillful and unskillful actions; whereas for an advantage estimate learned without using hindsight, this difference is dominated by the extrinsic randomness.}
\label{tab.hvktd_advantages}
\end{table*}

To motivate our approach, Table~\ref{tab.hvktd_advantages} shows the advantage estimates for either picking up the key or not on High-Variance-Key-To-Door, for an agent that has a perfect apple-phase policy, but never picks up the key or door. Since there are 10 apples which can be worth 1 or 10, the return will be either 10 or 100. Thus the forward baseline in the key phase, i.e. before it has seen how much an apple is worth in the current episode, will be 55. As seen in Table~\ref{tab.hvktd_advantages}, the difference in advantage estimates due to `luck' is far larger than the difference in advantage estimates due to `skill' when not using hindsight. This makes learning difficult and leads to the policy never learning to start picking up the key or opening the door.
However, when we use a hindsight-conditioned baseline, we are able to learn a $\Phi$ (such as the value of a single apple in the current episode) that is completely independent from the actions taken by the agent, but which can provide a perfect hindsight-conditioned baseline of either 10 or 100.

\subsubsection{Architecture}
\label{sec:ktd_architecture}

The agent architecture is as follows:
\begin{itemize}
    \item The observations are first fed to 2-layer CNN with $(16, 32)$ output channels, kernel shapes of $(3, 3)$ and strides of $(1, 1)$. The output of the CNN is flattened and fed to a linear layer of size $128$.
    \item The agent state is computed by a forward LSTM with a state size of $128$. The input to the LSTM is the output of the previous linear layer, concatenated with the reward at the previous timestep.
    \item The hindsight feature $\Phi$ is computed either by a backward LSTM (i.e CCA-PG RNN) with a state size of 128 or by an attention mechanism~\cite{vaswani2017attention} (i.e CCA-PG Att) with value and key sizes of $64$, $1$ transformer block with $2$ attention heads, a 1 hidden layer MLP of size $1024$, an output size of 128 and a rate of dropout of $0.1$. The input provided is the concatenation of the output of the forward LSTM and the reward at the previous timestep.
    \item The policy is computed as the output of a single-layer MLP with $64$ units where the output of the forward LSTM is provided as input.
    \item The forward baseline is computed as the output of a 3-layer MLP of $128$ units each where the output of the forward LSTM is provided as input.
    \item The hindsight baseline is computed as the sum of the forward baseline and a hindsight residual baseline; the hindsight residual baseline is the output of a 3-layer MLP of $128$ units each where the concatenation of the output of the forward LSTM and the hindsight feature $\Phi$ is provided as input. It is trained to learn the residual between the return and the forward baseline.
    \item For CCA, the hindsight classifier $h_\omega$ is computed as the sum of the log of the policy outputs and the output of an MLP, with four hidden layers with $256$ units each where the concatenation of the output of the forward LSTM and the hindsight feature $\Phi$ is provided as input.
    \item For State HCA, the hindsight classifier $h_\omega$ is computed as the output of an MLP, with four hidden layers with $256$ units each, where the concatenation of the outputs of the forward LSTM at two given time steps is provided as input.
    \item For Return HCA, the hindsight classifier $h_\omega$ is computed as the output of an MLP, with four hidden layers with $256$ units each, where the concatenation of the output of the forward LSTM and the return is provided as input. 
    \item All weights are jointly trained with RMSprop~\citep{hinton2012neural} with epsilon $1e{\shortminus4}$, momentum 0 and decay $0.99$.
\end{itemize}

For High-Variance-Key-To-Door, the optimal hyperparameters found for each algorithm can be found in Table~\ref{tab.hvktd_hypers}.

For Key-To-Door, the optimal hyperparameters found for each algorithm can be found in Table~\ref{tab.ktd_hypers}.

The agents are trained on full-episode trajectories, using a discount factor of 0.99.

\subsubsection{Additional results}
\label{sec:ktd_additional_results}

\begin{figure}
\centering
\begin{minipage}{.45\textwidth}
  \centering
    \includegraphics[width=0.95\linewidth]{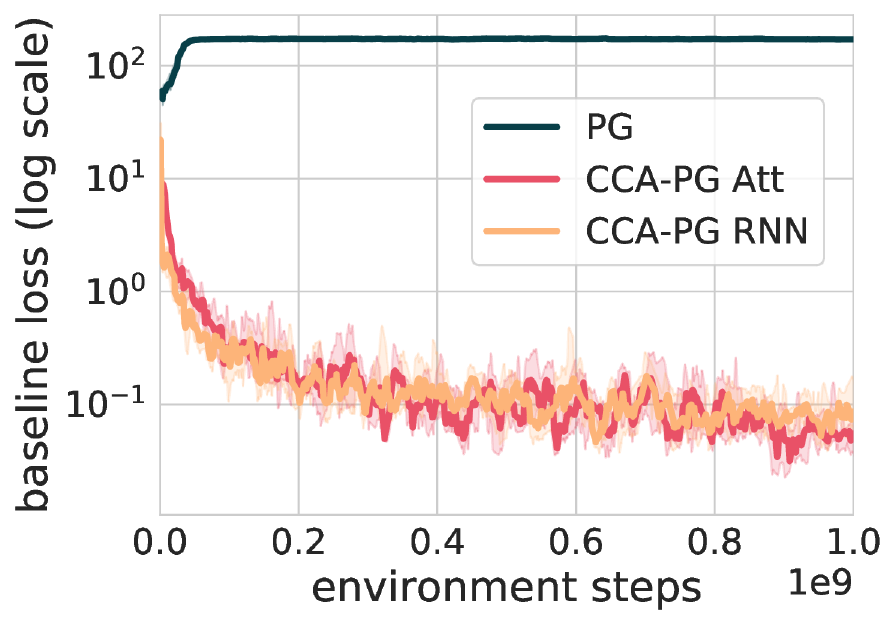}
    \caption{Baseline loss for vanilla PG versus hindsight baseline loss for CCA in {\bf High-Variance-Key-To-Door}.
    }
    \label{fig.hvktd.baseline}
\end{minipage}\hfill
\begin{minipage}{.45\textwidth}
  \centering
    \includegraphics[width=0.95\linewidth]{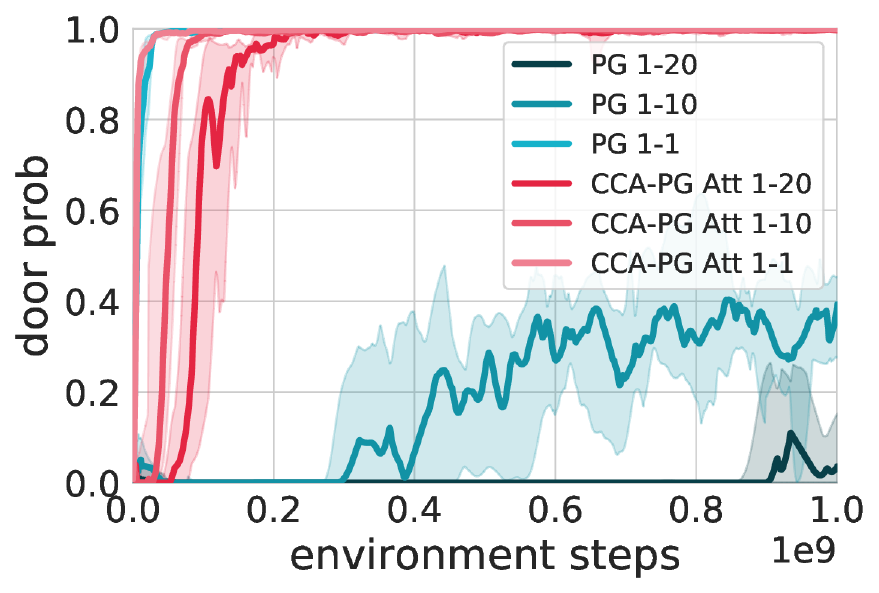}
    \caption{{\bf Impact of variance over credit assignment performances.} Probability of picking up the key and opening the door as a function of the variance level induced by the apple reward discrepancy between episodes.}
    \label{fig.hvktd.var}
\end{minipage}
\end{figure}

As shown in Fig.~\ref{fig.hvktd.baseline}, in the case of vanilla policy gradient, the baseline loss increases at first. As the reward associated with apples varies from one episode to another, getting more apples also means increasing the forward baseline loss. On the other hand, as CCA is able to take into account trajectory specific exogenous factors, the hindsight baseline loss can nicely decrease as learning takes place.

Fig.~\ref{fig.hvktd.var} shows the impact of the variance level induced by the apple reward discrepancy between episodes on the probability of picking up the key and opening the door. Thanks to the use of hindsight in its value function, CCA-PG is almost not impacted by this whereas vanilla PG sees its performances drop dramatically as variance increases.

\begin{figure}[h!]
    \centering
    \includegraphics[width=0.3\textwidth]{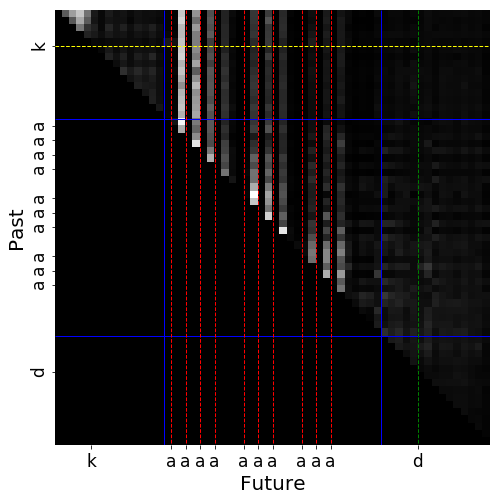}
    \includegraphics[width=0.44\textwidth]{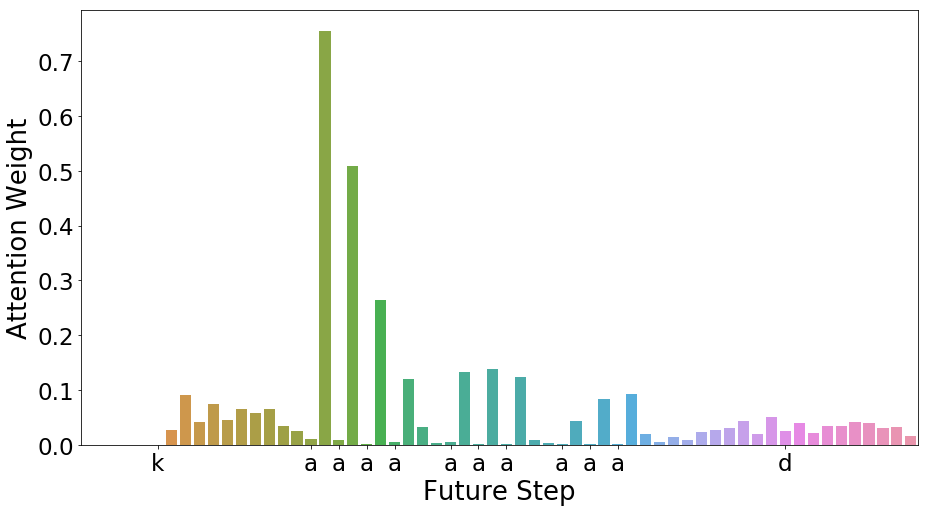}
    \caption{Visualization of attention weights on the High-Variance-Key-To-Door task. \textbf{Top:} a 2-dimensional heatmap showing how the hindsight function at each step attends to each step in the future. Red lines indicate the timesteps at which apples are picked up (marked as `a'); green indicates the door (marked as `d'); yellow indicates the key (marked as `k'). \textbf{Bottom:} A bar plot of attention over future timesteps, computed at the step where the agent is just about to pick up the key.}
  \label{fig.hvktd.viz}
\end{figure}

Fig.~\ref{fig.hvktd.viz} shows a qualitative analysis of the attention weights learned by CCA-PG Att on the High-Variance-Key-To-Door task. For this experiment, we used only a single attention head for easier interpretation of the hindsight function, and show both a heatmap of the attention weights over the entire episode, and a histogram of attention weights at the step where the agent picks up the key.
As expected, the most attention is paid to timesteps just after the agent picks up an apple - since these are the points at which the apple reward is provided to the $\Phi$ computation. In particular, very little attention is paid to the timestep where the agent opens the door. These insights further show that the hindsight function learned is highly predictive of the episode return, while not having mutual information with the action taken by the agent, thus ensuring an unbiased policy gradient estimator.

\begin{table*}[ht]
\centering
\begin{tabular}{|l|c|c|c|c|c|}\hline
                          & CCA Att  & CCA RNN &   PG   & State HCA & Return HCA  \\\hline
Policy cost               &    1     &    1     &  1     &    1      &    1        \\
Entropy cost              &    5e-3  &   5e-3   &  5e-3  &    5e-3   &    5e-3     \\
Forward baseline cost     &    5e-2  &   5e-2   &  5e-2  &    5e-2   &    5e-2     \\
Hindsight residual baseline cost &    5e-2  &   5e-2   &\hrulefill  & \hrulefill&\hrulefill             \\
Hindsight classifier cost &    1e-2  &   1e-2   & \hrulefill&    1e-2   &    1e-2     \\
Action independence cost  &    1e2   &   1e2    &\hrulefill &\hrulefill &\hrulefill             \\
Learning rate             &    5e-4  &   5e-4   &  1e-4  &    5e-4   &    1e-3     \\\hline
\end{tabular}
\caption{High-Variance-Key-To-Door hyperparameters}
\label{tab.hvktd_hypers}
\end{table*}

\begin{table*}[ht]
\centering
\begin{tabular}{|l|c|c|c|c|c|}\hline
                          & CCA Att & CCA RNN &   PG    & State HCA & Return HCA  \\ \hline
Policy cost               &    1     &    1     &  1     &    1      &    1        \\
Entropy cost              &    5e-3  &   5e-3   &  5e-3  &    5e-3   &    5e-3     \\
Forward baseline cost     &    5e-2  &   5e-2   &  5e-2  &    5e-2   &    5e-2     \\
Hindsight residual baseline cost &    5e-2  &   5e-2   &\hrulefill&\hrulefill& \hrulefill            \\
Hindsight classifier cost &    1e-2  &   1e-2   &\hrulefill&    1e-2   &    1e-2     \\
Action independence cost  &    1e2   &   1e2    &\hrulefill & \hrulefill &\hrulefill             \\
Learning rate             &    5e-4  &   5e-4   &  5e-4  &    5e-4   &    5e-4     \\\hline
\end{tabular}
\caption{Key-To-Door hyperparameters}
\label{tab.ktd_hypers}
\end{table*}

\subsection{Task Interleaving}
\label{sec:interleaving}

\subsubsection{Environment details}
\label{sec:add_exps.interleaving.env}

For each task, a random, but fixed through training, set of 5 out of 10 colored squares are leading to a positive reward. Furthermore, a small reward of 0.5 is provided to the agent when it picks up any colored square. As mentioned previously, each episode are 140 steps long and it takes at least 9 steps for the agent to reach one colored square from its initial position.

The 6 tasks we consider (numbered $\#1$ to $\#6$) are respectively associated with a reward of 80, 4, 100, 6, 2 and 10. Tasks $\#$2, $\#$4, $\#$5 and $\#$6 are referred to as `hard’ while tasks $\#$1 and $\#$3 as `easy’ because of their large associated rewards. The settings 2, 4 and 6-task are respectively considering tasks 1-2, 1-4 and 1-6.

\subsubsection{Architecture}
\label{sec:architecture.interleaving}
We use the same architecture setup as reported in Appendix~\ref{sec:ktd_architecture}. The agents are also trained on full-episode trajectories, using a discount factor of 0.99.

For Task Interleaving, the optimal hyperparameters found for each algorithm can be found in Table~\ref{tab.interleaving_hypers}.

\subsubsection{Additional results}
\label{sec:add_exps.interleaving}

\begin{figure}[h!]
    \centering
    \includegraphics[width=0.25\textwidth]{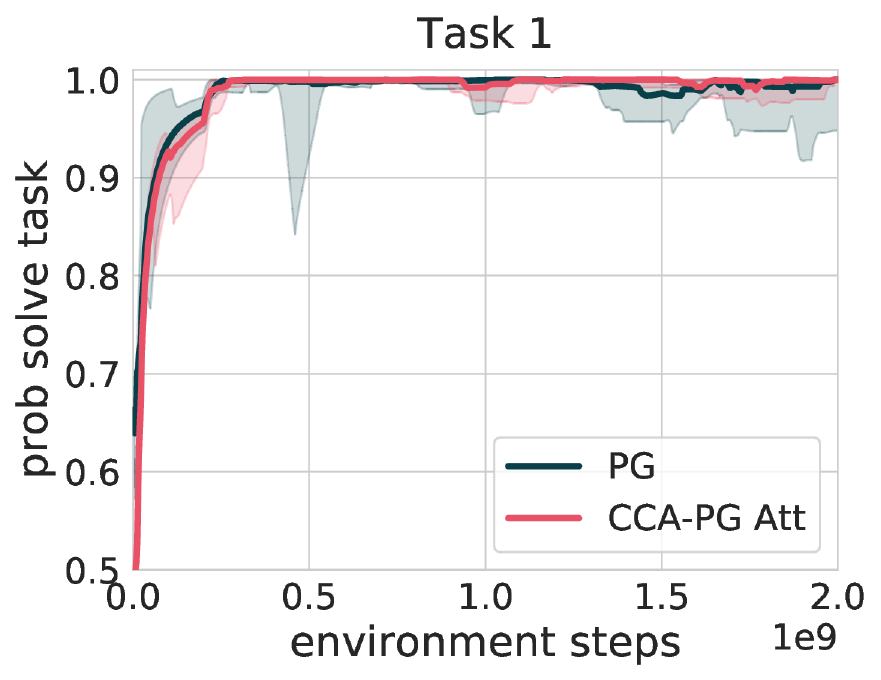}
    \includegraphics[width=0.25\textwidth]{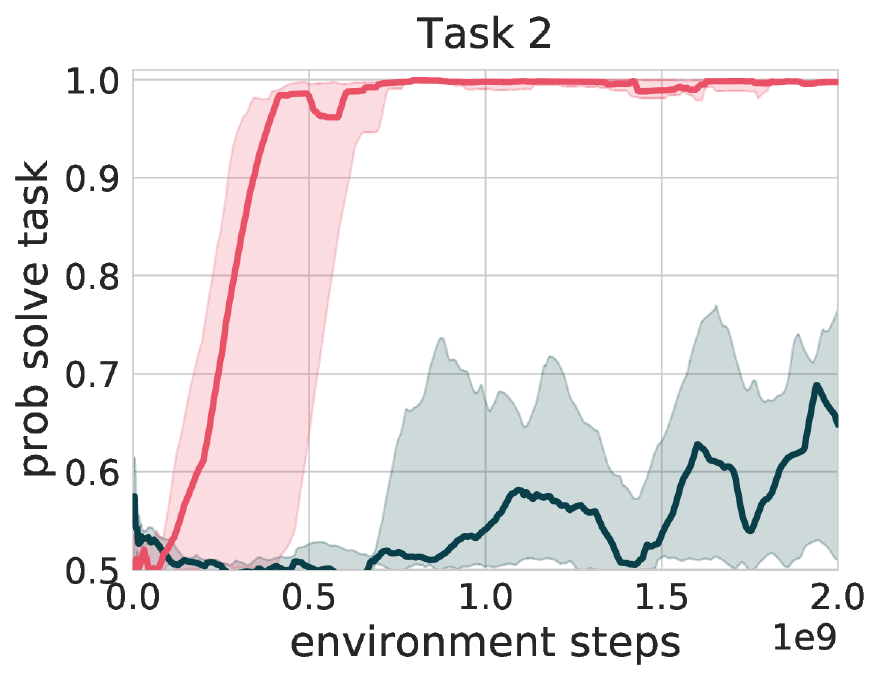}
    \includegraphics[width=0.25\textwidth]{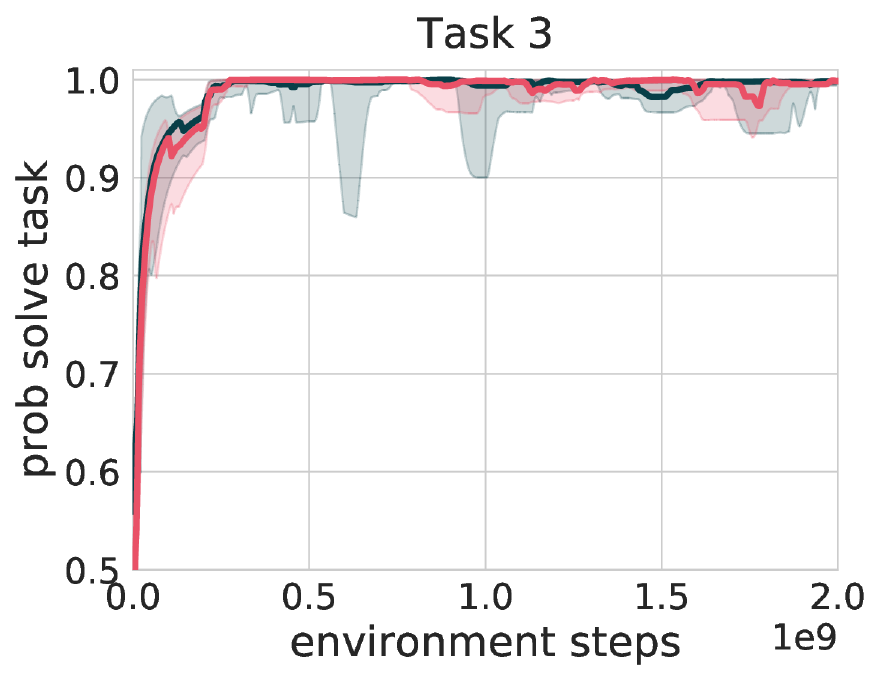}
    \includegraphics[width=0.25\textwidth]{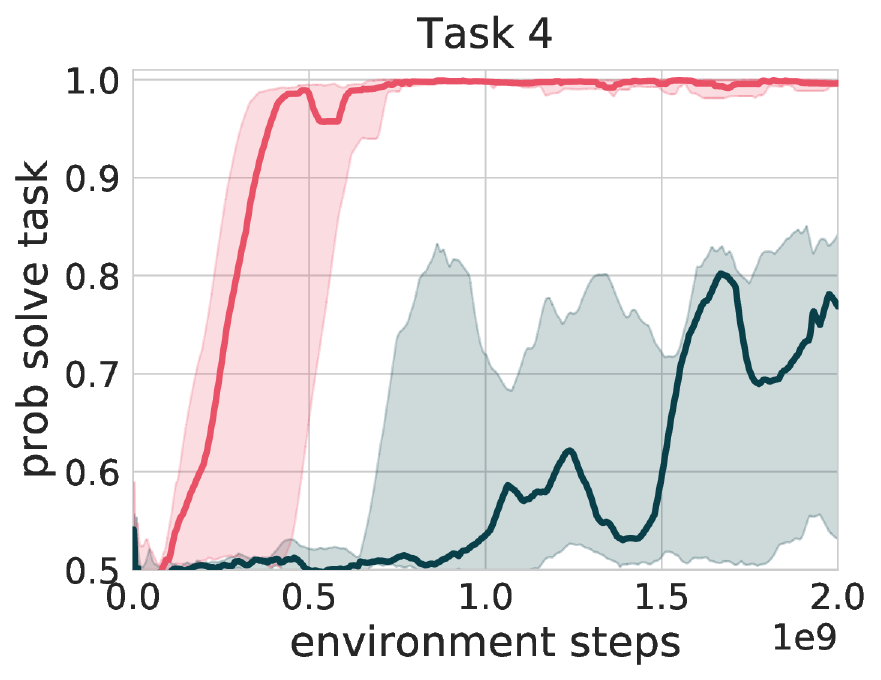}
    \includegraphics[width=0.25\textwidth]{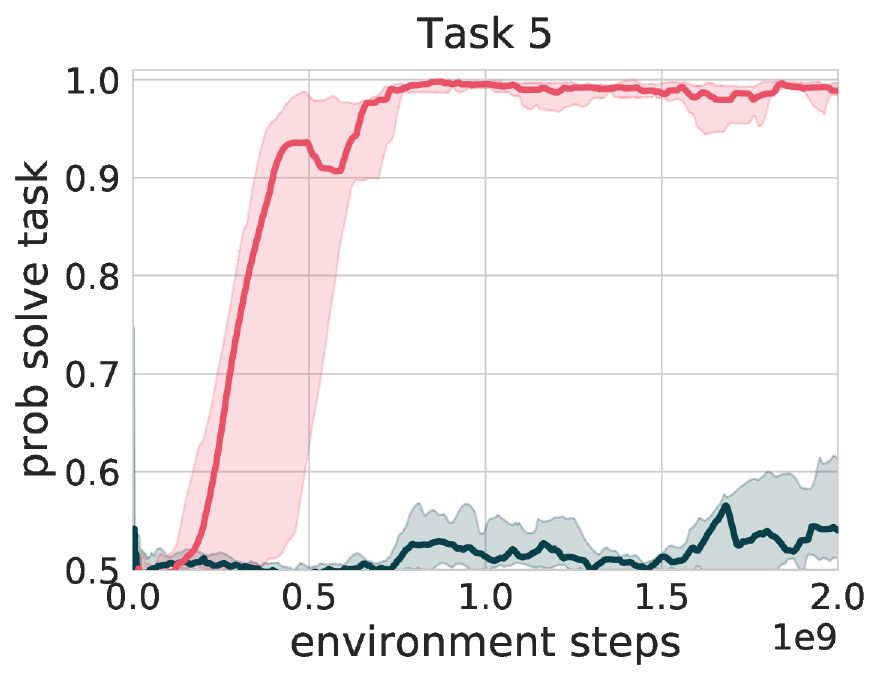}
    \includegraphics[width=0.25\textwidth]{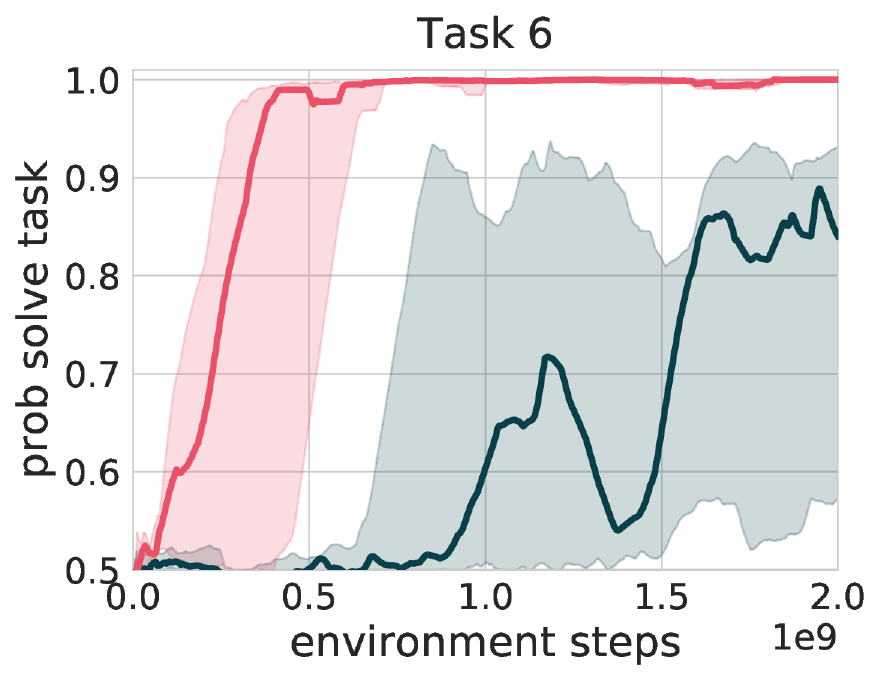}
    \caption{
    Probability of solving each task in the 6-task setup for {\bf Task Interleaving}.}
    \label{fig.interleaving_prob_per_task}
\end{figure}

Fig.~\ref{fig.interleaving_prob_per_task} shows that CCA is able to solve all 6 tasks quickly despite the variance induced by the exogenous factors. Vanilla PG on the other hand despite solving the `easy’ tasks 1 and 3 for which the agent receives big rewards, it is incapable of reliably solve the 4 remaining tasks for which the associated reward is smaller. This helps unpacking~Fig.~\ref{fig.interleaving_results}.

\subsubsection{Ablation Study}
\label{sec:ablation.interleaving}

\begin{figure}[h!]
\centering
\begin{minipage}{.31\textwidth}
  \centering
    \includegraphics[width=0.95\linewidth]{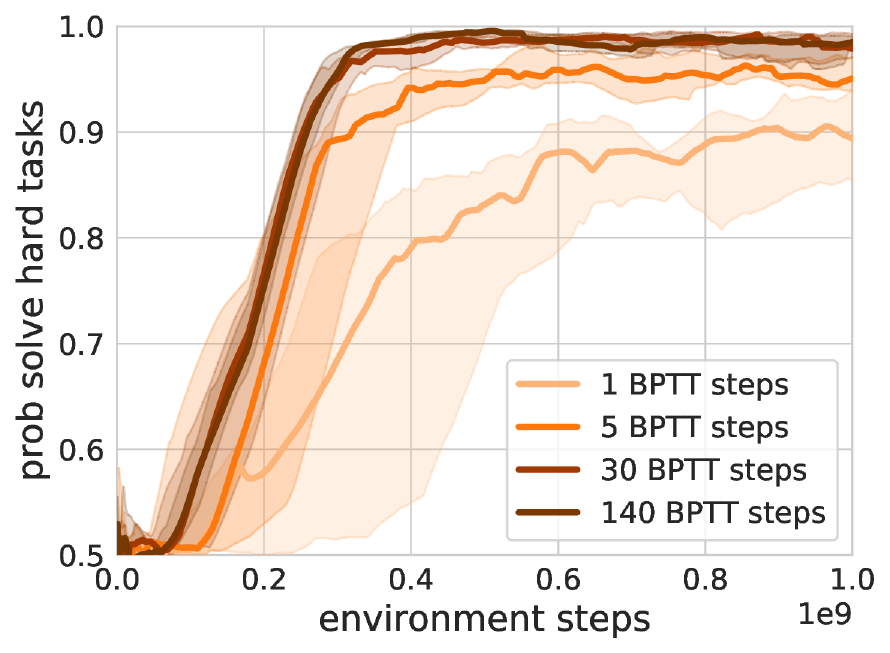}
    \caption{{\bf Impact of the number of back-propagation through time steps performed into the hindsight function for CCA-RNN.} Probability of solving the `hard’ tasks in the 6-task setup of {\bf Task Interleaving}.} 
    \label{fig.ablation.bptt}
\end{minipage}\hfill
\begin{minipage}{.62\textwidth}
  \centering
    \includegraphics[width=0.47\linewidth]{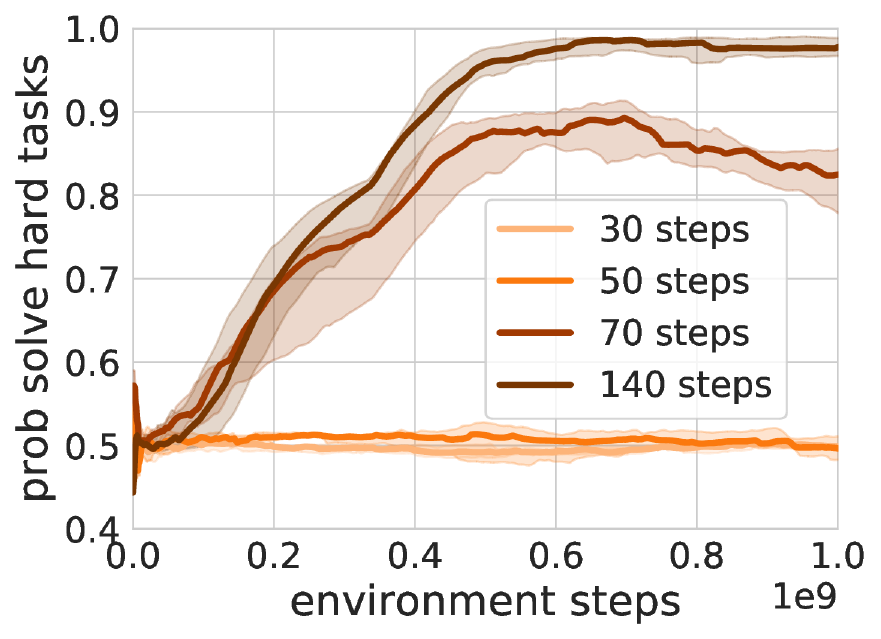}
    \includegraphics[width=.47\textwidth]{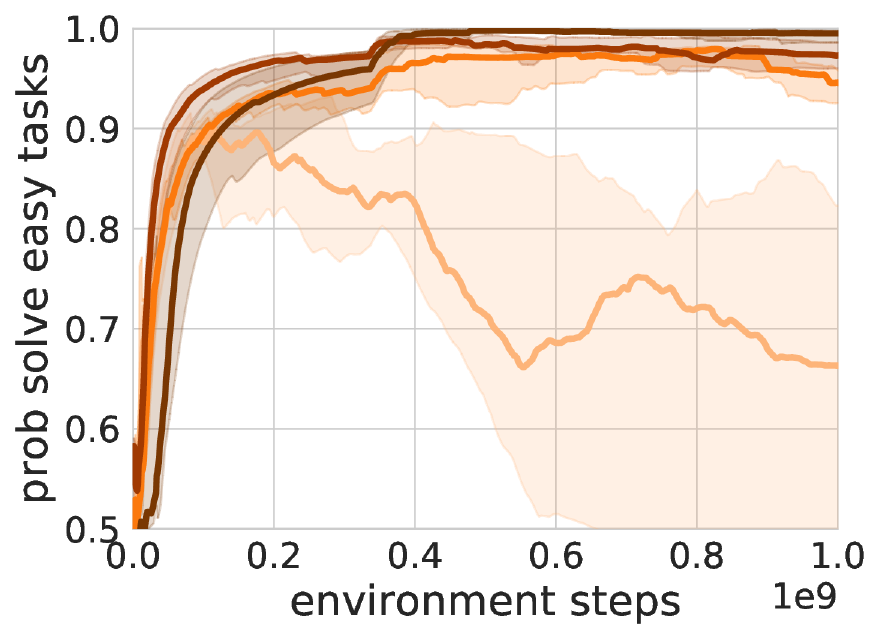}
    \label{fig.interleaving_ablation_unroll_easy}
    \caption{
    {\bf Impact of the unroll length for CCA-RNN}. Probability of solving the `hard’ and `easy’ tasks in the 6-task setup of {\bf Task Interleaving}.}
    \label{fig.interleaving_ablation_unroll}
\end{minipage}
\end{figure}

Fig.\ref{fig.ablation.bptt} shows the impact of the number of back-propagation through time steps performed into the backward RNN of the hindsight function while performing full rollouts. This shows that learning in `hard’ tasks, i.e.\ where hindsight is crucial for performances, is not much impacted by the number of back-propagation steps performed into the backward RNN. This is great news as this indicates that learning in challenging credit assignment tasks still works when the hindsight function sees the whole future but can only backprop through a limited window.  

Fig.\ref{fig.interleaving_ablation_unroll} shows how performances of CCA-RNN are impacted by the unroll length. As expected, the less it is able to look into the future, the harder it becomes to solve hard credit assignment tasks as it is limited in its capacity to take into account exogenous effects.

The two previous results are promising since CCA seems to only require to have access to as many steps into the future as possible while not needing to do back-propagation through the full sequence. 

\begin{table}[h!]
\centering
\begin{tabular}{|l|c|c|c|c|c|}
\hline
                          & CCA Att & CCA RNN &   PG    \\\hline
Policy cost               &    1     &    1     &  1     \\
Entropy cost              &    5e-2  &   5e-2   &  5e-2  \\
Forward baseline cost     &    1e-2  &   5e-3   &  5e-2  \\
Hindsight residual baseline cost &    1e-2  &   5e-3   &  \hrulefill \\
Hindsight classifier cost &    1e-2  &   1e-2   &  \hrulefill \\
Action independence cost  &    1e1   &   1e1    &  \hrulefill \\
Learning rate             &    5e-4  &   5e-4   &  1e-3  \\\hline
\end{tabular}
\caption{Task Interleaving hyperparameters}
\label{tab.interleaving_hypers}
\end{table}

\newpage
\FloatBarrier
\FloatBarrier
\section{Relation between HCA, CCA, and FC estimators}
\label{hca-cca-fc}
The FC estimators generalize both the HCA and CCA estimators. From FC, we can derive CCA by assuming that $\Phi_t$ and $A_t$ are conditionally independent (see next section). We can also derive state and return HCA from FC.

For return HCA, we obtain both an all-action and baseline version of return HCA by choosing $\Phi_t=G_t$. For state HCA, we first need to decompose the return into a sum of rewards, and apply the policy gradient estimator to each reward separately. For a pair $(X_t, R_{t+k})$, and assuming that $R_{t+k}$ is a function of $X_{t+k}$ for simplicity, we choose $\Phi_t=X_{t+k}$. We then sum the different FC estimators for different values of $k$ and obtain both an all-action and single-action version of state HCA.

Note however that HCA and CCA \emph{cannot} be derived from one another. Both estimators leverage different approaches for unbiasedness, one (HCA) leveraging importance sampling, and the other (CCA) eschewing importance sampling in favor of constraint satisfaction (in the context of inference, this is similar to the difference between obtaining samples of the posterior by importance sampling versus directly parameterizing the posterior distribution).

\FloatBarrier

\section{Proofs}

\subsection{Policy gradients}

\begin{proof}[Proof of equation~\ref{eq:REINFORCE}]
By linearity of expectation, the expected return can be written as $\E[G] = \sum_t \gamma^t \E[R_t]$. Writing the expectation as an integral over trajectories, we have:
\begin{align*}
    \E[R_t] = \sum_{\substack{x_0,\ldots, x_t\\a_0,\ldots, a_t}} \left(\prod_{s\leq t} \left(\pi_\theta(a_s|x_s) P(x_{s+1}|x_s, a_s)\right)\right) R(x_t, a_t)
\end{align*}
Taking the gradient with respect to $\theta$:
\begin{multline*}
    \nabla_\theta \E[R_t] = \sum_{\substack{x_0,\ldots, x_t\\a_0,\ldots, a_t}} \Bigg( \Bigg.\sum_{s'\leq t} \nabla_\theta  \pi_\theta(a_{s'}|x_{s'})P(x_{s'+1}|x_{s'}, a_{s'})
    \bigg(\prod_{s\leq t, s\not=s'} (\pi_\theta(a_s|x_s) P(x_{s+1}|x_s, a_s))\bigg)\Bigg. \Bigg)  R(x_t, a_t)
\end{multline*}
We then rewrite $\nabla_\theta \pi_\theta(a_{s'}|x_{s'}) = \nabla_\theta \log \pi_\theta(a_{s'}|x_{s'})\pi_\theta(a_{s'}|x_{s'})$, and obtain
\begin{align*}
    \nabla_\theta \E[R_t] =& \sum_{\substack{x_0,\ldots, x_t\\a_0,\ldots, a_t}} \Bigg(\Bigg.\sum_{s'\leq t} \nabla_\theta \pi_\theta(a_{s'}|x_{s'})\bigg(\prod_{s\leq t, s} (\pi_\theta(a_s|x_s) P(x_{s+1}|x_s, a_s))\bigg)\Bigg. \Bigg)  R(x_t, a_t)\\
    =& \E\left[\sum_{s'\leq t} \nabla_\theta \log \pi_\theta(A_{s'}|X_{s'}) R_t\right]
\end{align*}

Summing over $t$, we obtain
\begin{align*}
    \nabla_\theta \E[G]=& \E\left[\sum_{t\geq 0} \gamma^t \sum_{s'\leq t} \nabla_\theta \log \pi_\theta(A_{s'}|X_{s'}) R_t\right]
\end{align*}
which can be rewritten (with a change of variables):
\begin{align*}
    \nabla_\theta \E[G]=& \E\left[\sum_{t\geq 0} \nabla_\theta \log \pi_\theta(A_{t}|X_{t}) \sum_{t'\geq t}\gamma^{t'}  R_{t'}\right]\\
    =& \E\left[\sum_{t\geq 0} \gamma^{t} \nabla_\theta \log \pi_\theta(A_{t}|X_{t}) \sum_{t'\geq t} \gamma^{t'-t} R_{t'}\right]\\
    =& \E\left[\sum_{t\geq 0} \gamma^t S_t G_t\right]
\end{align*}

To complete the proof, we need to show that $\E[S_t V(X_t)]=0$. By iterated expectation, $\E[S_t V(X_t)] = \E[\E[S_t V(X_t)|X_t]]=\E[V(X_t)\E[S_t |X_t]]$, and we have $\E[S_t |X_t]=\sum_a \nabla_\theta \pi(a|X_t)= \nabla_\theta (\sum_a \pi(a|X_t)) = \nabla_\theta 1 =0$.
\end{proof}

\begin{proof}[Proof of equation~\ref{eq:AAPG}]
We start from the single action policy gradient $\nabla_\theta \E[G]=\E\left[\sum_{t\geq 0} \gamma^t S_t G_t\right]$ and analyse the term for time t, $\mathbb{E}[S_t G_t]$. 
\begin{align*}
    \mathbb{E}[S_t G_t] =& \mathbb{E}[\mathbb{E}[S_t G_t|X_t,A_t]]\\
    =& \mathbb{E}[S_t\mathbb{E}[G_t|X_t,A_t]]\\
    =& \mathbb{E}[S_t Q(X_t, A_t)]\\
    =& \mathbb{E}\left[\mathbb{E}[S_t Q(X_t, A_t)|X_t]\right]\\
    =& \mathbb{E} \left[\sum_a \nabla_\theta \pi_\theta(a|X_t) Q(X_t,a)\right]
\end{align*}
The first and fourth inequality come from different applications of iterated expectations, the second from the fact $S_t$ is a constant conditional on $X_t,A_t$, and the third from the definition of $Q(X_t,A_t)$.
\end{proof}

\subsection{Proof of FC-PG theorem}

\begin{proof}[Proof of theorem~\ref{thm:FC-PG} (single action)]
We need to show that 
$\E\Big[S_t \frac{\pi(A_t|X_t)}{P(A_t|X_t,\Phi_t)}V(X_t,\Phi_t)\Big]=0$, so that\\ $\frac{\pi(A_t|X_t)}{P(A_t|X_t,\Phi_t)}V(X_t,\Phi_t)$ is a valid baseline.
As previously, we proceed with the law of iterated expectations, by conditioning successively on $X_t$ then $\Phi_t$
\begin{align*}
\E\left[S_t \frac{\pi(A_t|X_t)}{P(A_t|X_t,\Phi_t)}V(X_t,\Phi_t)\right]
=&\E\bigg[\E\bigg[S_t \frac{\pi(A_t|X_t)}{P(A_t|X_t,\Phi_t)}
V(X_t,\Phi_t)\bigg\rvert X_t, \Phi_t\bigg.\bigg]\bigg.\bigg]\\
=&\E\bigg[\bigg.V(X_t,\Phi_t)\E\bigg[\bigg.S_t\frac{\pi(A_t|X_t)}{P(A_t|X_t,\Phi_t)}\bigg\rvert X_t, \Phi_t\bigg.\bigg]\bigg.\bigg]
\end{align*}
Then we note that 
\begin{align*}
\E\left[S_t \frac{\pi(A_t|X_t)}{P(A_t|X_t,\Phi_t)}\bigg\rvert X_t, \Phi_t\right]=&\sum_a P(a|X_t, \Phi_t) \grad \log \pi(a|X_t)\frac{\pi(a|X_t)}{P(a|X_t,\Phi_t)}\\
=&\sum_a \grad \pi(a|X_t)=0.
\end{align*}
\end{proof}

\begin{proof}[Proof of theorem~\ref{thm:FC-PG} (all-action)]
We start from the definition of the $Q$ function:
\begin{align*}
Q(X_t,a)=\:&\mathbb{E}\left[G_t|X_t,A_t=a\right]\\
        =\:&\mathbb{E}_{\Phi_t}\left[\mathbb{E}\left[G_t|X_t,\Phi_t,A_t=a\right]|X_t,A_t=a\right]\\
        =\:&\int_\phi P(\Phi=\varphi|X_t,A_t=a)Q(X_t,\Phi_t=\varphi,a)
\end{align*}
We also have $$P(\Phi=\varphi|X_t,A_t)= \frac{P(\Phi=\varphi|X_t)P(A_t=a|X_t,\Phi_t=\phi)}{P(A_t=a|X_t)},$$which combined with the above, results in:
\begin{align*}
Q(X_t,a)=&\int_\phi P(\Phi=\varphi|X_t)\frac{P(A_t=a|X_t,\Phi_t=\phi)}{\pi_\theta(a|X_t)} Q(X_t,\Phi_t,a)\\
=& \mathbb{E}\left[\frac{P(A_t=a|X_t,\Phi_t=\phi)}{\pi_\theta(a|X_t)}Q(X_t,\Phi_t,a)\bigg\rvert X_t\right]
\end{align*}
For the compatibility with policy gradient, we start from:
\begin{align*}
    \mathbb{E}[S_t G_t]=& \mathbb{E} \left[\sum_a \nabla_\theta \pi_\theta(a|X_t) Q(X_t,a)\right]
\end{align*}
We replace $Q(X_t,a)$ by the expression above and obtain
\begin{align*}
\mathbb{E}[S_t G_t]=&\mathbb{E} \bigg[\bigg.\sum_a \nabla_\theta \pi_\theta(a|X_t) \mathbb{E}\bigg[\bigg.\frac{P(A_t=a|X_t,\Phi_t=\phi)}{\pi_\theta(a|X_t)} Q(X_t,\Phi_t,a)\bigg\rvert X_t\bigg.\bigg]\bigg.\bigg]\\
=& \mathbb{E} \bigg[\bigg. \mathbb{E}\bigg[\bigg.\sum_a \nabla_\theta \pi_\theta(a|X_t)\frac{P(A_t=a|X_t,\Phi_t=\phi)}{\pi_\theta(a|X_t)}Q(X_t,\Phi_t,a)\bigg\rvert X_t\bigg.\bigg]\bigg.\bigg]\\
=& \mathbb{E} \bigg[\bigg. \mathbb{E}\bigg[\bigg.\sum_a \nabla_\theta \log \pi_\theta(a|X_t)P(A_t=a|X_t,\Phi_t=\phi) Q(X_t,\Phi_t,a)\bigg\rvert X_t\bigg.\bigg]\bigg.\bigg]\\
=& \mathbb{E} \bigg[\bigg.\sum_a \nabla_\theta \log \pi_\theta(a|X_t)P(A_t=a|X_t,\Phi_t=\phi)Q(X_t,\Phi_t,a)\bigg.\bigg]
\end{align*}

Note that in the case of a large number of actions, the above can be estimated by$$\frac{\grad_\theta \log \pi_\theta(A_t'|X_t)P(A_t'|X_t,\Phi_t=\phi)}{\pi_\theta(A_t'|X_t)} Q(X_t,\Phi_t,A_t'),$$ where $A_t'$ is an independent sample from $\pi(.|X_t)$; note in particular that $A_t'$ shall NOT be the action $A_t$ that gave rise to $\Phi_t$, which would result in a biased estimator.

\subsection{Proof of proposition~\ref{propIM} (existence of independence-maximization loss)}

Since the KL is non-negative, $E_\Phi[\text{KL}[P(A_t|X_t))||P(A_t|X_t,\Phi_t)]=0$ implies that $\text{KL}[P(A_t|X_t))||P(A_t|X_t,\Phi_t]$ is $0$ almost everywhere. Furthermore, for a given $\phi$, $\text{KL}[P(A_t|X_t))||P(A_t|X_t,\Phi_t=\phi)]=0$ implies $P(A_t=a|X_t)=P(A_t=a|X_t, \Phi_t=\phi)$ for all $a$. Multiplying by $p(\Phi_t=\phi|X_t)$, we obtain $P(A_t=a, \Phi_t=\phi|X_t) = P(A_t=a|X_t) P(\Phi_t=\phi|X_t)$, from which conditional independence follows.

\subsection{Proof of CCA-PG theorem}

Assume that $\Phi_t$ and $A_t$ are conditionally independent on $X_t$. Then, $\frac{P(A_t=a|X_t,\Phi_t=\phi)}{P(A_t=a|X_t)}=1$. In particular, it is true when evaluating at the random value $A_t$. From this simple observation, both CCA-PG theorems follow from the FC-PG theorems.

To prove the lower variance of the hindsight advantage estimate, note that 
\begin{align*}
    \V[G_t-V(X_t,\Phi)]=&\E[(G_t-V(X_t,\Phi_t))^2]\\
    =& \E[G_t^2] - \E[V(X_t,\Phi_t)^2]\\
    \V[G_t-V(X_t)]=&\E[(G_t-V(X_t))^2]\\
    =& \E[G_t^2] - \E[V(X_t)^2]
\end{align*}
To prove the first statement, we have $(G_t-V(X_t,\Phi_t))^2 = G_t^2 + V(X_t,\Phi_t)^2-2G_t V(X_t,\Phi_t)$, and apply the law of iterated expectations to the last term:
\begin{align*}
    \E[G_t V(X_t,\Phi_t)] =&\E[\E[G_t V(X_t,\Phi_t)|X_t,\Phi_t]]\\
    =& \E[V(X_t,\Phi_t)\E[G_t|X_t,\Phi_t]]\\
    =& \E[V(X_t,\Phi_t)^2]
\end{align*}
The proof for the second statement is identical. Finally, we note that by Jensen's inequality, we have $\E[V(X_t,\Phi_t)^2] \leq \E[V(X_t)^2]$, from which we conclude that $\mathbb{V}[G_t-V(X_t,\Phi_t)]\leq \mathbb{V}[G_t-V(X_t)]$.

\subsection{Proof of generative-CCA results}

\begin{figure}[h!]
\centering
\centering
    \resizebox{0.35\textwidth}{!}{%
        \begin{tikzpicture}[scale=0.85, >=stealth]

\tikzstyle{empty}=[]
\tikzstyle{lat}=[circle, inner sep=1pt, minimum size = 6.5mm, thick, draw =black!80, node distance = 20mm, scale=0.9]
\tikzstyle{directed}=[->, thick, shorten >=0.5 pt, shorten <=1 pt]

\node[lat] at (0,0) (x) {$X_t$};
\node[lat] at (1,2) (a) {$A_t$};
\node[lat] at (1,-2) (e) {$\varepsilon$};
\node[lat] at (2,0) (y) {$Y_t$};
\node[lat] at (4,0) (phi) {$\Phi_\varepsilon$};

\path   (x) edge [directed] (a)
        (x) edge [directed] (y)
        (a) edge [directed] (y)
        (e) edge [directed] (y)
        (a) edge [directed] (phi)
        (y) edge [directed] (phi)
        (x) [bend right] edge [directed] (phi);

\end{tikzpicture}
    }  
\caption{Graphical model for generative CCA
    }
\label{fig:genCCA}
\end{figure}
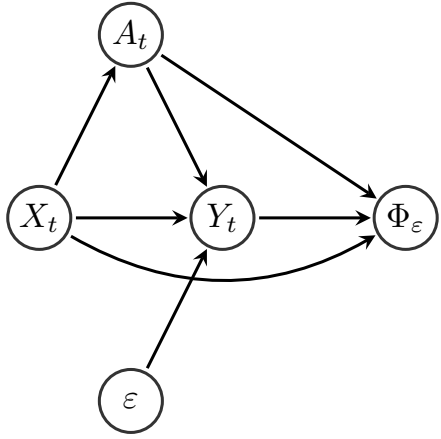
We begin by decomposing the conditional joint probability between $\Phi_\varepsilon$ and $A_t$ by using $Y_t$; for any $\epsilon$, we therefore have:
\begin{align*}
    P(\Phi_\varepsilon=\epsilon,A_t|X_t) = &\int_{Y_t} P(A_t, Y_t|X_t) P(\Phi_\varepsilon=\epsilon|X_t,A_t,Y_t) \:d Y_t
\end{align*}
Recall that we sampled $\Phi_\varepsilon$ from $p(\varepsilon|X_t,A_t,Y_t)$; since we assume an exact model and posterior, $p(\varepsilon|X_t,A_t,Y_t)=P(\varepsilon|X_t,A_t,Y_t)$, and from this we obtain
\begin{align*}
    P(\Phi_\varepsilon=\epsilon,A_t=a|X_t)= & \int_{Y_t} P(A_t, Y_t|X_t) P(\varepsilon=\epsilon|X_t,A_t,Y_t) \:d Y_t\\
    = & \int_{Y_t} p(\varepsilon=\epsilon, A_t, Y_t|X_t) \:d Y_t
\end{align*}

Recall that under the prior, $\varepsilon$ and $A_t$ are conditionally independent, which implies $P(\varepsilon=\epsilon, A_t, Y_t|X_t)=P(\varepsilon=\epsilon|X_t) P(A_t|X_t) P(Y_t|A_t,X_t,\varepsilon=\epsilon)$, which results in 
\begin{align*}
    P(\Phi_\varepsilon=\epsilon,A_t=a|X_t)= 
    & \int_{Y_t} P(\varepsilon=\epsilon|X_t) P(A_t|X_t) P(Y_t|X_t, A_t, \varepsilon=\epsilon) \:d Y_t\\
    = & P(\varepsilon=\epsilon|X_t) P(A_t|X_t) \int_{Y_t}  P(Y_t|X_t, A_t, \varepsilon=\epsilon) \:d Y_t = P(\varepsilon=\epsilon|X_t) P(A_t|X_t)
\end{align*}
from which independence follows. 
For the proof of the corollary, consider an i.i.d. sequence $\Phi_\varepsilon^1,\Phi_\varepsilon^2,\ldots$. 
Note the sequence itself \emph{is not} independent from the action - indeed, a large enough sequence captures exactly the information of the distribution $P(\epsilon|X_t,A_t,Y_t)$, which is ostensibly and intuitively a function of the action.
However, from the theorem above, each $\Phi_\varepsilon^i$ is independent from the action, and therefore $V(X_t, \Phi_\varepsilon^i)$ is valid baseline, i.e. uncorrelated to the action (and the score). Their average is therefore also a valid baseline (since if two variables are both uncorrelated to a third variable, so is their sum), and from the strong law of large numbers, $\int_\epsilon V(X_t,\epsilon) p(\epsilon|X_t,A_t, Y_t)$ is also uncorrelated to the action and therefore a valid baseline.

\end{proof}

\FloatBarrier
\section{Variance analysis}

\subsection{Relation between variance of advantage and variance of policy gradient}

Consider an advantage estimate $Y_t$, i.e. a variable such that $\E[Y_t|X_t=x, A_t=a] = Q(x,a)-V(x)$.  Possible choices for $Y_t$ include the CCA estimate $G_t-V(X_t,\Phi_t)$ as well as the actual advantage $\mathcal{A}(x,a)=Q(x,a)-V(x)$.
Note that $\grad_\theta \mathbb{E}[G_t] = \E[\sum_t \gamma^t S_t Y_t]$. We aim to analyze the variance of a single term $S_t Y_t$ (understanding the variance of the sum is more involved). More precisely, we compare the variance $\mathbb{V}[S_t Y_t|X_t]$ of the policy gradient term $S_t Y_t$ given $X_t$ when using $Y_t$ to that of $S_t \mathcal{A}_t$.

We use the conditional variance formula:
\begin{align*}
    \mathbb{V}[S_t Y_t|X_t] = \E[\mathbb{V}[S_t Y_t|X_t, A_t]|X_t] + \mathbb{V}[\E[S_t Y_t|X_t, A_t]|X_t],
\end{align*}
where $S_t$ is constant given $X_t,A_t$. 
Therefore the first term becomes $ \E[S_t^2 \mathbb{V}[Y_t|X_t, A_t]|X_t]$, and the second one $\mathbb{V}[S_t\E[Y_t|X_t, A_t]|X_t]=\mathbb{V}[S_t \mathcal{A}_t|X_t]$; this term does not depend on the actual advantage estimate used - it is equal to the variance of the policy gradient estimate when using the exact advantage $\mathcal{A}_t$. The additional variance incurred by using an unbiased advantage estimate $Y_t$ instead of the exact advantage $\mathcal{A}_t$ is therefore:
\begin{align*}
    \mathbb{V}[S_t Y_t|X_t]-\mathbb{V}[S_t \mathcal{A}_t|X_t]=\E[S_t^2 \mathbb{V}[Y_t|X_t, A_t]|X_t].
\end{align*}
We see that the (conditional) advantage variance $\mathbb{V}[Y_t|X_t, A_t]$ (as well as the variance of the score function, and their correlation) drives the variance of the policy gradient estimator.  We can further find a loose upper bound purely in terms of the unconditional variance of the advantage. First, suppose that the actions are discrete, and that the action distribution is parametrized by a softmax over logits $l_1,\ldots, l_k$, where $k$ is the number of actions. 
Note that the score is $S_t= \dbyd{\log \pi(A_t)}{\theta}=\sum_{a'}\dbyd{\log \pi(A_t)}{l_{a'}} \dbyd{l_{a'}}{\theta}$, so 
\begin{align*}
|S_t| =& |\sum_{a'}\dbyd{\log \pi(A_t)}{l_{a'}} \dbyd{l_{a'}}{\theta}|\\
\leq& \sum_{a'}|\dbyd{\log \pi(A_t)}{l_{a'}}| |\dbyd{l_{a'}}{\theta}|\\
\leq& \sum_{a'} |\dbyd{l_{a'}}{\theta}| \leq ||J||_1
\end{align*}
where $J$ is the jacobian of the function mapping parameters $\theta$ to logits. The second inequality is due to $|\dbyd{\log \pi(a)}{l_{a'}}|=|\delta_{a,a'}-\pi(a')|\leq 1$. It follows that:
\begin{align*}
    \mathbb{V}[S_t Y_t|X_t]-\mathbb{V}[S_t \mathcal{A}_t|X_t]\leq & ||J||_1^2 \: \E[ \mathbb{V}[Y_t|X_t, \mathcal{A}_t]|X_t]\\
    \leq &||J||_1^2 \: (\mathbb{V}[Y_t|X_t] - \mathbb{V}[\mathcal{A}_t|X_t]))
\end{align*}
again using the law of conditional variance $\mathbb{V}[Y_t|X_t] = \mathbb{V}[\E[Y_t|X_t,\mathcal{A}_t]|X_t]+\mathbb{E}[\mathbb{V}[Y_t|X_t,\mathcal{A}_t]|X_t]$. We thus see that the excess variance incurred by using $Y_t$ in the policy gradient estimate can be upper bounded by a constant times the excess variance of the advantage estimate.

\subsection{Variance analysis in the bandit problem}
Here we provide a back-of-the-envelope variance analysis of the bandit problem. For simplicity (but reasoning can easily be extended), we assume no context and only two actions $\{0,1\}$, and three vectors $W, V_0, V_1\in \mathbb{R}^K$ (randomly sampled from a Gaussian and kept constant across all episodes).
$\epsilon_r$ and $\epsilon_f$ are the reward and observation noise respectively, with standard deviations $\sigma_r \gg  \sigma_f$.
The feedback vector for action $a$ is $W\epsilon_r + V_a + \epsilon_f$. The reward for action $0$ is $\epsilon_r$, and for action $1$ is $\epsilon_r + 1$.

A forward (in this case, constant) baseline for this problem will have square advantage roughly scale as a $\sigma_r^2$.

Let's consider linear hindsight baseline $\alpha^T F$, which is equal to $\epsilon_r (\alpha^T W) + \alpha^T V_a + \alpha^T \epsilon_f$. The expected square advantage  $\E[(G-\alpha^T F)^2]$ is therefore
\begin{align*}
\E[(G-\alpha^T F)^2]=&\pi_0\E[\left(\epsilon_r (\alpha^T W) + \alpha^T  V_0 + \alpha^T \epsilon_f\right)^2]+\pi_1\E[\left(\epsilon_r (\alpha^T W-1) + (\alpha^T  V_1-1) + \alpha^T \epsilon_f\right)^2] \\
=& (\alpha^T W-1)^2 \sigma_r^2 + (\alpha^T \alpha) \sigma_f^2+\pi_0 (\alpha^T  V_0)^2 + \pi_1 (\alpha^T  V_1-1)^2
\end{align*}
The vectors $W, V_0$ and $V_1$ are independent with probability one (in fact they are nearly orthogonal), one can find a hindsight baseline such that $\alpha^T W-1=\alpha^T  V_0=\alpha^T  V_1-1=0$, which leaves an expected squared advantage of $\sigma_f^2 \alpha^T \alpha$ which is small (for random vectors the matrices will be well-conditioned, the resulting $\alpha$ will have small norm); however that advantage leads to a biased update since the advantage is independent of the action. However, choosing $\alpha^T W=1$ but $\alpha^T V_0=\alpha^T V_1=0$ leads to a hindsight baseline which is equal to $\epsilon_r + \alpha^T \epsilon_f$, independent from the action; the effect of the noise $\epsilon_r$ will be removed entirely from the squared advantage, leading to an unbiased gradient estimator with a considerably lower variance (of order $\sigma_f^2$).

\FloatBarrier

\section{RL algorithms, common randomness, structural causal models}\label{sec:causalRL}

In this section, we provide an alternative view and intuition behind the CCA-PG algorithm by investigating credit assignment through the lens of causality theory, in particular \emph{structural causal models} (SCMs)~\citep{pearl2009causality}. These ideas are very related to the use of common random numbers (CRN), a standard technique in optimization with simulators~\citep{glasserman1992some}.

\subsection{Structural causal model of the MDP}

\begin{figure*}[h!]
    \centering
    \resizebox{0.3\textwidth}{!}{%
    \begin{tikzpicture}[scale=0.85, >=stealth]
\tikzstyle{empty}=[]
\tikzstyle{lat}=[circle, fill=black!10, inner sep=1pt, minimum size = 6.5mm, thick, draw =black!80, node distance = 20mm, scale=0.9]
\tikzstyle{directed}=[->, thick, shorten >=0.5 pt, shorten <=1 pt]

\node[lat] at (0,0) (s0) {$X_{0}$};
\node[lat] at (2,0) (s1) {$X_{1}$};
\node[lat] at (4,0) (s2) {$X_{2}$};
\node[lat] at (6,0) (s3) {$X_{3}$};

\node[lat] at (1,-1) (a0) {$A_{0}$};
\node[lat] at (3,-1) (a1) {$A_{1}$};
\node[lat] at (5,-1) (a2) {$A_{2}$};

\node[lat] at (0, 1.25) (r0) {$R_{0}$};
\node[lat] at (2, 1.25) (r1) {$R_{1}$};
\node[lat] at (4, 1.25) (r2) {$R_{2}$};
\node[lat] at (6, 1.25) (r3) {$R_{3}$};

\path   (s0) edge [directed] (s1)
        (s1) edge [directed] (s2)
        (s2) edge [directed] (s3)
        (s0) edge [directed] (a0)
        (s1) edge [directed] (a1)
        (s2) edge [directed] (a2)
        (a0) edge [directed] (s1)
        (a1) edge [directed] (s2)
        (a2) edge [directed] (s3)
        (s0) edge [directed] (r0)
        (s1) edge [directed] (r1)
        (s2) edge [directed] (r2)
        (s3) edge [directed] (r3);
\end{tikzpicture}} 
    \hspace{5cm}
    \resizebox{0.3\textwidth}{!}{%
    \begin{tikzpicture}[scale=0.85, >=stealth]
\tikzstyle{empty}=[]
\tikzstyle{lat}=[circle, inner sep=1pt, minimum size = 6.5mm, thick, draw =black!80, node distance = 20mm, scale=0.75]
\tikzstyle{obs}=[circle, fill =black!10, inner sep=1pt, minimum size = 6.5mm, thick, draw =black, node distance = 20mm, scale=0.75]
\tikzstyle{detobs}=[rectangle, fill =black!10, inner sep=1pt, text=black, minimum size = 6mm, thick, draw=black, node distance = 20mm, scale=0.75]
\tikzstyle{detlat}=[rectangle, fill =white, inner sep=1pt, text=black, minimum size = 6mm, thick, draw=black, node distance = 20mm, scale=0.75]
\tikzstyle{connect}=[-latex, thick]
\tikzstyle{undir}=[thick]
\tikzstyle{directed}=[->, thick, shorten >=0.5 pt, shorten <=1 pt]

\node[lat] at (0,2) (us0) {$\epsilon_0^X$};
\node[detobs] at (0,1) (s0) {$X_{0}$};

\node[obs] at (0.5,0) (ua0) {$\epsilon^\pi_{0}$};
\node[detobs] at (1.5,0) (a0) {$A_0$};

\node[lat] at (2,2) (us1) {$\epsilon^X_{1}$};
\node[detobs] at (2,1) (s1) {$X_1$};

\node[obs] at (2.5,0) (ua1) {$\epsilon^\pi_1$};
\node[detobs] at (3.5,0) (a1) {$A_1$};

\node[lat] at (4,2) (us2) {$\epsilon^X_2$};
\node[detobs] at (4,1) (s2) {$X_2$};

\node[empty] at (5,1) (s3) [] {};

\path   (us0) edge [directed] (s0)
        (us1) edge [directed] (s1)
        (ua0) edge [directed] (a0)
        (s0) edge [directed] (a0)
        (a0) edge [directed] (s1)
        (s0) edge [directed] (s1)
        (s1) edge [directed] (a1)
        (ua1) edge [directed] (a1)
        (a1) edge [directed] (s2)
        (us2) edge [directed] (s2)        
        (s1) edge [directed] (s2)
        (s2) edge [-, thick] (s3)
        ;
\end{tikzpicture}} \\
    \resizebox{0.3\textwidth}{!}{%
    \begin{tikzpicture}[scale=0.85, >=stealth]
\tikzstyle{empty}=[]
\tikzstyle{latobs}=[circle, fill=black!10, inner sep=1pt, minimum size = 6.5mm, thick, draw =black!80, node distance = 20mm, scale=0.9]
\tikzstyle{lat}=[circle, inner sep=1pt, minimum size = 6.5mm, thick, draw =black!80, node distance = 20mm, scale=0.9]
\tikzstyle{det}=[rectangle, fill =black!10, inner sep=1pt, text=black, minimum size = 6mm, thick, draw=black, node distance = 20mm, scale=0.9]
\tikzstyle{directed}=[->, thick, shorten >=0.5 pt, shorten <=1 pt]

\node[lat] at (0,-2) (e0) {$E_{0}$};
\node[lat] at (2,-2) (e1) {$E_{1}$};
\node[lat] at (4,-2) (e2) {$E_{2}$};
\node[lat] at (6,-2) (e3) {$E_{3}$};

\node[latobs] at (0,-1) (o0) {$O_{0}$};
\node[latobs] at (2,-1) (o1) {$O_{1}$};
\node[latobs] at (4,-1) (o2) {$O_{2}$};
\node[latobs] at (6,-1) (o3) {$O_{3}$};

\node[det] at (0,0) (s0) {$X_{0}$};
\node[det] at (2,0) (s1) {$X_{1}$};
\node[det] at (4,0) (s2) {$X_{2}$};
\node[det] at (6,0) (s3) {$X_{3}$};

\node[latobs] at (1,-1) (a0) {$A_{0}$};
\node[latobs] at (3,-1) (a1) {$A_{1}$};
\node[latobs] at (5,-1) (a2) {$A_{2}$};

\node[latobs] at (0, 1.25) (r0) {$R_{0}$};
\node[latobs] at (2, 1.25) (r1) {$R_{1}$};
\node[latobs] at (4, 1.25) (r2) {$R_{2}$};
\node[latobs] at (6, 1.25) (r3) {$R_{3}$};

\path   (s0) edge [directed] (s1)
        (s1) edge [directed] (s2)
        (s2) edge [directed] (s3)
        (s2) edge [directed] (s3)
        (e0) edge [directed] (e1)
        (e1) edge [directed] (e2)
        (e2) edge [directed] (e3)
        (e0) edge [directed] (o0)
        (e1) edge [directed] (o1)
        (e2) edge [directed] (o2)
        (e3) edge [directed] (o3)
        (o0) edge [directed] (s0)
        (o1) edge [directed] (s1)
        (o2) edge [directed] (s2)
        (o3) edge [directed] (s3)        
        (s0) edge [directed] (a0)
        (s1) edge [directed] (a1)
        (s2) edge [directed] (a2)
        (a0) edge [directed] (e1)
        (a1) edge [directed] (e2)
        (a2) edge [directed] (e3)
        (a0) edge [directed] (s1)
        (a1) edge [directed] (s2)
        (a2) edge [directed] (s3)        
        (s0) edge [directed] (r0)
        (s1) edge [directed] (r1)
        (s2) edge [directed] (r2)
        (s3) edge [directed] (r3);
\end{tikzpicture}} 
    \resizebox{0.65\textwidth}{!}{%
    \begin{tikzpicture}[scale=1.0, >=stealth]
\tikzstyle{empty}=[]
\tikzstyle{lat}=[circle, inner sep=1pt, minimum size = 6.5mm, thick, draw =black!80, node distance = 20mm, scale=0.75]
\tikzstyle{obs}=[circle, fill =black!10, inner sep=1pt, minimum size = 6.5mm, thick, draw =black, node distance = 20mm, scale=0.75]
\tikzstyle{detobs}=[rectangle, fill =black!10, inner sep=1pt, text=black, minimum size = 6mm, thick, draw=black, node distance = 20mm, scale=0.75]
\tikzstyle{detlat}=[rectangle, fill =white, inner sep=1pt, text=black, minimum size = 6mm, thick, draw=black, node distance = 20mm, scale=0.75]
\tikzstyle{connect}=[-latex, thick]
\tikzstyle{undir}=[thick]
\tikzstyle{directed}=[->, thick, shorten >=0.5 pt, shorten <=1 pt]

\node[empty] at (-2,2) (hm1) [] {};
\node[detlat] at (0,0) (s0) {$E_{0}$};
\node[lat] at (0,-1) (us0) {$\epsilon_0^S$};
\node[detobs] at (1,1) (o0) {$O_1$};
\node[lat] at (0,1) (uo0) {$\epsilon^O_{1}$};
\node[detobs] at (2,2) (h0) {$X_1$};
\node[detobs] at (3,1) (a0) {$A_1$};
\node[obs] at (2,1) (ua0) {$\epsilon^\pi_{1}$};
\node[detlat] at (4,0) (s1) {$E_2$};
\node[lat] at (4,-1) (us1) {$\epsilon^S_{2}$};
\node[detobs] at (5,1) (o1) {$O_2$};
\node[lat] at (4,1) (uo1) {$\epsilon^O_2$};
\node[detobs] at (6,2) (h1) {$X_2$};
\node[detobs] at (7,1) (a1) {$A_2$};
\node[obs] at (6,1) (ua1) {$\epsilon^\pi_2$};
\node[detlat] at (8,0) (s2) {$E_3$};
\node[lat] at (8,-1) (us2) {$\epsilon^S_3$};
\node[empty] at (9,2) (h2) [] {};
\node[empty] at (9,1.85) (h2a) [] {};
\node[empty] at (9,1) (o2) [] {};
\node[empty] at (10,2) (a2) [] {};
\node[empty] at (9,0) (s3) [] {};
\path   (s0) edge [directed] (s1)
        (s1) edge [directed] (s2)
        (s2) edge [-, thick] (s3)
        (s2) edge [-, thick] (o2)
        (s0) edge [directed] (o0)
        (o0) edge [directed] (h0)
        (uo0) edge [directed] (o0)
        (ua0) edge [directed] (a0)
        (s1) edge [directed] (o1)
        (o1) edge [directed] (h1)
        (h0) edge [directed] (a0)
        (a0) edge [directed] (s1)
        (h1) edge [directed] (a1)
        (a1) edge [directed] (s2)
        (a0) edge [directed, bend left=10] (h1)
        (a1) edge [directed, bend left=10, -] (h2a)
        (us0) edge [directed] (s0)
        (us1) edge [directed] (s1)
        (uo1) edge [directed] (o1)
        (ua1) edge [directed] (a1)
        (us2) edge [directed] (s2)
        (h0) edge [directed] (h1)
        (h1) edge [-, thick] (h2)
        ;
\end{tikzpicture}} 
    \caption{Graphical models and corresponding SCMs for RL problems. Top: MDP, bottom: POMDP; left: graphical model, right: structural causal model. Squares represent deterministic nodes, while circles represent stochastic nodes. Observed nodes are shaded in gray.}
    \label{figSCMs}
\end{figure*}
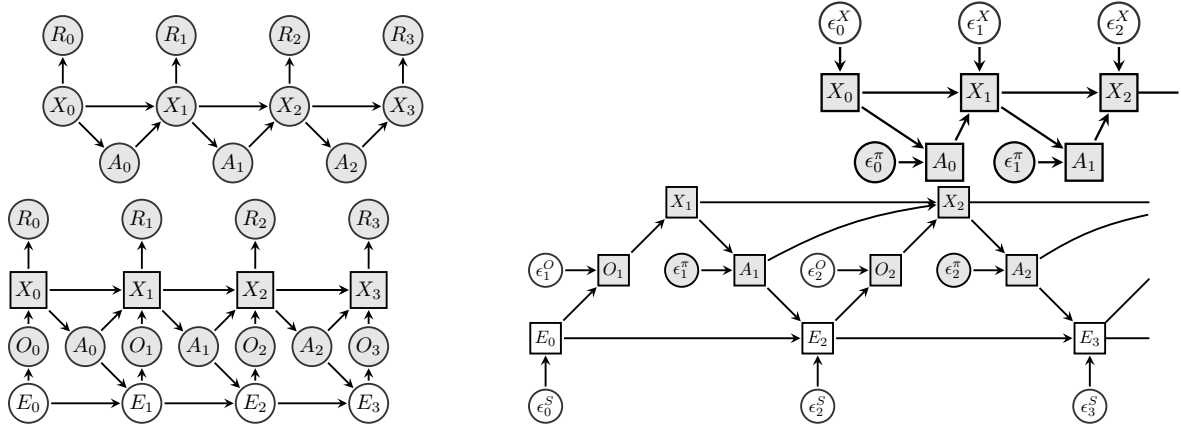

\emph{Structural causal models} (SCM) \citep{pearl2009causality} are, informally, models where all randomness is exogenous, and where all variables of interest are modeled as deterministic functions of other variables and of the exogenous randomness. 
They are of particular interest in causal inference as they enable reasoning about interventions, i.e.\ how would the \emph{distribution} of a variable change under external influence (such as forcing a variable to take a given value, or changing the process that defines a variable), and about counterfactual interventions, i.e.\ how would a particular observed outcome (sample) of a variable have changed under external influence.
Formally, a SCM is a collection of model variables $\{V \in \bm{V}\}$, exogenous random variables $\{\calE \in \bm{\calE}\}$, and distributions $\{p_\calE(\varepsilon), \calE \in \bm{\calE}\}$, one per exogenous variable, and where the exogenous random variables are all assumed to be independent.
Each variable $V$ is defined by a function $V=f_V(\pa(V), \bm{\calE})$, where $\text{pa}(V)$ is a subset of $\bm{V}$ called the parents of $V$. The model can be represented by a directed graph in which every node has an incoming edge from each of its parents. For the SCM to be valid, the induced graph has to be a directed acyclic graph (DAG), i.e.\ there exists a topological ordering of the variables such that for any variable $V_i$, $\pa(V_i) \subset \{V_1, \ldots, V_{i-1}\}$; in the following we will assume such an ordering. This provides a simple sampling mechanism for the model, where the exogenous random variables are first sampled according to their distribution, and each node is then computed in topological order. Note that any probabilistic model can be represented as a SCM by virtue of reparametrization \cite{kingma2014adam, buesing2018woulda}. However, such a  representation is not unique, i.e.\ different SCMs can induce the same distribution.

In the following we give an SCM representation of a MDP (see Fig.\ref{figSCMs} for the causal graphical model and corresponding SCM for MDPs and POMDPs). The transition from $X_t$ to $X_{t+1}$ under $A_t$ is given by the transition function $\ft$: $X_{t+1}=\ft(X_t,A_t,\calE_t^X)$ with exogenous variable / random number $\calE_t^X$. The policy function $\fpi$ maps a random number $\calE_t^\pi$, policy parameters $\theta$, and current state $X_t$ to the action $A_t=\fpi(X_t, \calE_t^\pi, \theta)$. Together, $\fpi$ and $\calE_t^\pi$ induce the policy, a distribution $\pi_\theta(A_t|X_t)$ over actions. Without loss of generality we assume that the reward is a deterministic function of the state and action: $R_t=f^R(X_t, A_t)$. $\calE^X$ and $\calE^\pi$ are random variables with a fixed distribution; all changes to the  policy are absorbed by changes to the deterministic function $\fpi$. 
Denoting $\calE_t=(\calE_t^X, \calE_t^\pi)$, note the next reward and state $(X_{t+1}, R_t)$ are deterministic functions of $X_t$ and $\calE_t$, since we have $X_{t+1}=\ft(X_t, \fpi(X_t, \calE_t^\pi, \theta), \calE_t^X)$ and similarly $R_t=R(X_t, \fpi(X_t, \calE_t^\pi, \theta)$. Let $\Xtp=(X_{t'})_{t'> t}$ and similarly, $\Etp=(\calE^X_{t},\calE_{t'})_{t'> t}$ 
Through the composition of the functions $f^X$, $f^\pi$ and $R$, the return $G_t$ (under policy $\pi$) is a deterministic function $f^G$ of $X_t$, $A_t$ and $\Etp$.

\subsection{Proof of theorem~\ref{thm:causal}}

For notation purposes, in the rest of this section, we will focus on credit assignment for action $A_t$ (since policy gradient terms are additive with respect to time), and will denote $X=X_t$, $A=A_t$, $\varepsilon=\Etp$, and $\tau=(X_s,R_r,A_s)_{s\geq t}$.
Furthermore, we will denote $\Phi=\Phi_t$.

From the arguments in the section above, one can write $\tau=f^\tau(X,A,\varepsilon)$, $G=f^G(\tau)$, and $\Phi=f^\phi(\tau)$.  We may integrate out $\tau$, in which case the graph only contains $X,A,\varepsilon$ and $G$. In that graph, by the faithfulness assumption, there can be no causal path from $A$ to $\Phi$, as this would violate the conditional independence assumption. It follows that there are functions $g_G$ and $g_\Phi$ such that $G=g_G(X,A,\varepsilon)$ and $\Phi=g_\Phi(X,\varepsilon)$. 

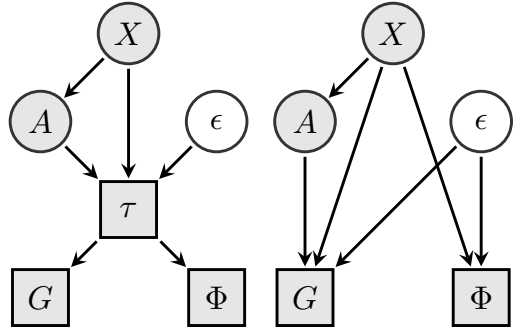
\begin{figure}[h!]
    \centering
    \resizebox{0.2\textwidth}{!}{\begin{tikzpicture}[scale=0.85, >=stealth]
\tikzstyle{empty}=[]
\tikzstyle{latobs}=[circle, fill=black!10, inner sep=1pt, minimum size = 6.5mm, thick, draw =black!80, node distance = 20mm, scale=0.9]
\tikzstyle{lat}=[circle, inner sep=1pt, minimum size = 6.5mm, thick, draw =black!80, node distance = 20mm, scale=0.9]
\tikzstyle{det}=[rectangle, fill =black!10, inner sep=1pt, text=black, minimum size = 6mm, thick, draw=black, node distance = 20mm, scale=0.9]
\tikzstyle{directed}=[->, thick, shorten >=0.5 pt, shorten <=1 pt]

\node[latobs] at (0,0) (x) {$X$};
\node[latobs] at (-1,-1) (a) {$A$};
\node[lat] at (1,-1) (eps) {$\epsilon$};
\node[det] at (0,-2) (tau) {$\tau$};
\node[det] at (-1, -3) (G) {$G$};
\node[det] at (1, -3) (phi) {$\Phi$};

\path   (x) edge [directed] (a)
        (x) edge [directed] (tau)
        (a) edge [directed] (tau)
        (eps) edge [directed] (tau)
        (tau) edge [directed] (G)
        (tau) edge [directed] (phi);
\end{tikzpicture} }
    \resizebox{0.2\textwidth}{!}{\begin{tikzpicture}[scale=0.85, >=stealth]
\tikzstyle{empty}=[]
\tikzstyle{latobs}=[circle, fill=black!10, inner sep=1pt, minimum size = 6.5mm, thick, draw =black!80, node distance = 20mm, scale=0.9]
\tikzstyle{lat}=[circle, inner sep=1pt, minimum size = 6.5mm, thick, draw =black!80, node distance = 20mm, scale=0.9]
\tikzstyle{det}=[rectangle, fill =black!10, inner sep=1pt, text=black, minimum size = 6mm, thick, draw=black, node distance = 20mm, scale=0.9]
\tikzstyle{directed}=[->, thick, shorten >=0.5 pt, shorten <=1 pt]

\node[latobs] at (0,0) (x) {$X$};
\node[latobs] at (-1,-1) (a) {$A$};
\node[lat] at (1,-1) (eps) {$\epsilon$};
\node[det] at (-1, -3) (G) {$G$};
\node[det] at (1, -3) (phi) {$\Phi$};

\path   (x) edge [directed] (a)
        (x) edge [directed] (G)
        (x) edge [directed] (phi)
        (a) edge [directed] (G)
        (eps) edge [directed] (G)
        (eps) edge [directed] (phi);
\end{tikzpicture} }
    \caption{SCMs for the reduced action selection problem; left: including the trajectory; right: trajectory is integrated out. There is no arrow from $A$ to $\Phi$ on the right since the graph is assumed to be faithful and $A$ and $\Phi$ are conditionally independent given $X$.}
    \label{fig.PhiSCM}
\end{figure}

The resulting structural causal models can be seen in Fig.~\ref{fig.PhiSCM}.

The conditional expectation $Q(x,a,\phi)$ is given by $Q(x,a,\phi) = \int_\varepsilon p(\varepsilon|x,\phi,a) G(x,a,\varepsilon)$
The counterfactual return for action $a$, having observed $\phi$ is given by $\mathbb{E}[G(\tau')| \tau' \sim P(\tau'|X=x, \text{observe}(\Phi=\phi))]$ is equal to $\int_\varepsilon p(\varepsilon|x,\phi) G(x,a,\phi)$.

Finally, note that from d-separation $p(\varepsilon|x,\phi)=p(\varepsilon|x,a,\phi)$, and the result follows.

\section{Individual Treatment Effects, (Conditional) Average Treatment Effects, Counterfactuals and Counterfactual identifiability}\label{sec:simple_ex_causality}

In this section, we will further link the ideas developed in this report to causality theory. In particular we will connect them to two notions of causality theory known as individual treatment effect (ITE) and average treatment effect (ATE). In the previous section, we extensively leveraged the framework of structural causal models. It is however known that distinct SCMs may correspond to the same distribution; learning a model from data, we may learn a model with correct distribution but with with incorrect structural parametrization and counterfactuals. We may therefore wonder whether counterfactual-based approaches may be flawed when using such a model. We investigate this question, and analyze our algorithm in very simple settings for which closed-form computations can be worked out.

\subsection{Individual and Average Treatment Effects}

\begin{figure}[h!]
\centering
\centering
    \resizebox{0.35\textwidth}{!}{%
        \begin{tikzpicture}[scale=0.85, >=stealth]
\tikzstyle{empty}=[]
\tikzstyle{cir}=[circle, inner sep=1pt, minimum size = 6.5mm, thick, draw =black!80, node distance = 20mm, scale=0.9]
\tikzstyle{sqr}=[rectangle, inner sep=1pt, minimum size = 6.5mm, thick, draw =black!80, node distance = 20mm, scale=0.9]
\tikzstyle{directed}=[->, thick, shorten >=0.5 pt, shorten <=1 pt]

\node[cir] at (0,0) (s) {$S$};
\node[sqr] at (1.5,0) (h) {$H$};
\node[cir] at (2, 1) (eps) {$\epsilon$};
\node[sqr] at (3,0) (t) {$T$};
\node[sqr] at (5,0) (o) {$O$};
\node[sqr] at (5,-1) (f) {$F$};

\path   (s) edge [directed] (h)
        (h) edge [directed] (t)
        (eps) edge [directed] (t)
        (s) [bend right] edge [directed] (o)
        (s) [bend right] edge [directed] (f)
        (t) [bend right] edge [directed] (o)
        (t) [bend right] edge [directed] (f);
\end{tikzpicture}
    }  
\caption{The medical treatment example as a structured causal model.
    }
\label{fig:medEX}
\end{figure}
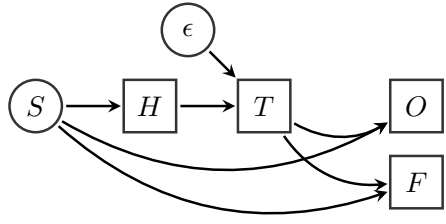

Consider a simple medical example which we model with an SCM as illustrated in Fig.~\ref{fig:medEX}. We assume population of patients, each with a full medical state denoted $S$, which summarizes all factors, known or unknown, which affect a patient's future health such as genotype, phenotype etc. While $S$ is never known perfectly, some of the patient's medical history $H$ may be known, including current symptoms. On the basis of $H$, a treatment decision $T$ is taken; as is often done, for simplicity we consider $T$ to be a binary variable taking values in \{$1$=`treatment', $0$=`no treatment'\}. Finally, health state $S$ and treatment $T$ result in a observed medical outcome $O$, a binary variable taking values in \{$1$=`cured', $0$=`not cured'\}. For a given value $S=s$ and $T=t$, the outcome is a function (also denoted $O$ for simplicity) $O(s,t)$.
Additional medical information $F$ may be observed, e.g.\ further symptoms or information obtained after the treatment, from tests such as X-rays, blood tests, or autopsy.

In this simple setting, we can characterize the effectiveness of the treatment for an individual a patient with profile $S$ by the Individual Treatment Effect (ITE) which is defined as the difference between the outcome under treatment and no treatment. 
\begin{defn}[Individual Treatment Effect]
\begin{align}\label{eq:ITE}
    \text{ITE(s)} =& \E[O|S=s,\doop(T=1)]-\E[O|S=s, \doop(T=0)]\notag\\
    =& O(s,T=1)-O(s,T=0)
\end{align}
\end{defn}

The conditional average treatment effect is the difference in outcome between the choice of $T=1$ and $T=0$ when averaging over all patients with the same set of symptoms $H=h$
\begin{defn}[Conditional Average Treatment Effect]
\begin{align}\label{eq:ATE}
        \text{ATE}(h)=& \E[O|H=h, \doop(T=1)]-\E[O|H=h, \doop(T=0)]\notag\\
        =& \int_s p(S=s|H=h) (O(s,T=1)-O(s,T=0))
\end{align}
\end{defn}

Since the exogenous noise (here, $S$) is generally not known, the ITE is typically an unknowable quantity. For a particular patient (with hidden state $S$), we will only observe the outcome under $T=0$ or $T=1$, depending on which treatment option was chosen; the counterfactual outcome will typically be unknown. Nevertheless, for a given SCM, it can be counterfactually estimated from the outcome and feedback.

\begin{defn}[Counterfactually Estimated Individual Treatment Effect]
\begin{align}
    \textrm{CF-ITE}[H=h,F=f,T=1] =& \delta(o=1)-\int_{s'} P(S=s'|H=h,F=f,T=1) O(s', T=0)\\
    \textrm{CF-ITE}[H=h,F=f,T=0] =&\int_{s'} P(S=s'|H=h,F=f, T=1) O(s',T=0)-\delta(o=1)
\end{align}
\end{defn}
In general the counterfactually estimated ITE will not be exactly the ITE, since there may be remaining uncertainty on $s$. However, the following  statements relate CF-ITE, ITE and ATE:
\begin{itemize}
    \item If $S$ is identifiable from $O$ and $F$ with probability one, then the counterfactually-estimated ITE is equal to the ITE.
    \item The average (over $S$, conditional on $H$) of the ITE is equal to the ATE.
    \item The average (over $S$ and $F$, conditional on $H$) of CF-ITE is equal to the ATE.
\end{itemize}

Assimilating $O$ to a reward, the above illustrates that the ATE (equation~\ref{eq:ATE}) essentially corresponds to a difference of Q functions, the ITE (equation~\ref{eq:ITE}) to a difference of returns under common randomness, and the counterfactual ITE to CCA-like advantage estimates. In contrast, the advantage estimate $G_t-V(H_t)$ is a difference between a return (a sample-level quantity) and a value function (a population-level quantity, which averages over all individuals with the same medical history $H$); this discrepancy explains why the return-based advantage estimate can have very high variance.

As mentioned previously, for a given joint distribution over observations, rewards and actions, there may exist distinct SCMs that capture that distribution. Those SCMs will all have the same ATE, which measures the effectiveness of a policy on average. But they will generally have different ITE and counterfactual ITE, which, when using model-based counterfactual policy gradient estimators, will lead to different estimators. Choosing the `wrong' SCM will lead to the wrong counterfactual, and so we may wonder if this is a cause for concern for our methods.

\emph{We argue that in terms of learning optimal behaviors (in expectation), estimating inaccurate counterfactual is not a cause for concern.}
Since all estimators have the same expectation, they would all lead to the correct estimates for the effect of switching a policy for another, and therefore, will all lead to the optimal policy given the information available to the agent. In fact, one could go further and argue that for the purpose of finding good policies in expectations, we should only care about the counterfactual for a precise patient inasmuch as it enables us to quickly and correctly taking better actions for future patients for whom the information available to make the decision ($H$) is very similar. This would encourage us to choose the SCM for which the CF-ITE has minimal variance, regardless of the value of the true counterfactual. In the next section, we elaborate on an example to highlight the difference in variance between different SCMs with the same distribution and optimal policy.

\subsection{Betting against a fair coin}

We begin from a simple example, borrowed from \cite{pearlbook}, to show that two SCMs that induce the same interventional and observational distributions can imply different counterfactual distributions. 
The example consists of a game to guess the outcome of a fair coin toss. The action $A$ and state $S$ both take their values in $\{h,t\}$. 
Under model $\mathbf{I}$, the outcome $O$ is $1$ if $A=S$ and $0$ otherwise. 
Under model $\mathbf{II}$, the guess is ignored, and the outcome is simply $O=1$ if $S=h$. 
For both models, the average treatment effect $E[O|A=h]-E[O|A=t]$ is $0$ implying that in both models, one cannot do better than random guessing.
Under model $\mathbf{I}$, the counterfactual for having observed outcome $O=1$ and changing the action, is always $O=0$, and vice-versa (intuitively, changing the guess changes the outcome).
Therefore, the ITE is $\pm1$. 
Under model $\mathbf{II}$, all counterfactual outcomes are equal to the observed outcomes, since the action has in fact no effect on the outcome. The ITE is always $0$. 

In the next section, we will next adapt the medical example into a problem in which the choice of action does affect the outcome. Using the CF-ITE as an estimator for the ATE, we will find how the choice of the SCM affects the variance of that estimator (and therefore how the choice of the SCM should affect the speed at which we can learn which is the optimal treatment decision).

\subsection{Medical example}

Take the simplified medical example from Fig.$\ref{fig:medEX}$, where a population of patients with the same symptoms come to the doctor, and the doctor has a potential treatment $T$ to administer. The state $S$ represents the genetic profile of the patient, which can be one of three $\{\hlaa, \hlab, \hlac\}$ (each with probability $1/3$). We assume that genetic testing is not available and that we do not know the value of $S$ for each patient. The doctor has to make a decision whether to administer drugs to this population or not, based on repeated experiments; in other words, they have to find out whether the average treatment effect is positive or not. We consider the two following models:
\begin{itemize}
    \item In model $\mathbf{I}$, patients of type $\hlaa$ always recover, patients of type $\hlac$ never do, and patients of type $\hlab$ recover if they get the treatment, and not otherwise; in particular, in this model, administering the drug never hurts.

    \item In model $\mathbf{II}$, patients of type $\hlaa$ and $\hlab$ recover when given the drug, but not patients of type $\hlac$; the situation is reversed ($\hlaa$ and $\hlab$ patients do not recover, $\hlac$ do) when not taking the drug. 
\end{itemize}

In both models - the true value of giving the drug is $2/3$, and not giving the drug $1/3$, which leads to an ATE of $1/3$. 
For each model, we will evaluate the variance of the CF-ITE, under one of the four possible treatment-outcome pair. The results are summarized in table~\ref{table:medical-model}. Under model $\mathbf{A}$, the variance of the CF-ITE estimate (which is the variance of the advantage estimate used in CCA-PG gradient) is $1/6$, while it is $1$ under model $\mathbf{B}$, which would imply $\mathbf{A}$ is a better model to leverage counterfactuals into policy decisions.

\begin{table*}[p]
\centering
\resizebox{\textwidth}{!}{
\begin{tabular}{lll|cc|cc|cc|cc|cc|c|}
\cmidrule{1-14}
Treatment & Outcome & Type & \multicolumn{2}{c|}{CF-Prob.} & \multicolumn{2}{c|}{CF-O} & \multicolumn{2}{c|}{ITE} & \multicolumn{2}{c|}{CF-V} &\multicolumn{2}{c|}{CF-ITE} & Var \\
\cmidrule{1-14}

Drug & Cured & $\hlaa$ &\red{1/2}&\blue{1/2}&\red{1}&\blue{0} & \red{0}&\blue{+1} & & & & &   \\
\cmidrule{3-9}
 &  & $\hlab$ &  \red{1/2}&\blue{1/2} & \red{0}&\blue{0} & \red{+1}&\blue{+1} & \red{1/2}&\blue{0} & \red{1/2}&\blue{1} & \\
\cmidrule{3-9}
 &  & $\hlac$ & \red{0}&\blue{0} &\notableentry&\notableentry    & &  & && \red{1/6}  \\

\cmidrule{2-13}

 & Not cured & $\hlaa$ & \red{0}&\blue{0}&\notableentry&\notableentry    & &  & &&\blue{1}\\
\cmidrule{3-9}
 &   & $\hlab$ & \red{0}&\blue{0} &\notableentry&\notableentry& \red{0}&\blue{1}  & \red{0}&\blue{-1}&\\
\cmidrule{3-9}
 &   & $\hlac$ & \red{1}&\blue{1} & \red{0}&\blue{1} & \red{0}&\blue{-1} & & & & &    \\

\cmidrule{1-14}

No Drug & Cured & $\hlaa$ & \red{1}&\blue{0} & \red{1}&\blue{1} & \red{0}&\blue{0} & & & & &    \\
\cmidrule{3-9}
&   & $\hlab$ & \red{0}&\blue{0} &\notableentry&\notableentry& \red{1}&\blue{0}  & \red{0}&\blue{1}&\\
\cmidrule{3-9}
 &   & $\hlac$ & \red{0}&\blue{1} & \red{0}&\blue{0} & \red{1}&\blue{1} & & & & & \red{1/6}   \\

\cmidrule{2-13}

 & Not cured & $\hlaa$ & \red{0}&\blue{1/2} & \red{1}&\blue{1} & \red{-1}&\blue{-1} & & & & & \blue{1} \\
\cmidrule{3-9}
 &   & $\hlab$ & \red{1/2}&\blue{1/2} & \red{1}&\blue{1} & \red{-1}&\blue{-1} & \red{1/2}&\blue{1} & \red{-1/2}&\blue{-1} &   \\
\cmidrule{3-9}
&   & $\hlac$ & \red{1/2}&\blue{0} & \red{0}&\blue{0} & \red{0}&\blue{0} & & & & &   \\
\cmidrule{1-14}
\end{tabular}}
\caption{CCA-PG variance estimates in the medical example. CF-Probs. \red{Red} value are estimates for model $\mathbf{I}$, \blue{blue} ones are for model $\mathbf{II}$. CF-Prob denotes posterior probabilities of the genetic state $S$ given the treatment $T$ and outcome $O$. CF-O is the counterfactual outcome. The ITE is the individual treatment effect (difference between outcome and counterfactual outcome). CF-V is the counterfactual value function, computed as the average of CF-O under the posterior probabilities for $S$. CF-ITE is the counterfactual advantage estimate (difference between O and CF-V). Var is the variance of CF-ITE under the prior probabilities for the outcome. }
\label{table:medical-model}
\end{table*}

\end{document}